\documentclass[manuscript]{acmart}

\AtBeginDocument{%
  \providecommand\BibTeX{{%
    \normalfont B\kern-0.5em{\scshape i\kern-0.25em b}\kern-0.8em\TeX}}}

\setcopyright{acmcopyright}
\copyrightyear{2024}
\acmYear{2024}
\acmDOI{XXXXXXX.XXXXXXX}

\acmConference[FAccT '24]{ACM Conference on Fairness, Accountability, and Transparency}{June 03--06,
  2024}{Rio de Janeiro, Brazil}
\acmBooktitle{FAccT '24: ACM Conference on Fairness, Accountability, and Transparency,
 June 03--06, 2024, Rio de Janeiro, Brazil} 
\acmPrice{15.00}
\acmISBN{978-1-4503-XXXX-X/18/06}

\usepackage{color-edits}
\addauthor[Nil-Jana]{na}{blue}
\addauthor[Alex]{alex}{green}

\usepackage{chngcntr}

\usepackage{array}
\newcolumntype{L}[1]{>{\raggedright\let\newline\\\arraybackslash\hspace{0pt}}m{#1}}
\newcolumntype{C}[1]{>{\centering\let\newline\\\arraybackslash\hspace{0pt}}m{#1}}
\newcolumntype{R}[1]{>{\raggedleft\let\newline\\\arraybackslash\hspace{0pt}}m{#1}}

\usepackage{tikz}
\tikzset{>=latex}
\usetikzlibrary{patterns}

\usepackage{caption}
\usepackage{subcaption}

\usepackage{bm}
\usepackage{enumitem}
\usepackage{array}
\newcolumntype{P}[1]{>{\raggedright\arraybackslash}p{#1}}

\usepackage{dsfont}
\newcommand{\E}[1]{\mathds{E}\left[{#1}\right]}
\newcommand{\V}[1]{\mathds{V}\left[{#1}\right]}
\newcommand{\Cov}[2]{\text{Cov}\left[{#1},{#2}\right]}
\newcommand{\R}{\mathds{R}}

\usepackage{amsthm}
\newtheorem{theorem}{Theorem}
\newtheorem{lemma}[theorem]{Lemma}
\newtheorem{definition}[theorem]{Definition}
\newtheorem{proposition}[theorem]{Proposition}
\newtheorem{corollary}[theorem]{Corollary}

\usepackage{colortbl}
\usetikzlibrary{calc}
\usepackage{zref-savepos}
\newcounter{NoTableEntry}
\renewcommand*{\theNoTableEntry}{NTE-\the\value{NoTableEntry}}

\begin{document}

\title{The Impact of Differential Feature Under-reporting on Algorithmic Fairness}

\author{Nil-Jana Akpinar}
\email{nakpinar@amazon.com}
\affiliation{%
  \institution{Amazon AWS AI/ML}
  \streetaddress{}
  \country{USA}
  \postcode{}
  \footnote{}
}
\footnotetext[1]{Work done while at Carnegie Mellon University.}
\author{Zachary C. Lipton}
\affiliation{%
  \institution{Carnegie Mellon University}
  \streetaddress{}
  \country{USA}}

\author{Alexandra Chouldechova}
\affiliation{%
  \institution{Carnegie Mellon University}
  \country{USA}
}
\renewcommand{\shortauthors}{Akpinar et al.}

\begin{abstract}
    Predictive risk models in the public sector are commonly developed using administrative data that is more complete for subpopulations that more greatly rely on public services. In the United States, for instance, information on health care utilization is routinely available to government agencies for individuals supported by Medicaid and Medicare, but not for the privately insured. Critiques of public sector algorithms have identified such ``differential feature under-reporting'' as a driver of disparities in algorithmic decision-making. Yet this form of data bias remains understudied from a technical viewpoint. While prior work has examined the fairness impacts of additive feature noise and features that are clearly marked as missing, little is known about the setting of data missingness absent indicators (i.e. differential feature under-reporting). 
    In this work, we study an analytically tractable model of differential feature under-reporting to characterizethe impact of under-report on algorithmic fairness.
    We demonstrate how standard missing data methods typically fail to mitigate bias in this setting, and propose a new set of augmented loss and imputation methods. 
    Our results show that, in real world data settings, under-reporting typically exacerbates disparities. The proposed solution methods show some success in mitigating disparities attributable to feature under-reporting.
\end{abstract}

\maketitle

\section{Introduction}
Regional and local governments around the world are using their increasingly digitized data systems to develop AI-driven decision-support technologies.
The hope is that these tools improve decision quality, reduce inefficiencies, eliminate fraud, and improve outcomes for their citizens \cite{engin2019algorithmic, levy2021algorithms}.  
Often, these technologies take the form of predictive risk models that are trained on administrative data to assess the likelihood that a case will go on to have poor outcomes.
Such models have been developed and deployed in criminal justice \cite{barnes2012classifying}, child welfare \cite{vaithianathan2017developing}, welfare fraud detection \cite{van2021digital}, federal tax audits \cite{houser2016use, black2022algorithmic}, homelessness services \cite{kithulgoda2022predictive}, health care \cite{mccarthy2015predictive}, and many other settings.  

Predictive risk models in the public sector have come under criticism over concerns that they are trained on biased data \cite{mayson2019bias, richardson2019dirty, chouldechova2018case,Berk2018}.
In this paper, we consider a specific form of bias.
We use the term `differential feature under-reporting' to describe the phenomenon whereby administrative data records are more complete for individuals who have more greatly relied on public services.
In the United States, for instance, administrative records often contain medical claims data for those who receive services through public insurance programs (Medicaid / Medicare), but lack information on physical, mental and behavioral healthcare utilization for the privately insured.
A lack of \emph{recorded} medical claims for an individual in this context is often indistinguishable from instances in which no medical claims have been made.
In her critique of the Allegheny Family Screening Tool (AFST) used in screening child maltreatment referrals, \citet{eubanks2018} writes, ``by relying on data that is only collected on families using public resources, the AFST unfairly targets low-income families for child welfare scrutiny.''

We provide a technical analysis of this problem.
First, we introduce a statistical model of data collection with differential feature under-reporting.
We then present theoretical results that characterize the impact of under-reporting on disparities in selection rates across groups.
We describe why standard missing data methods generally fail to mitigate unfairness and, instead, propose new methods based on augmented loss estimation and optimal prediction imputation that are tailored to the under-reporting setting. Lastly, we present empirical results on semi-synthetic and real world data. Our results show that, while in theory under-reporting can decrease disparities, in practice, under-reporting usually leads to increasing disparities and our proposed mitigation methods alleviate this increase.

\begin{figure}
\begin{subfigure}[b]{0.4\linewidth}
\centering
\begin{tikzpicture}
\footnotesize{
\node (g) [circle, draw, fill=gray!50!white, minimum size=0.6cm,pattern=horizontal lines, pattern color=gray!50!white] {$G$};
\node (z) [circle, draw, fill=white, right of=g, minimum size=0.6cm] {$Z$};
\node (y) [circle, draw, fill=gray!50!white, right of=z, minimum size=0.6cm] {$Y$};
\node (xi) [circle, draw, fill=white, below of=g, minimum size=0.6cm] {$\xi$};
\node (x) [circle, draw, fill=gray!50!white, right of=xi, minimum size=0.6cm] {$X$};
\node (haty) [circle, draw, fill=gray!50!white, right of=x, minimum size=0.6cm] {$\hat{Y}$};

\draw[dotted,->,black,fill=black] (g) -- (z);
\draw[->,black,fill=black] (z) -- (y);
\draw[->,black,fill=black] (xi) -- (x);
\draw[->,black,fill=black] (x) -- (haty);
\draw[->,black,fill=black] (g) -- (xi);
\draw[->,black,fill=black] (z) -- (x);
}
\end{tikzpicture}
\subcaption[]{General graph}
\end{subfigure}
\begin{subfigure}[b]{0.4\linewidth}
\centering
\footnotesize{
\begin{align*}
    &G\text{: High vs. low income group}\\
    &Z\text{: Number of doctor visits in the past year}\\
    &\xi\text{: Publicly insured ($\xi=1$) or privately insured ($\xi=0$)}\\
    &Y\text{: Health risk}
\end{align*}
\vspace{-3ex}
    }
\subcaption[]{Illustrative example}
\end{subfigure}
\caption{We study a prediction model on feature vectors with differential under-reporting $X$ where true outcomes $Y$ are a function of the latent `true' features $Z$. Missingness $\xi$ is influenced by group membership $G$. We consider both cases in which feature distributions vary by group membership and cases with $G\perp Z$.
In our setting, missingness indicators $\xi$ are unobserved and group membership $G$ is only used for model evaluation and not as a feature. The graph reflects the dependencies at prediction time.}
\label{fig:tikzgraph}
\end{figure}
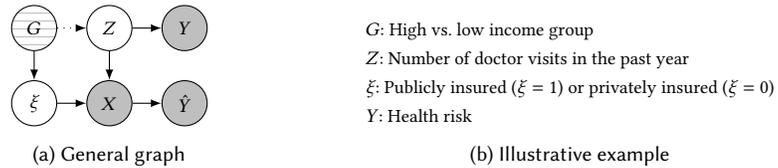

\section{Background and related work}
\label{sec:relatedWork}

\paragraph{Under-reporting vs. missingness}
The problem of differential feature under-reporting is illustrated in Figure~\ref{fig:tikzgraph}. An individual's risk prediction $\hat Y$ is formed based on observed administrative data features $X$ which are a mismeasured version of a ``true'' latent feature vector $Z$.  We assume that certain features in $Z$, such as demographic information, are correctly observed, whereas others, such as use of mental health services, are only correct for individuals who rely on publicly funded services.  
Problematically, we generally lack indicators on who is privately or publicly funded and for which services.
For indicators and count features, e.g. the number of episodes in inpatient mental health treatment in the past year, the mismeasured feature will simply show the value $0$ for individuals who received those services through privately funded mechanisms.
This means that when we observe $X_j=0$, we do not know whether $Z_j = 0$, or if $Z_j \neq 0$ and the feature has been mismeasured. In the graph, the \textit{unobserved} missingness indicators are denoted by $\xi$. This distinguishes the under-reporting setting from standard missingness, wherein $\xi$ is assumed to be fully observed.

Missing data literature distinguishes three types of mechanisms: (1) Missing Completely At Random (MCAR) where missing values are independent of both observed and unobserved data, (2) Missing At Random (MAR) where missingness depends on observed variables, and (3) Missing Not At Random (MNAR) where missing values depend on unobserved data \cite{RUBIN1976}.
In administrative data, records are more available for individuals who rely more greatly on public services which often correlates with demographic attributes excluded from modeling. This implies an MNAR setting.

\paragraph{Under-reporting in real-world applications}
The problem of feature under-reporting extends beyond the administrative data context.
In health applications, Electronic Health Record (EHR) data is often under-reported at different levels for different population sub-groups \citep{Zou23,Rajkomar2018,Gianfrancesco2018,Groenwold2020,Cismondi2013}.
Socioeconomically disadvantaged patients may be missing more diagnostic tests due to limited health care access \citep{ahmad2019challenge,Arpey2017-nh}.
Reliance on clinical decision support systems trained on EHR data could exacerbate already existing health care disparities \citep{Gianfrancesco2018,Char2018,Jeanselme_De-Arteaga_Zhang_Barrett_Tom_2022}.
Similarly, the extent of under-reporting often varies across domains (e.g. hospitals) which has been studied by \citet{zhou22}.
While \citet{zhou22} consider model adaptation when shifting to unlabeled target data with different level of under-reporting, we focus on a single domain with varying levels of under-reporting across groups and study fairness implications. 

\paragraph{Under-reporting in biomedical research}
In epidemiological surveys, social stigma can lead participants to provide false negative responses (e.g. true maternal smoking status $Z$ vs. reported status $X$) \cite{King2001,McKnight2007,Greenland2014,Sechidis2017}.
Researchers have proposed various methods to estimate association between $Z$ and outcome $Y$ by leveraging $X$ including correction factors for independence tests \cite{Sechidis2017,Bross1954-vr}, adjusted mutual information \cite{Sechidis2017}, odds-ratios \cite{Chu2006,Edwards2013-jl,DOSEMECI1990}, risk-ratios \cite{Rahardja2021,Brenner1994}, and full likelihood approaches \cite{Adams2019}.
These methods are typically limited to binary features and outcomes, and focus on inferring the relationship between $Y$ and some $Z$ rather than on prediction.
In the study of geographical disease counts, \cite{Bailey2005,deOliveira2017,Stoner_2019,Gelman2013-da} Bayesian methods have been used to make inference in the under-reporting setting of $X \leq Z$.
Such methods require a host of parametric and distributional assumptions as well as informative priors.
In single-cell RNA sequencing, under-report arises as `zero-inflation,' which refers to genes going undetected despite being expressed in a cell due to low levels of RNA.
Methods correcting for zero-inflation are often Bayesian and highly specialized for the single-cell RNA sequencing task.  Note that even if some of the above approaches were applicable in our setting to learn
a correctly specified model $f(z)=\E{Y\mid Z=z}$ from observations of $X$ and $Y$, it is unclear how to use such models \textit{for prediction} when only the under-reported features $X$ are available at prediction time.

\paragraph{Additive noise and fairness}

\begin{table}
\renewcommand{\arraystretch}{1.5}
\footnotesize{
\begin{tabular}{|P{1.1cm}| P{2.3cm} | P{2.6cm} |P{3.1cm} | P{2.8cm}|}
  \hline
  &\multicolumn{1}{c|}{\textbf{Complete data}}&\multicolumn{1}{c|}{\textbf{Additive noise}}  & \multicolumn{1}{c|}{\textbf{Missing with indicator}} &\multicolumn{1}{c|}{\textbf{Under-reporting}} \\
  \hline
  \textbf{Setting}&
  \multicolumn{1}{c|}{
  \renewcommand{\arraystretch}{1}
  \begin{tabular}{cccc}
  $g$ & $\bm{z_1}$ & $z_2$ & $y$\\
  \hline
  0 & \textbf{10} & 2 & 1\\
  0 & \textbf{7} & 1 & 0\\
  1 & \textbf{0} & 3 & 1\\
  \end{tabular}} &
  \multicolumn{1}{c|}{
  \renewcommand{\arraystretch}{1}
  \begin{tabular}{cccc}
  $g$ & $\bm{x_1}$ & $z_2$ & $y$\\
  \hline
  0 & \textbf{10.2} & 2 & 1\\
  0 & \textbf{6.5} & 1 & 0\\
  1 & \textbf{0.8} & 3 & 1\\
  \end{tabular}} &
  \multicolumn{1}{c|}{
  \renewcommand{\arraystretch}{1}
  \begin{tabular}{ccccc}
  $g$ & $\bm{x_1}$ & $\bm{r}$ &$z_2$ & $y$\\
  \hline
  0 & \textbf{10} & \textbf{1}& 2 & 1\\
  0 & $\bm{m}$ & \textbf{0}& 1 & 0\\
  1 & $\bm{m}$ & \textbf{0}& 3 & 1\\
  \end{tabular}} &
  \multicolumn{1}{c|}{
  \renewcommand{\arraystretch}{1}
  \begin{tabular}{cccc}
  $g$ & $\bm{x_1}$ & $z_2$ & $y$\\
  \hline
  0 & \textbf{10} & 2 & 1\\
  0 & $\bm{m}$ & 1 & 0\\
  1 & $\bm{m}$ & 3 & 1\\
  \end{tabular}} \\
  &
  Features fully observed
  &
  Feature values with added random noise~$\varepsilon$
  &
  Some feature values take default value $m$; $r$ indicates which values are observed
  &
  Some feature values take default value $m$; No indicators for missingness
  \\
  \hline
  \textbf{Previous fairness work}&No feature mismeasurement
  & 
\citet{khani2020feature}, \citet{Phelps1972}, \citet{Aigner1977}, \citet{chen2018my}&
  \citet{Zhang21Neurips}, \citet{Wang2021}, \citet{Jeanselme_De-Arteaga_Zhang_Barrett_Tom_2022}, \citet{Fernando2021}, \citet{Fricke2020}, \citet{ahmad2019challenge}
  & \textbf{This work}, \citet{eubanks2018}\\
  \hline
\end{tabular}}
\caption{Different types of feature mismeasurement and previous work addressing fairness implications. In the data examples, mismeasured features are denoted with $x$ while correctly observed features are denoted by $z$. Column $g$ encodes group membership.}
\label{fig:table}
\vspace{-7ex}
\end{table}

The algorithmic fairness literature has studied various types of feature mismeasurement as summarized in Table~\ref{fig:table}.
A commonly studied setting is additive feature noise where, instead of a feature $z_1$, we observe a noisy version $x_1=z_1+\varepsilon$.
The random noise $\varepsilon$ is often assumed to be zero-mean, of small variance, and independent of other variables. This implies that, while some of the information in the feature is diluted, large portions of the encoded information remains intact.
\citet{khani2020feature} show that adding the same amount of feature noise to a group-blind model can introduce statistical loss discrepancy.
This is in line with earlier observations from the statistical discrimination literature \cite{Phelps1972,Aigner1977}.
\citet{chen2018my} propose data collection strategies targeted at decreasing discrepancy and come to the conclusion that overcoming differential noise across protected groups may require collection of additional data.
In contrast to additive noise, under-reporting removes all information from impacted feature entries and typically biases the feature mean.
Some works \citep[e.g.][]{khani2020feature} suggest that feature missingness can be modeled as a special case of additive noise by selecting noise terms with very high variance. However, this only covers a special case of under-reporting in which feature entries are missing for \emph{all} observations.

\paragraph{Missing data methods and fairness}
\label{sec:missing_data_methods}

Feature missingness has been studied in the statistical literature for several decades \citep[e.g.][]{RUBIN1976}.
This line of work generally assumes that we observe missingness indicators or, equivalently, missing values are clearly marked (e.g. NaN).
Various methods with different fairness implications have been proposed (Table~\ref{fig:table}).
\vspace{-0.6cm}
\begin{enumerate}
    \item[(1)] \emph{Complete case analysis and reweighing.}
    In some cases, it may be desirable to remove incomplete rows which can lead to significant biases \citep{RUBIN1976}.
    Various reweighing procedures have been proposed to deal with this problem.
    \citet{Zhang21Neurips} suggest learning only from complete observations while employing an importance sampling procedure.
    \citet{Wang2021} suggest that reweighing and resampling methods in the context of categorical data can lead to considerable fairness improvements over learning with missing data directly.
    \item[(2)] \emph{Imputation.}
    \citet{Jeanselme_De-Arteaga_Zhang_Barrett_Tom_2022} compare different imputation strategies under clinical presence and find that there is no imputation strategy that reliably outperforms other imputation methods in terms of fairness. 
    \citet{Fernando2021} and \citet{Fricke2020} compare feature imputation to complete case analysis and find that rows with missing values can contribute to fairer outcomes via observed columns.
\end{enumerate}
\vspace{-0.5em}
In contrast to the feature missingness setting, we do not observe indicators for missing entries in this work. Despite this difficulty, we experiment with row omission and imputation in Section~\ref{sec:experiments}.

\section{Problem setup}

\paragraph{Setting}
We study the effect of feature under-reporting on algorithmic fairness through the lens of the regression setting displayed in Figure~\ref{fig:tikzgraph}.
Assume latent feature vectors $z\in\R^d$ and group information $g\in\{0,1\}$.
We assume a noiseless regression setting in which the outcome $y$ is a linear function of $z$, i.e. $y=\alpha+\beta^Tz$ with $\alpha\in\R$, $\beta\in\R_{\neq 0}^d$.
Instead of the true features $z$, we observe a mismeasured vector $x$ in which entries default to 0 with group-dependent probabilities. That is, we set $x = z\odot\xi^{g}$ where $\odot$ denotes element-wise multiplication, $\xi^g\sim\text{Bern}(m^g)$, and $(1-m^0),(1-m^1)\in(0,1]^d$ are under-reporting rates in the two groups.
More formally, we have a group random variable $G\sim\text{Bern}(r)$ and a random feature vector $Z$. The vector of under-reported features $X$ can be written as $X=Z\odot \xi$ where $\xi = G\xi^1 + (1-G)\xi^0$.
This setting allows for different dependence structures depending on whether we assume $G\perp Z$.

\paragraph{Two-step bias}
Differential under-reporting introduces bias in two ways: (1) Under-reporting in training data influences estimation of the prediction model \emph{(estimation step)}, and (2) input data with under-reporting leads to biased predictions at test time \emph{(prediction step)}. 
It is generally not sufficient to recover the true model parameters $\alpha,\beta$ as only biased features are available at prediction time. In fact, our experiments in Section~\ref{sec:experimentresults} demonstrate that using true parameters for prediction can lead to worse fairness outcomes than using a model estimated with biased data.

\paragraph{Thresholded prediction}
We assume a thresholded prediction setting reminiscent of predictive risk modeling in the public sector.
A predictor $f$ is fit on $(X,Y)$ to produce predictions $\hat{Y}=f(X)=f(Z\odot\xi^G)$. 
We consider group-wise shares of predictions above a given threshold $\tilde{y}$: $P(\hat Y \ge \tilde y \mid G = g)$ which we refer to as \textit{selection rates} at threshold $\tilde{y}$. 
This implicitly assumes a setting in which the \emph{highest} risk individuals are selected (e.g. child welfare screenings, fraud detection, federal tax audits). However, it is straightforward to reverse the analysis for scenarios in which \emph{low} risk leads to selection (e.g. bail decisions in criminal risk assessment).
In addition, we assume that being selected is \emph{undesirable}.
Crucially, this assumption is only made to simplify interpretation and we could easily consider the opposite case.

\paragraph{Excess selection rates}

We assume that the threshold on predictions $\hat{Y}$ is set to achieve a desired overall \textit{selection rate} ${P}(\hat Y \ge \tilde y) = C\in[0,1]$. 
Given the cumulative distribution function of predictions $F_{\hat{Y}}$, the percentile threshold $C$ implies an absolute threshold $\tilde{y}=F_{\hat{Y}}^{-1}(1-C)$ such that the selection rate for group $g$ can be written as $P(\hat{Y} \geq \tilde{y}\mid G=g)$.
In order to isolate the effect of under-reporting, we need to account for a ground truth difference in selection rates.
Let $\hat{Y}_X$ denote the predictions of a model trained on $(X,Y)$ and $\hat{Y}_Z$ the predictions of a model trained on $(Z,Y)$.
When distributions of $\hat{Y}_Z$ and $\hat{Y}_X$ differ, the predictions imply different thresholds $y'=F_{\hat{Y}_Z}^{-1}(1-C)$ and $\tilde{y}=F_{\hat{Y}_X}^{-1}(1-C)$.
With this notation, we define a metric for impact of differential feature under-reporting on disparities in selection rates.
\vspace{-0.5em}
\begin{definition}[Excess selection rate due to under-reporting]
\label{def:excessselectionrate}
The \emph{excess selection rate} for group $g\in\{0,1\}$ at overall selection rate $C\in[0,1]$  
\vspace{-1.5em}
\begin{align*}
    \Delta(g,C) := P(\hat{Y}_X \geq \tilde{y}\mid G=g) - P(\hat{Y}_Z\geq y'\mid G=g),
\end{align*}
is the difference in selection rates when ranking according to a model trained on $X$ compared to a model trained on $Z$.
We say that group $g$ is \emph{over-selected} at level $C$ if $\Delta(g,C) > 0$.
If $\Delta(g,C)<0$, we say that $g$ is \emph{under-selected}. 
\label{def:overselection}
\end{definition}
\vspace{-0.5em}
In principle, we could directly consider a ``difference in difference'': the difference in  selection rates between groups $g=0,1$ when selection occurs according to the model $\hat{Y}_X$ versus the unbiased predictions $\hat{Y}_Z$.
However, since we select a fixed share of the population $C$, an increase of the selection rate of group $g$ when moving from $\hat{Y}_Z$ to $\hat{Y}_X$ already implies a decrease for group $1-g$.
 It is generally difficult to argue about the excess selection rate $\Delta(g,C)$ analytically.
Even in a simple setting with group-dependent Gaussian features $Z \mid G\sim\mathcal{N}\left(\mu_G,\Sigma_G\right)$, there is no closed-form expression for the quantile $y'=F_{\hat{Y}_Z}^{-1}(1-C)$ and determining the sign of $\Delta(g,C)$ requires analysis of a difference in cdfs which is often intractable.
Instead, we simplify the setting and assume that $Z$ follows the same distribution across groups.
In this case, the selection rates on the true outcome $Y$ are the same in both groups at every threshold, and we can simplify.
\vspace{-0.5em}
\begin{definition}[Excess selection rate due to under-reporting, independent case]
\label{def:overselectionindependent}
    If $Z\perp G$, we say that group $g\in\{0,1\}$ is \emph{over-selected} at threshold $C\in[0,1]$ if
    $
        P(\hat{Y}_X\geq \tilde{y}\mid G=g) > P(\hat{Y}_X\geq \tilde{y}\mid G=1-g),
    $
    and \emph{under-selected} if the inequality is reversed.
\end{definition}
\vspace{-0.5em}
While the majority of our theoretical derivations assume the special case of $Z\perp G$, the empirical portion of this work explores the impact of feature under-reporting in the more general setting.
To clarify which assumptions are sufficient for which finding in the paper, we supply a summary table in Appendix~\ref{app:summary_assumptions}, alongside descriptions in the main text.

\section{Differential feature under-reporting in linear regression}
\label{sec:linreg}

In this section, we examine the bias that differential feature under-reporting introduces into regression parameter estimates. We consider a setting in which true outcomes are a linear function of latent features, i.e. $Y=\alpha + \beta^TZ$, which implies that a linear model with access to the true $Z$ recovers the true outcomes $Y$. Dropping subscripts, we write $Y=\hat{Y}_Z$ and $\hat{Y}=\hat{Y}_X$.
Note that this section focuses on population-level regression.

\paragraph{Estimates and attenuation bias}
\label{sec:generalestimates}
Feature mismeasurement in the form of under-reporting leads to inconsistent parameter estimates in linear regression. When fitting a linear model on $(X,Y)$, the least squares estimates become
\begin{alignat}{2}
    \hat{\beta} & = \Sigma_{X}^{-1}\Sigma_{XZ}\beta, & \quad \hat{\alpha} & = \alpha + \E{Z}^T\beta - \E{X}^T\hat{\beta},
    \label{eq:paramestimates}
\end{alignat}
where $\Sigma_{XZ}$ denotes the covariance matrix between $X$ and $Z$ and we write $\Sigma_{X}$ for $\Sigma_{XX}$.
At first glace, this solution resembles the regression estimates in the more commonly studied additive feature noise case. Assuming $X'=Z+U$ where $U$ is independent zero-mean feature noise, we obtain $\hat{\beta}=\Sigma_{X'}^{-1}\Sigma_{X'Z}\beta=(\Sigma_{Z}+\Sigma_{U})^{-1}\Sigma_{Z}\beta$. The factor $\lambda = (\Sigma_{Z}+\Sigma_{U})^{-1}\Sigma_{Z}$ is commonly interpreted as a noise-to-signal ratio and, if $Z$ is one-dimensional, we know that $| \hat{\beta}|=\lambda|\beta|<|\beta|$ which is generally referred to as attenuation bias \citep[e.g.][]{Hausman01,fuller1987}.
In the under-reporting setting, $\Sigma_{X}$ does not easily separate into terms depending on only the feature or only the mismeasurement. However, in the special case of one-dimensional $Z$ and $Z\perp G$, we can still show that the parameter $\hat{\beta}$ is biased towards zero.
\vspace{-0.5em}
\begin{lemma}[Attenuation bias]
\label{lemma:attenutationbiasonedim}
Assume the feature $Z$ is one-dimensional and has the same distribution across groups. Then, the least squares regression of $Y$ on the mismeasured feature $X$ yields an estimated slope $\hat{\beta}$ with $|\hat{\beta}| \leq |\beta|$.
\end{lemma}
\vspace{-0.5em}

\paragraph{The $d$-dimensional case}
\label{sec:estimationddimcase}
Real-world prediction settings typically include multiple, often correlated, features.
Assume the feature vector $Z$ is $d$-dimensional, and under-reporting only occurs in the first feature, $Z_1$. This means that $X$ coincides with $Z$ in all but the first entry which is computed as $X_1=Z_1\xi_1$ where $\xi_1=G\xi_1^1+(1-G)\xi_1^0$ and $\xi_1^1\sim\text{Bern}(m^0_1)$, $\xi_1^0\sim\text{Bern}(m^1_1)$.
We further assume that features $Z_2,\ldots,Z_d$ are uncorrelated and the feature dependence structure is characterized entirely by the correlations between $Z_1$ and $Z_{[2:d]}$.
This assumption is without loss of generality because we can always apply orthogonalization to the features.
We explicitly exclude cases in which $Z_1$ is a perfect linear combination of other features to avoid problems of multicollinearity and assume $\V{Z_k}>0$ for all $k\in[2:d]$.
Given this setting, Proposition~\ref{prop:params} provides a closed form representation for the parameter estimates $\hat{\beta}$. The bias in these estimates can be conceptualized as a generalization of omitted variable bias \cite{Angrist2008-oo} which is discussed Appendix~\ref{app:omitted_variable_bias}.
We now assume that $Z\perp G$ which implies that under-reporting is independent of the value that is under-reported. This resembles the assumptions made in previous work on the impact of additive feature noise on fairness \citep[e.g.][]{khani2020feature,Phelps1972,Aigner1977}, and allows us to gain analytical insights that would otherwise remain intractable.
First, we examine the behavior of the parameter estimate for the feature with under-reporting, i.e. $\hat{\beta}_1$.
\vspace{-0.5em}
\begin{proposition}[Properties of $\hat{\beta}_1$]
\label{prop:propertieshatbeta1}
If $Z\perp G$, the parameter estimate $\hat{\beta}_1$ has the following properties.
\begin{enumerate}
    \item \textbf{Sign invariance:} $\hat{\beta}_1$ has the same sign as $\beta_1$.
    \item \textbf{Attenuation bias: }$|\hat{\beta}_1|\leq | \beta_1|$.
    \item \textbf{Attenuation bias increasing with under-reporting:} 
    If under-reporting $1-m_1^g$ is increasing for one (or both) groups $g\in\{0,1\}$, ceteris paribus, the magnitude of the parameter estimate, $|\hat{\beta}_1|$, is decreasing.
\end{enumerate}
\end{proposition}
\vspace{-0.5em}
This finding shows that feature under-reporting leads to attenuation bias in the respective parameter estimate even when other correlated and fully observed features are available. The attenuation bias gets more pronounced with more under-reporting. Next, we turn towards the estimates for the fully observed features $\hat{\beta}_k$ for $k\in[2:k]$.
\vspace{-0.5em}
\begin{proposition}[Properties of $\hat{\beta}_k$]
\label{prop:propertieshatbetak}
If $Z\perp G$, the parameter estimates $\hat{\beta}_k$ for $k\in[2:d]$ have the following properties.
\begin{enumerate}
    \item \textbf{Correlation bias:} 
    If $\hat{\beta}_k\neq\beta_k$, then $\rho(Z_1,Z_k)>0$.
    \item \textbf{Shifting weight:} If under-reporting $1-m_1^g$ is is increasing for one (or both) groups $g\in\{0,1\}$, ceteris paribus, $\hat{\beta}_k$ is increasing if $\text{sign}\left(\beta_1\Cov{Z_1}{Z_k}\right) = +1$, and decreasing if $\text{sign}\left(\beta_1\Cov{Z_1}{Z_k}\right) = -1$.
\end{enumerate}
\end{proposition}
\vspace{-0.5em}
In line with general intuition, under-reporting in $Z_1$ has no effect on the parameter estimate $\hat{\beta}_k$ if features $Z_k$ and $Z_1$ are uncorrelated. If the features are correlated, the direction of the under-reporting effect on the parameter estimate depends on the signs of $\beta_1$ and $\Cov{Z_1}{Z_k}$. Note that this is independent of the value and sign of $\beta_k$.

\paragraph{Take-away}
Proposition~\ref{prop:propertieshatbeta1} and \ref{prop:propertieshatbetak} tell a compelling story about the effect of under-reporting on parameter estimates in the studied setting. As more feature values default, the regression model places less weight on the mismeasured feature and instead shifts weight to fully observed features with non-zero correlation. This can lead to increasing or decreasing parameter estimates.
Analytically, there are cases in which `shifting weight' means that the magnitude of a parameter estimate, $| \hat{\beta}_k|$, is decreasing, which may appear counterintuitive. For example, consider a setting in which both $Z_1$ and $Z_k$ have positive true parameters but negative correlation, i.e. $\beta_1,\beta_k>0$ and $\Cov{Z_1}{Z_k}<0$.
In practice, this could occur when there are several mutually exclusive paths to the same outcome. For example, consider prediction of general health risk scores with features including both the number of pediatrician visits in the last year and the number of internist visits. Presumably, these features are negatively correlated because they are relevant for two mutually exclusive parts of the population, i.e. children and adults, but in both columns larger values can be indicative of a high general health risk.

\section{Impact on selection rate disparity}
\label{sec:xx}

\paragraph{Selection rate disparity in Gaussian setting}
\label{sec:propwhobenefits}
We study the effect of differential feature under-reporting on selection rate disparities in linear regression.
Similar to our previous discussion, we assume a $d$-dimensional feature setting in which only the first feature is subject to under-reporting and $Z\perp G$.
We further assume that features are jointly Gaussian, i.e. $Z\sim\mathcal{N}\left(\mu,\Sigma\right)$ where $\mu\in\R^d$ and the covariance matrix $\Sigma\in\R^{d\times d}$ is positive definite. This has the benefit that predictions $\hat{Y}=\hat{Y}_X$ follow a Gaussian mixture distribution which allows us to directly analyze group selection rates.
If under-reporting rates are the same across groups, there is no selection rate disparity as both groups have the same feature distributions. 
If the under-reporting rates vary between groups, we observe the following. 
\vspace{-0.5em}
\begin{proposition}
\label{prop:whobenefits_short}
For a sufficiently high threshold $\tilde{y}$, the group with more under-reporting is
over-selected if
$$
    \V{\hat{\alpha}+\hat{\beta}_{[2:d]}Z_{[2:d]}}>\V{\hat{\alpha}+\hat{\beta}Z}
$$
(Case 1), or under-selected if the inequality is reversed (Case 2).
For low thresholds, the cases are reversed.
\end{proposition}
\vspace{-0.5em}
We refer to Proposition~\ref{prop:whobenefits} for an expanded version of this finding including a discussion of sufficiently high thresholds.
Proposition~\ref{prop:whobenefits_short} shows that over-selection primarily depends on variance in predictions. When cutting off at a high threshold, the group with more feature under-reporting is over-selected if the variance in predictions for examples with defaulted feature exceeds the prediction variance for fully observed examples (Case 1). It is under-selected if the variance of predictions is larger for the examples with fully observed features (Case 2).
Intuitively, at high thresholds, information deficiency in a group always leads to under-selection because it prompts the group's risk distribution to concentrate more closely around its mean moving more mass below the threshold. 
We find that, analytically, outcome disparities can go into either direction and sometimes groups with information deficiency are over-selected.
While our findings suggest that this is mostly a question of variance in predictions, this is likely only part of the story in settings with group-dependent feature distributions. 
We study more general settings empirically in Section~\ref{sec:experiments}.

\paragraph{Combining parameter estimation and prediction steps}
Feature under-reporting introduces bias both at estimation and prediction time. 
In the following, we combine our previous findings to examine the conditions under which under-reporting leads to over-selection and under-selection.
As before, we assume a $d$-dimensional feature setting in which only the first feature $Z_1$ is impacted by under-reporting. Features are jointly Gaussian, and we further assume that $Z_2,\ldots,Z_d$ are uncorrelated.
\vspace{-0.5em}
\begin{corollary}
\label{cor:maincorollary}
Given the first and second moments of $Z_1$, the expected share of observed values $\E{\xi_1}$, and the fraction of variance in $Z_1$ that is explained by the remaining features $S^2=\sum_{i=2}^d\rho(Z_1,Z_i)^2$,
there exists a positive constant $c=c(\E{Z_1},\V{Z_1},\E{\xi_1},S^2)$ such that, at high thresholds, the group with more under-reporting is over-selected if
$$
    \frac{1}{\beta_1}\sum_{j=2}^d\beta_j\Cov{Z_1}{Z_j} < -c,
$$
(Case 1), and under-selected if the inequality is reversed (Case 2).
\end{corollary}
\vspace{-0.5em}
Thresholds are considered high if they exceed the turning point defined in Proposition~\ref{prop:whobenefits}.
The corollary shows that over-selection due to feature under-reporting depends on the signs and magnitudes of the true parameters $\beta$ and the covariances between features. If
\mbox{$
\text{sign}\left(\frac{1}{\beta_1}\sum_{j=2}^d\beta_j\Cov{Z_1}{Z_j}\right) = 1,
$}
e.g. if all true parameters and covariances are non-negative, the group with more missingness will always
be under-selected at high thresholds.
If the sign is negative, the group with more under-reporting is over-selected if the covariance-weighted sum of true parameters 
is sufficiently large in absolute value. Otherwise, feature under-reporting still leads to under-selection.

\section{Solution approaches}
\label{sec:solution_approaches}
Sections~\ref{sec:linreg} and \ref{sec:xx} show how ignoring differential feature under-reporting can lead to disparities in selection rates across groups.
In the following, we explore how conventional missing data methods can be adapted to the under-reporting setting. We then propose a new set of methods that is specifically tailored to this setting by separating the problem into two steps---estimation and prediction. For the estimation step, we provide a method that recovers the ground truth data generating model from observed data. For the prediction step, we derive optimal group-dependent imputation values.
As before, we assume under-reporting occurs only in the first feature which is observed as $X_1=Z_1\xi_1$.

\paragraph{Standard missing data methods}
\label{sec:standard_solution_approaches} 

Existing missing data methods typically assume that defaulted values are clearly marked which is not the case in the under-reporting setting. We explore adaptations of several methods.
\begin{enumerate}
    \item [(1)] \emph{Feature omission.}
    Discarding the mismeasured feature vector $X_1$ doesn't require missingness indicators and mitigates the bias introduced by under-reporting. 
    However, this approach may decrease model performance significantly, and may itself introduce bias.
    When assuming a linear ground truth, feature omission leads to omitted variable bias in parameter estimates $\hat{\beta}_{[2:d]}$, which has been studied previously in the econometrics literature \cite{Angrist2008-oo} (also see Appendix~\ref{app:omitted_variable_bias}).
    \item [(2)] \emph{Multiple imputation.}
    Multiple imputation draws plausible feature values while retaining variability.
    Since we do not observe indicators for missingness, we experiment with imputing all 0-entries in $X_1$ which includes correctly observed 0s.
    In each imputation run, we estimate the posterior $P(Z_1\mid Z_2,\ldots,Z_d)$ on data rows with $X_1\neq 0$, impute 0-entries, and train a prediction model for $Y$.
    At prediction time, we average the imputations and predictions over models to obtain a single prediction $\hat{Y}$.
    While this procedure successfully alleviates bias in some standard feature missingness settings, it is not a priori clear how well the method works with under-reporting.
\item [(3)] \emph{Row omission.}
    Omission of rows with missing feature entries provides a convenient complete case analysis.
    Since true and false 0-entries are indistinguishable in our setting, we experiment with discarding all rows with $X_1=0$. If there is no model misspecification, e.g. in the linear case with $f(Z)=\alpha + \beta^TZ$ where we train a linear model on observed features, training on only complete rows is guaranteed to asymptotically retrieve the true parameters $\hat{\alpha}=\alpha$ and $\hat{\beta}=\beta$ if $Z_1$ is not binary. Even with access to the ground truth model, under-reporting introduces bias via the prediction step and $\hat{Y}=f(X)$ may not be the most accurate (or fairest) prediction.
\end{enumerate}

\label{sec:new_solution_approaches}

\paragraph{Model estimation with augmented loss}
Without model misspecification, row omission can recover the ground truth model $Y=f(Z)$.
In practice, models are usually misspecified and discarding rows can significantly decrease performance.
Instead, we propose an augmented loss function to recover the ground truth model.
This proxy loss uses observed features $X$ to provide an unbiased estimate of the loss of a model $f$ on latent features $Z$.
Similar approaches have previously been used in the label noise setting \citep{Natarajan13,Nemirovski2009}.
Assume $Z\in\R^d$ has support $\mathcal{Z}$ and $y\in\R$ has support $\mathcal{Y}$. Let $\mathcal{F}:\mathcal{Z}\to\R$ be a class of real-valued functions and $l:\mathcal{F}\times \mathcal{Z}\times \mathcal{Y}\to \R$ be a bounded loss function.
We assume $Z\perp G$ and denote the rate of observed values as $m_1:=\E{\xi_1}=rm_1^1 + (1-r)m_1^0$.
\vspace{-0.5em}
\begin{lemma}[Augmented loss]
    \label{lem:augmented_loss}
    Assume fixed $f\in\mathcal{F},z\in\mathcal{Z}$, $y\in\mathcal{Y}$ and $X\in\R^d$ defined by $X_1=Z_1\xi_1$ and $X_{[2:d]}=z_{[2:d]}$.
    Define
    \vspace{-1em}
    \begin{align*}
        \tilde{l}(f,X,y) := \frac{1}{m_1}l(f,X,y) - \frac{1-m_1}{m_1}l(f,[0,X_{[2:d]}]^T,y).
    \end{align*}
    If $Z\bot G$, we have that $\mathds{E}_{\xi_1}\left[\tilde{l}(f,X,y)\right]=l(f,z,y)$.
\end{lemma}
\vspace{-0.5em}
The fact that the augmented loss is unbiased with respect to under-reporting noise implies that a prediction model on observed data estimated with augmented loss, i.e. 
$
    \hat{f} = \arg\min_{f\in\mathcal{F}} \mathds{E}_{(X,Y)}\left[\tilde{l}(f,X,Y)\right],
$
asymptotically recovers the Bayes optimal model on the true features $Z$. If $Y=\alpha + \beta^TZ$, $\mathcal{F}$ is the class of linear functions $f:\R^d\to \R$ and $l(f,z,y)=(f(z)-y)^2$ denotes squared error loss, the true parameters $\hat{\alpha}=\alpha$ and $\hat{\beta}=\beta$ are retrieved. Note that squared error loss is not bounded and estimating $\hat{f}$ requires the additional constraint $\tilde{l}(f,X,Y)\geq 0$.
Lemma~\ref{lem:augmented_loss} operates in a group-agnostic setting with $Z\bot G$.
Lemma~\ref{lem:group_dep_augmented_loss} provides a more general group-dependent version of the finding.

\paragraph{Optimal prediction imputation value}
Assume we are in the linear case with $Y=\alpha + \beta^TZ$ and we have access to the true parameters $\alpha$ and $\beta$, e.g. obtained via augmented loss.
What is the best possible prediction for an example of the form $x = [0,z_2,\ldots,z_d]$? Since $x_1=0$ could mean $z_1=0$ or the entry is missing, it is intuitive that $\hat{y}=\alpha + \beta^T x$ does not minimize expected prediction error. Instead, we derive the optimal fixed prediction imputation value $x_1'^\ast$.
\vspace{-0.5em}
\begin{lemma}[Optimal prediction imputation value]
    \label{lem:optimal_prediction_imp_value}
    Assume $Z\bot G$, $f(Z)=\alpha+\beta^TZ$ is the ground truth model and $X$ the random vector of observed features. We set
    \vspace{-1.5em}
    \begin{align*}
        X' = \begin{cases}
            X \text{ if } X_1\neq 0,\\
            [x_1',X_{[2:d]}] \text{if } X_1=0,
        \end{cases}
    \end{align*}
    where $x_1'$ is fixed.
    Then,
    $
        x_1'^\ast := \arg\min_{x_1'} \mathds{E}_X[(f(X')-Y)^2] = \E{Z_1\mid X_1=0}
    $
    is the optimal prediction imputation value.
\end{lemma}
\vspace{-0.5em}
The Lemma shows that the loss-minimizing constant imputation value is the conditional mean $Z_1$ given the observed value is 0. This implies that, in alignment with earlier intuition, directly predicting with the observed $x=[0,z_2,\ldots,z_d]$ is sub-optimal in the under-reporting setting.
The optimal value $x_1'^\ast$ in the setting of Lemma~\ref{lem:optimal_prediction_imp_value} can be written as
\begin{align*}
    \E{Z_1\mid X_1 = 0} = \frac{\frac{1}{m_1}\E{X_1}-P(X_1\neq 0)\E{X_1\mid X_1\neq 0}}{P(X_1=0)},
\end{align*}
which can be estimated directly from observed data if the under-reporting rate $1-m$ is known. If feature distributions vary across groups, group-dependent optimal prediction imputation values can be derived as described in Lemma~\ref{lem:group_optimam_prediction_imp_value}.

\paragraph{Under-reporting rate estimation}
\label{sec:estimate_m}
Both augmented loss estimation and optimal prediction imputation require access to the reporting rate, $m$, which is typically unknown.
In some cases, it may be possible to obtain supplementary data that can be used to estimate $m$.
For example, in administrative data with under-reported health features for privately insured individuals, an external private insurance health claims dataset could be used to estimate the expected rate of true 0's. 
In most settings, estimation of under-reporting rates needs to rely directly on the observed data.
Assume we have access to a dataset $V=\{(x,y)_{i=1}^n\}$. We split $V$ into a training portion $V_{\text{train}}$ and evaluation portion $V_{\text{eval}}$. Let $P_{\text{eval}}$ denote the subset of examples from $V_{\text{eval}}$ for which $x_1\neq 0$.
We draw on the literature on Positive and Unlabeled (PU) learning \cite{Elkan2008} and estimate under-reporting rates as follows. First, we fit a model $h$ on $V_{\text{train}}$ to estimate $P(X_1\neq 0\mid X_{[2:d]},Y)$.
Second, we evaluate $h$ on $P_{\text{eval}}$. The estimator for the share of observed values $m$ is given by
$
\hat{m} = \frac{1}{\mid P_{\text{eval}}\mid} \sum_{(x,y)\in P_{\text{eval}}} h(x_{[2:d]},y).
$
The estimation procedure assumes that under-reporting occurs completely at random. Our experimental setting assumes under-reporting completely at random within groups and thus $m_0$ and $m_1$ can be estimated with the described procedure by restricting $V$ to examples from the respective group.
For more details on the estimation procedure, we refer to Appendix~\ref{app:missingness_estimation}.

\section{Experiments}
\label{sec:experiments}

\subsection{Publicly available datasets}

\paragraph{Data}
Both COMPAS data \cite{compas} and German credit data \cite{German} are widely used across the algorithmic fairness literature.
The American Community Survey (ACS) Income dataset is comprised of 2018 census data from California \citep{folktables}.
Datasets vary in size, number of features, and prediction tasks as shown in Table~\ref{table:datasets}. We conduct all experiments with both gender and race as group columns if available. All models are group-blind and race information is never included as predictive feature.
Results for the German credit dataset are discussed in Appendix~\ref{app:german}.

\label{sec:setup}
\paragraph{Semi-synthetic outcomes}
Since all of the prediction tasks are binary classification, we opt to generate semi-synthetic regression labels for our experiments.
We first fit a logistic regression model to the entire dataset and extract the fitted probabilities. For the ACS Income data, the values are rescaled to center around the \$50,000 income threshold.
Next, we fit a linear regression model using the same features and the predicted probabilities as outputs. The fitted values from this linear model are chosen as the new ``true'' labels for our experiment.
This outcome augmentation procedure allows us to generate artificial settings with a truly linear ground truth similar to the settings studied in Sections~\ref{sec:linreg} and \ref{sec:propwhobenefits} while leaving realistic covariance structures intact.
We further experiment with controlling the $R^2$ of the true linear model by adding additional noise to outcomes and report fairness implications in Appendix~\ref{app:noise}.

\paragraph{Experiment stratification}
Regression models are trained to predict semi-synthetic outcomes based on the features of the respective dataset. We select the top $C$ share of the predictions as high risk and evaluate excess selection rates to assess the fairness impact of under-reporting.
Artificial under-reporting is added to one feature column at a time and we repeat the experiments for each outcome column, group column, and under-reporting rate. Under-reporting rates range from 0-90\% in 10 percentage point increments, and we add under-reporting to only one group at a time (e.g., we set 10\% of a feature in the male group to 0 while leaving the features of the female group unchanged). Only numeric features are considered for under-reporting since, in administrative data, binary features are often categorical dummies or thresholded versions of continuous count features. 
All models are trained with 80\% of the datasets while withholding 20\% for testing.
We experiment with various solution approaches as described in Section~\ref{sec:solution_approaches}. This includes our proposed methods of group-dependent augmented loss estimation and group-dependent optimal prediction imputation.

\subsection{County-level birth data}
\paragraph{Data}
We present an analysis of a private administrative dataset we obtained from a county in the US.
The dataset contains information on newborn children and their families including demographics, child protective services history, birth record data, and mental and behavioral health information for those who used publicly funded services.
We set up a prediction task that attempts to mimic the analysis described in the Hello Baby model methodology report from Allegheny County \cite{hello-baby}.
The Hello Baby model was developed to predict which families are at greatest risk of having their child removed by Child Protective Services (CPS) during their first three years of life, and is used to prioritize families with newborn children for opt-in, voluntary supportive services.
Using our data, we train a similar model, and explore the effect of adding additional under-reporting to the behavioral and mental health data fields.

\paragraph{Experiment setup}
We use the birth dataset with its original prediction outcomes to showcase a realistic example of the effect of under-reporting.
As before, the data is separated into 80\% for training and 20\% for testing.
We fit separate logistic regression models on three datasets. (1) Data as observed. (2) Data with behavioral health features set to 0 for privately insured individuals, i.e. mothers that are not insured through Medicaid. (3) Data without behavioral health features. For illustration, results are stratified by whether individuals are covered by Medicaid, and by whether the mother's race is recorded as Black. Medicaid coverage and race are not used as features in any of the models.

\section{Results}
\label{sec:experimentresults}

\begin{figure}
    \centering
    \begin{subfigure}[b]{0.49\linewidth}
        \includegraphics[scale=0.4]{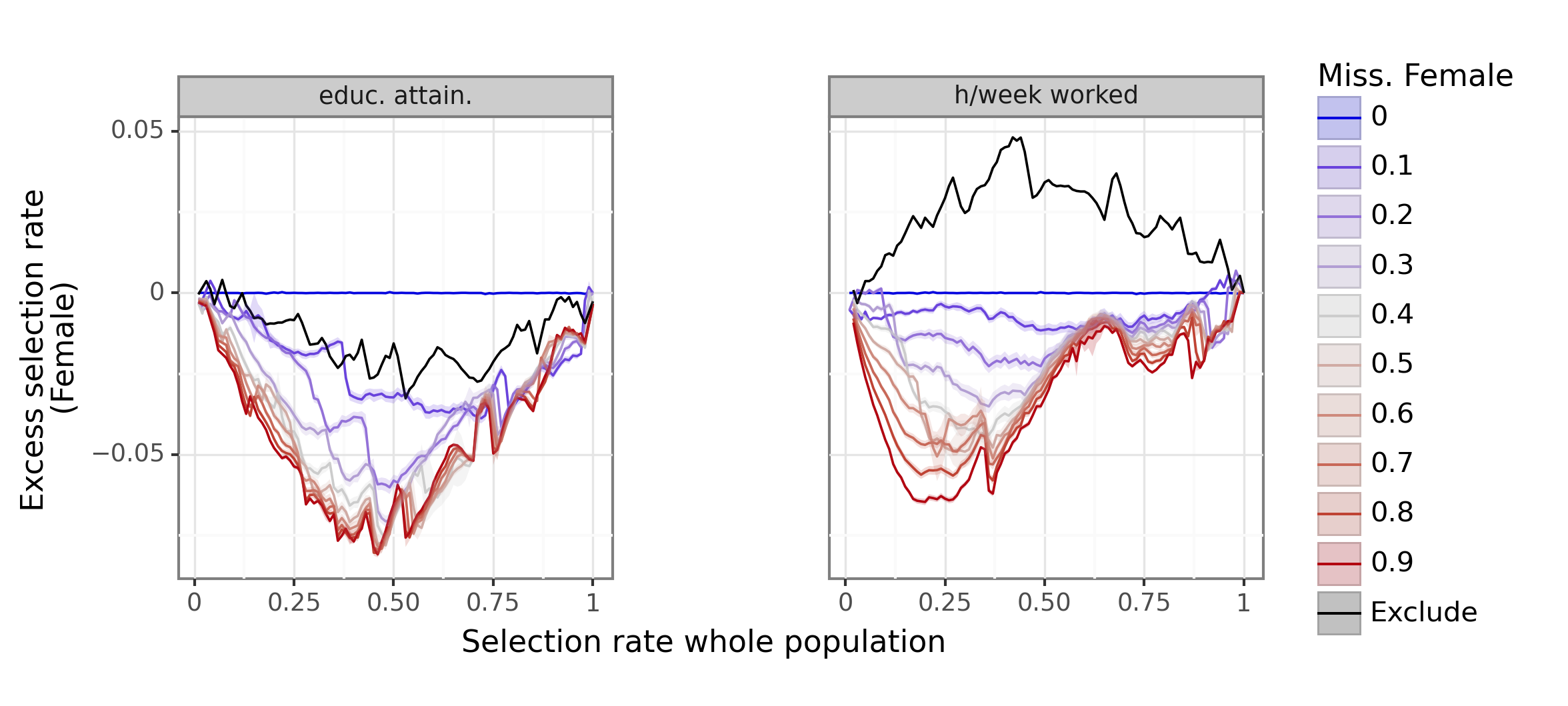}
        \subcaption{Female group}
    \end{subfigure}
    \begin{subfigure}[b]{0.49\linewidth}
        \includegraphics[scale=0.4]{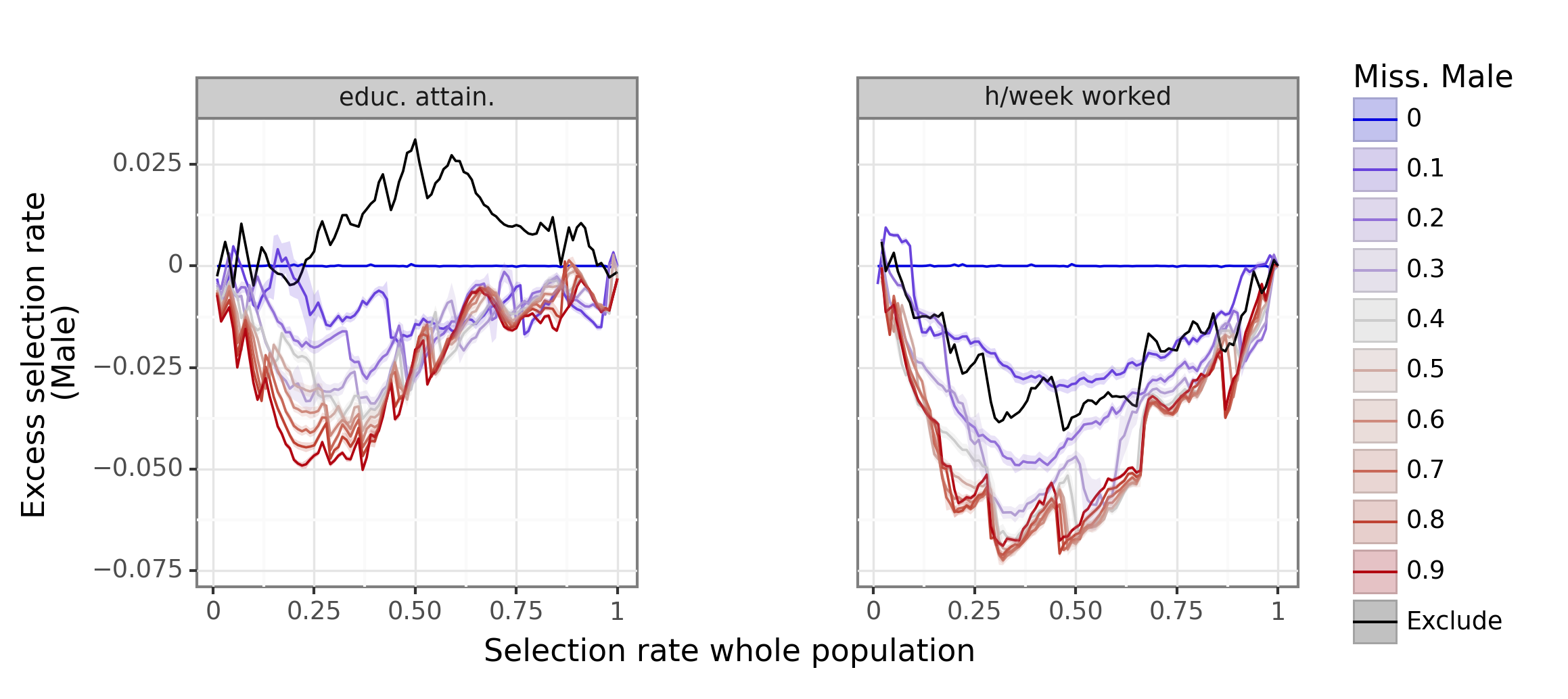}
        \subcaption{Male group}
    \end{subfigure}
    \caption{Group-wise excess selection rates using the ACS Income dataset. Each panel represents a feature that has been corrupted by under-reporting in independent runs of the experiment. Black curves show performance when omitting the entire feature column. Results are averaged over 50 runs on the test set. Shaded areas correspond to one standard deviation in each direction of the mean.}
    \label{fig:acsincome}
\end{figure}

\paragraph{ACS Income data}
Figure~\ref{fig:acsincome} summarizes the results for the experiments on the ACS Income data.
We see that feature under-reporting in `education attainment' and `hours worked per week' consistently leads to under-selection of the group with under-reporting.
This is true irrespective of whether feature under-reporting is injected into the female sub-group or the male sub-group, and we observe the same effect when under-reporting is added based on the individuals' racial group.
The figure additionally suggests that more under-reporting generally leads to increasing under-selection.
Intuitively, it makes sense that both education attainment and hours worked per week contribute positively to predicted income which is confirmed by the parameter estimates (Figure~\ref{fig:paramsacs}). Exploration of the covariance matrix of the unbiased features further reveals that all numeric columns in the dataset are positively correlated which together creates a setting reminiscent of the Case 2 scenario studied in Section~\ref{sec:xx}. At a high-level, our theoretical analysis predicts that the group with more under-reporting is under-selected in this setting which aligns with our observations.
In addition to selection rate disparity, feature under-reporting in the data also leads to decreased model accuracy as displayed in Figure~\ref{fig:acsincome_test_r2}, and the parameter estimates in Figure~\ref{fig:paramsacs} display an attenuation effect as predicted in Section~\ref{sec:linreg}.

\begin{figure}
    \centering
    \begin{subfigure}[b]{1\linewidth}
    \centering
    \includegraphics[scale=0.38]{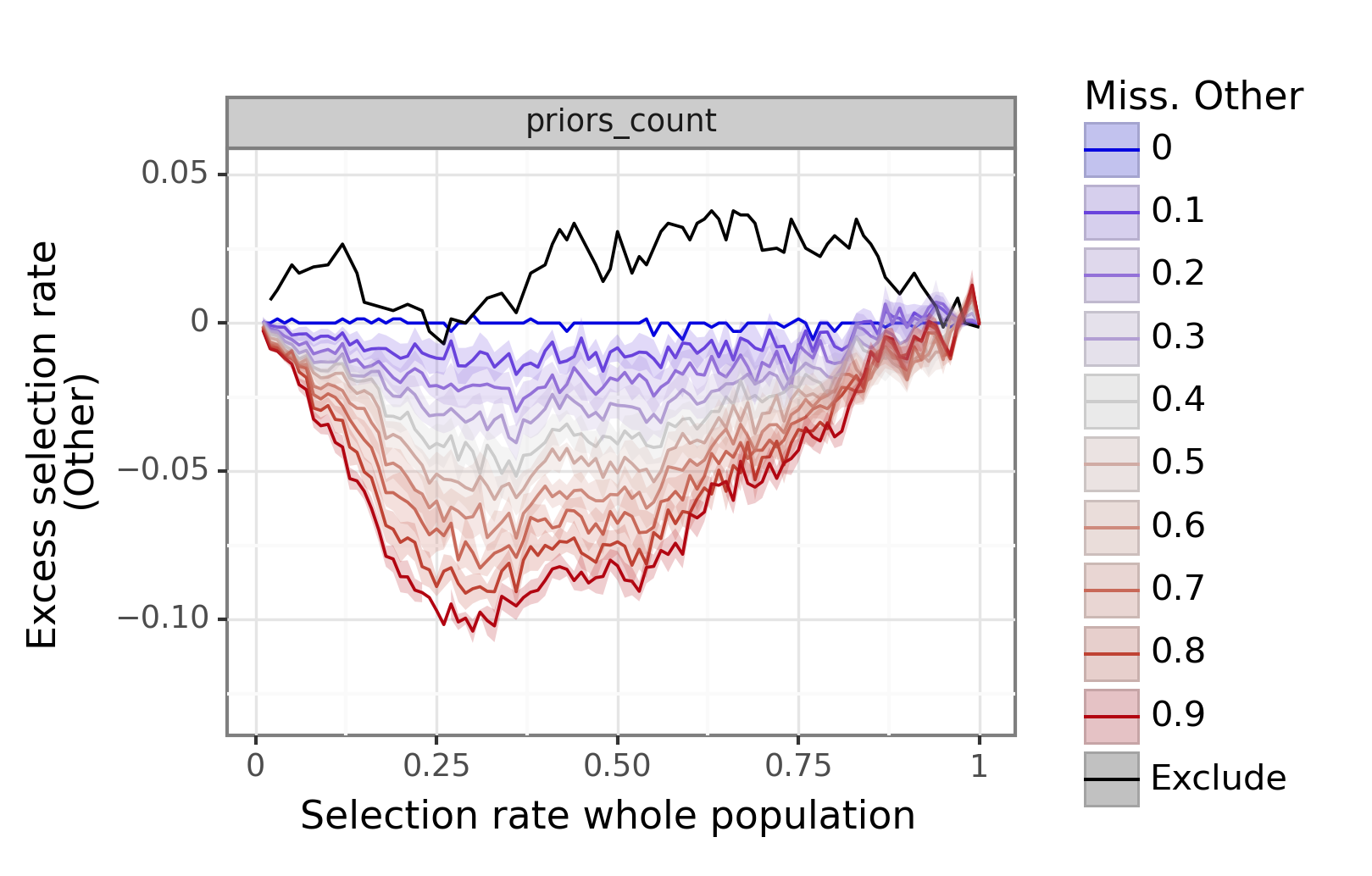}
    \includegraphics[scale=0.38]{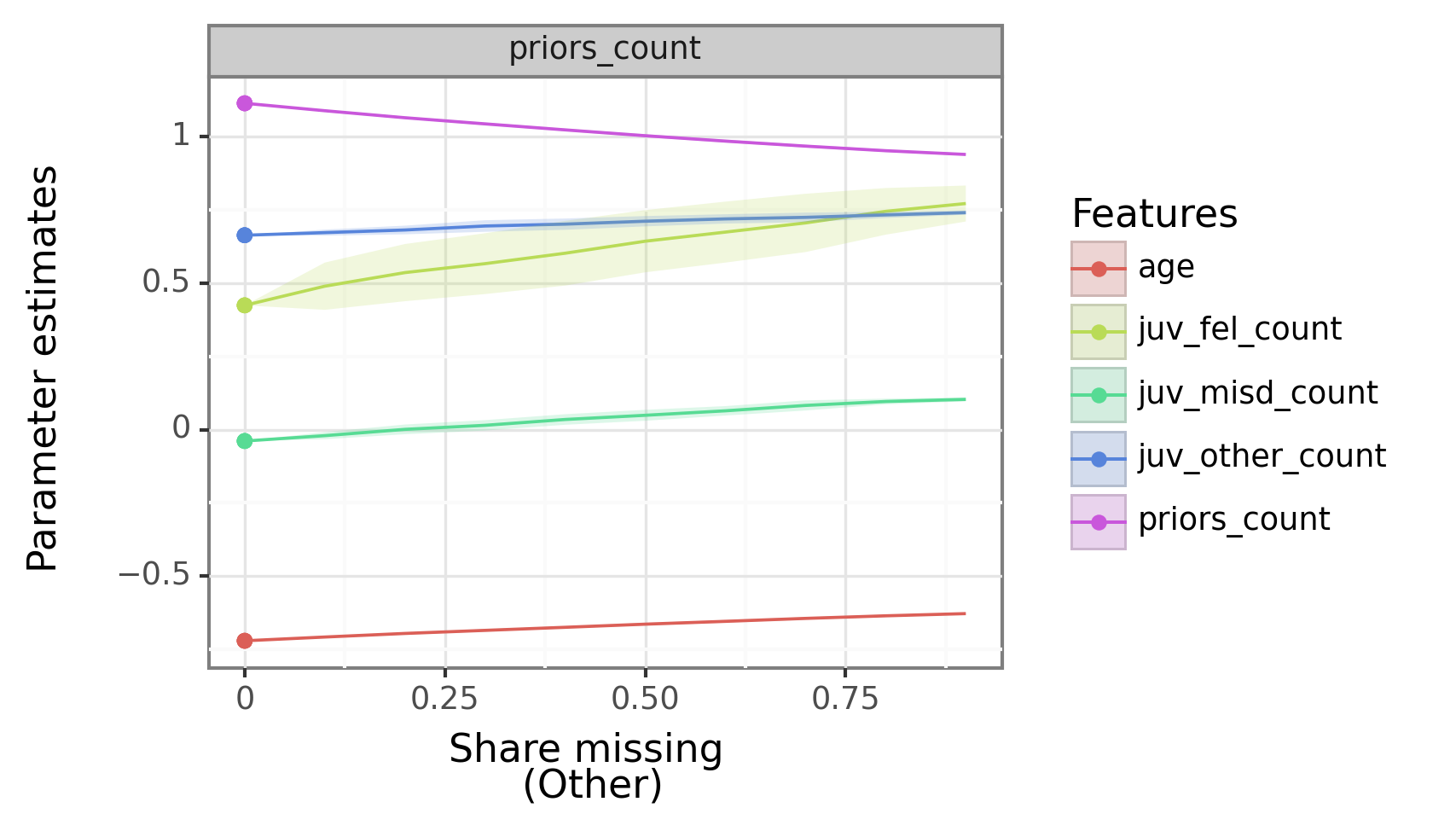}
    \includegraphics[scale=0.38]{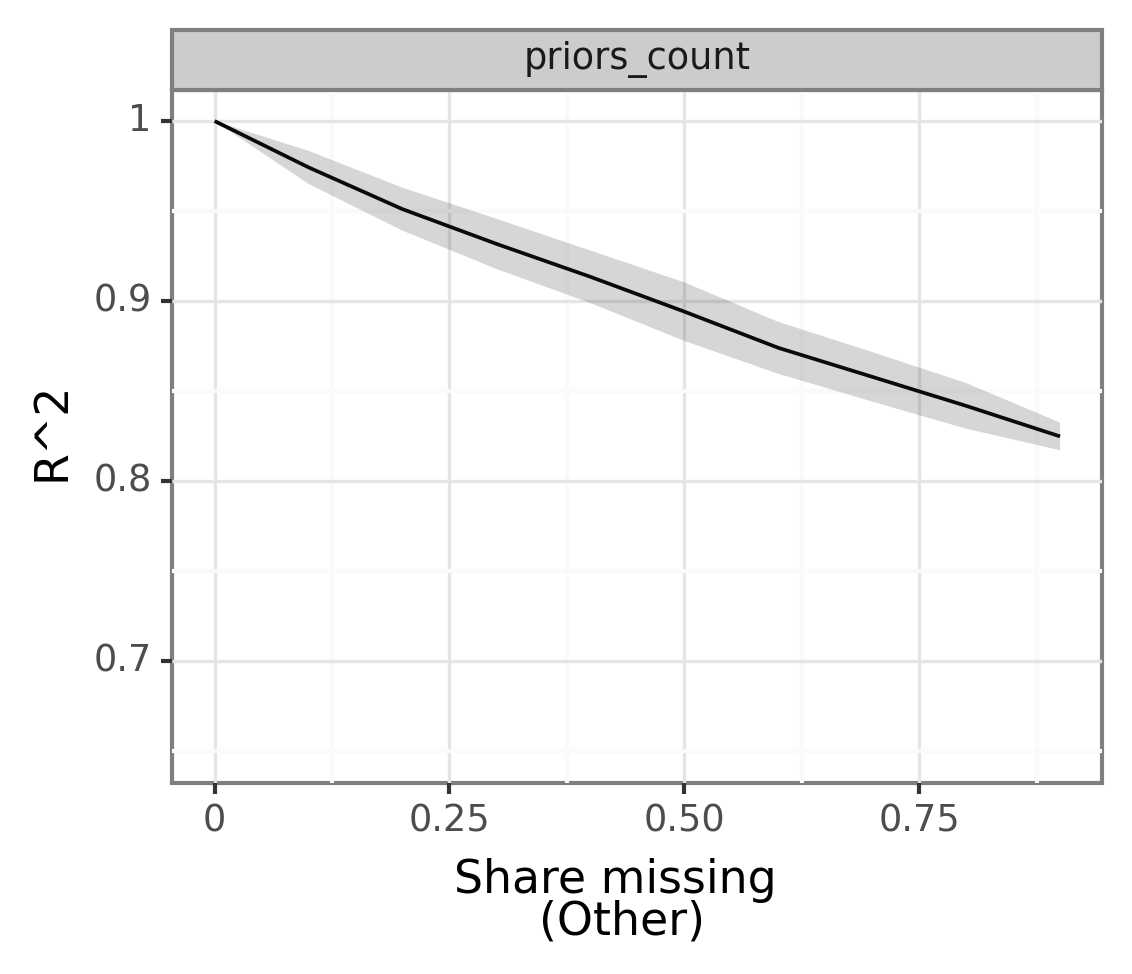}
    \subcaption{Estimation and prediction with under-reported feature}
    \end{subfigure}
    \begin{subfigure}[b]{1\linewidth}
    \centering
    \includegraphics[scale=0.38]{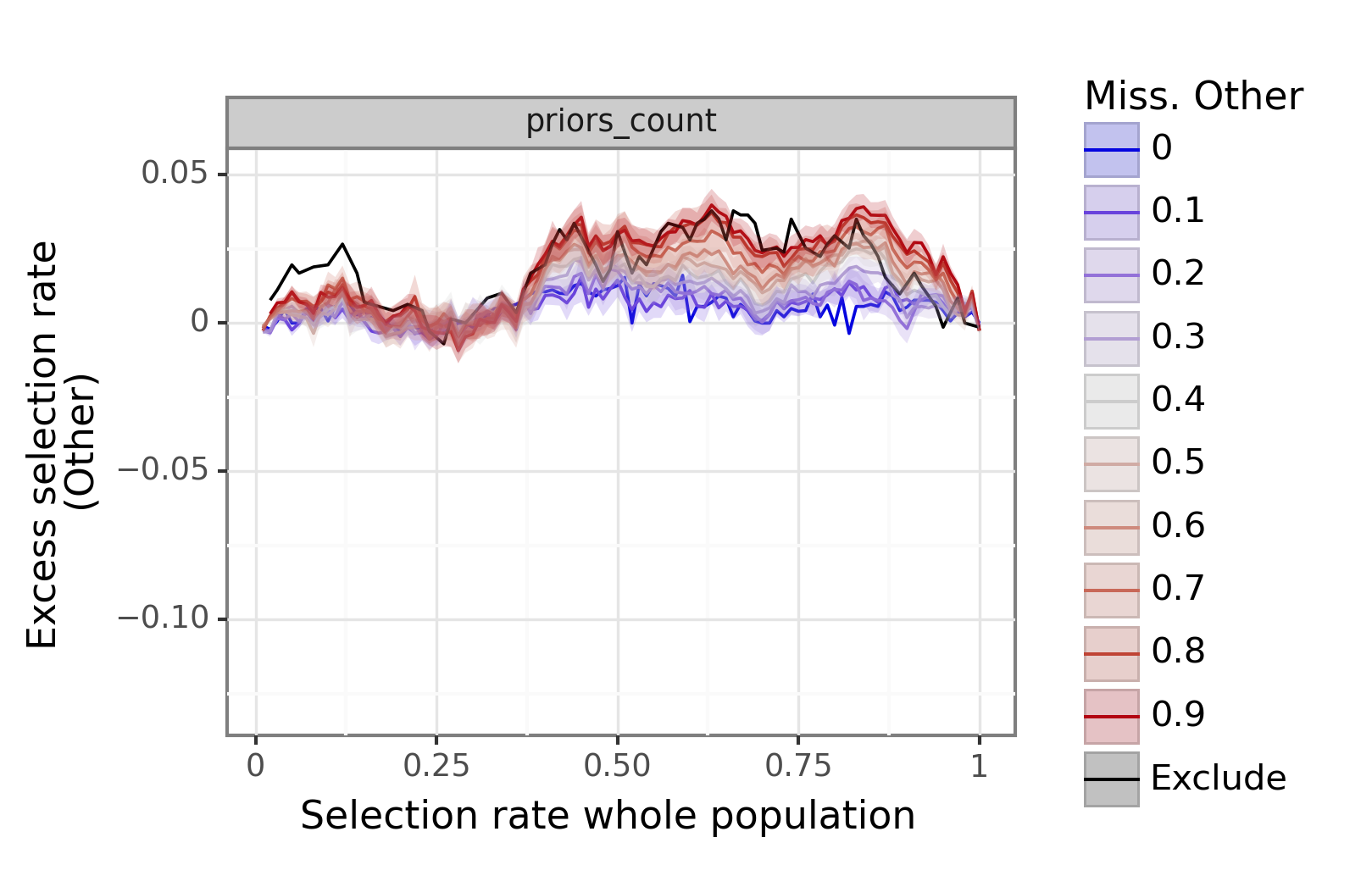}
    \includegraphics[scale=0.38]{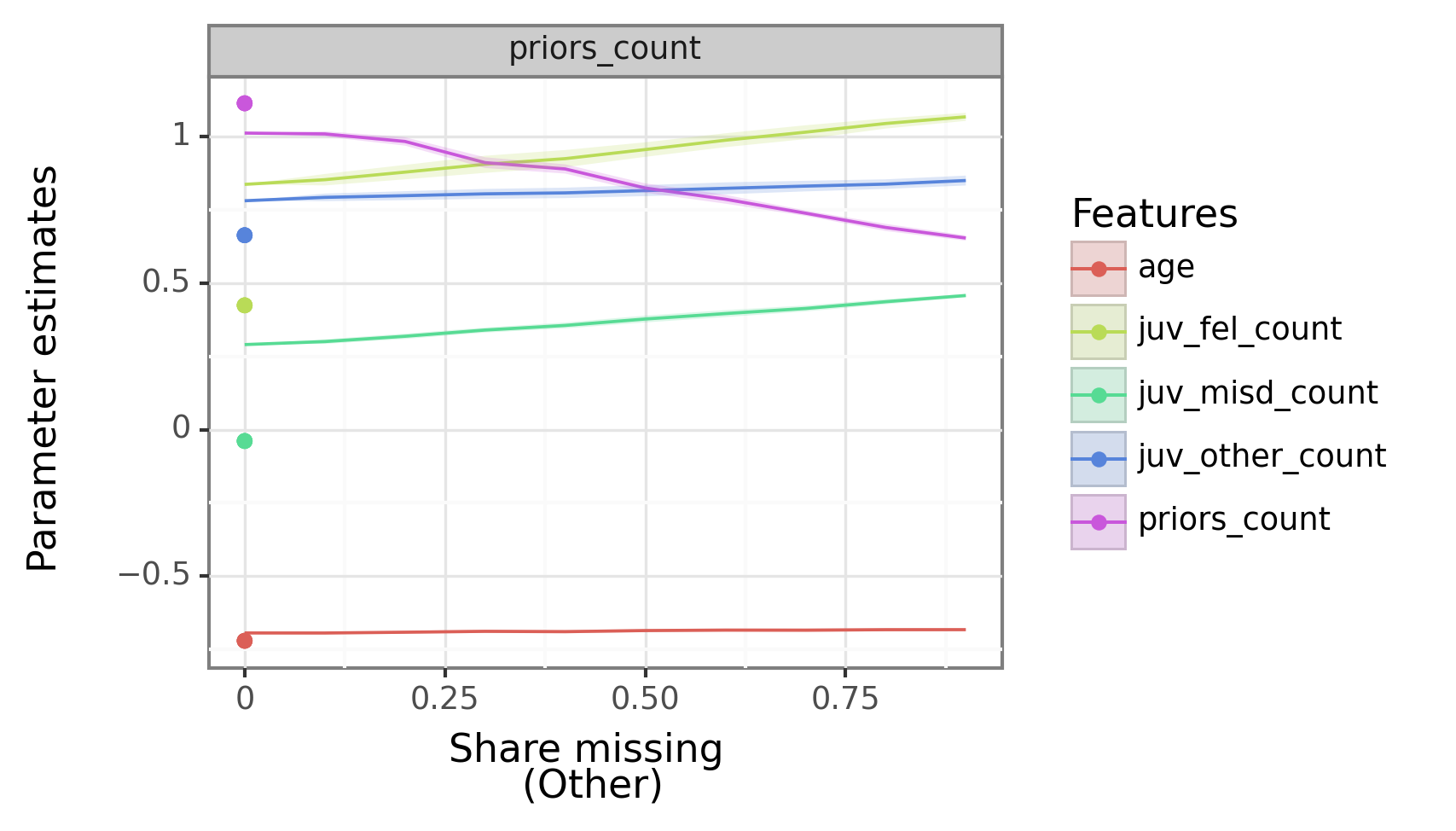}
    \includegraphics[scale=0.38]{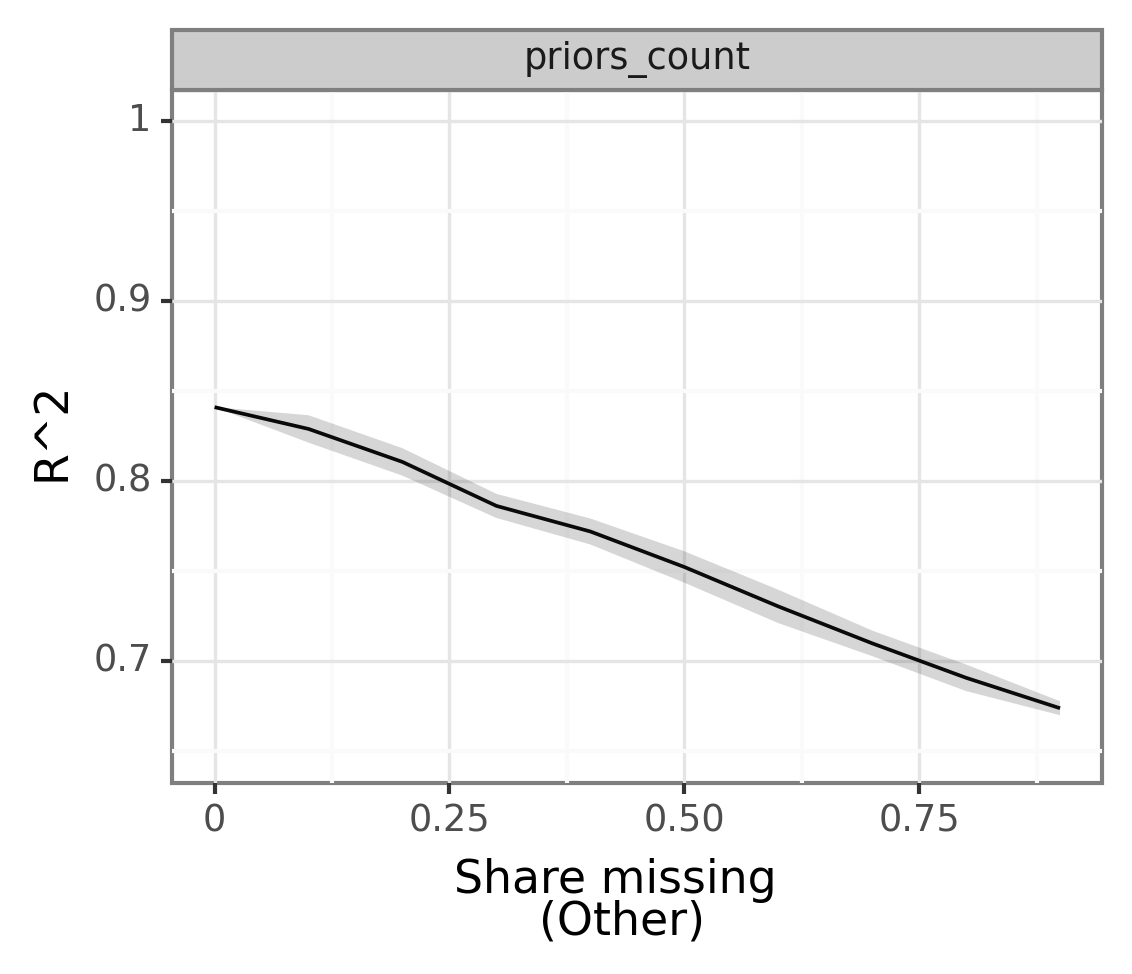}
    \subcaption{Estimation and prediction with multiple imputation}
    \end{subfigure}
    \begin{subfigure}[b]{1\linewidth}
    \centering
    \includegraphics[scale=0.38]{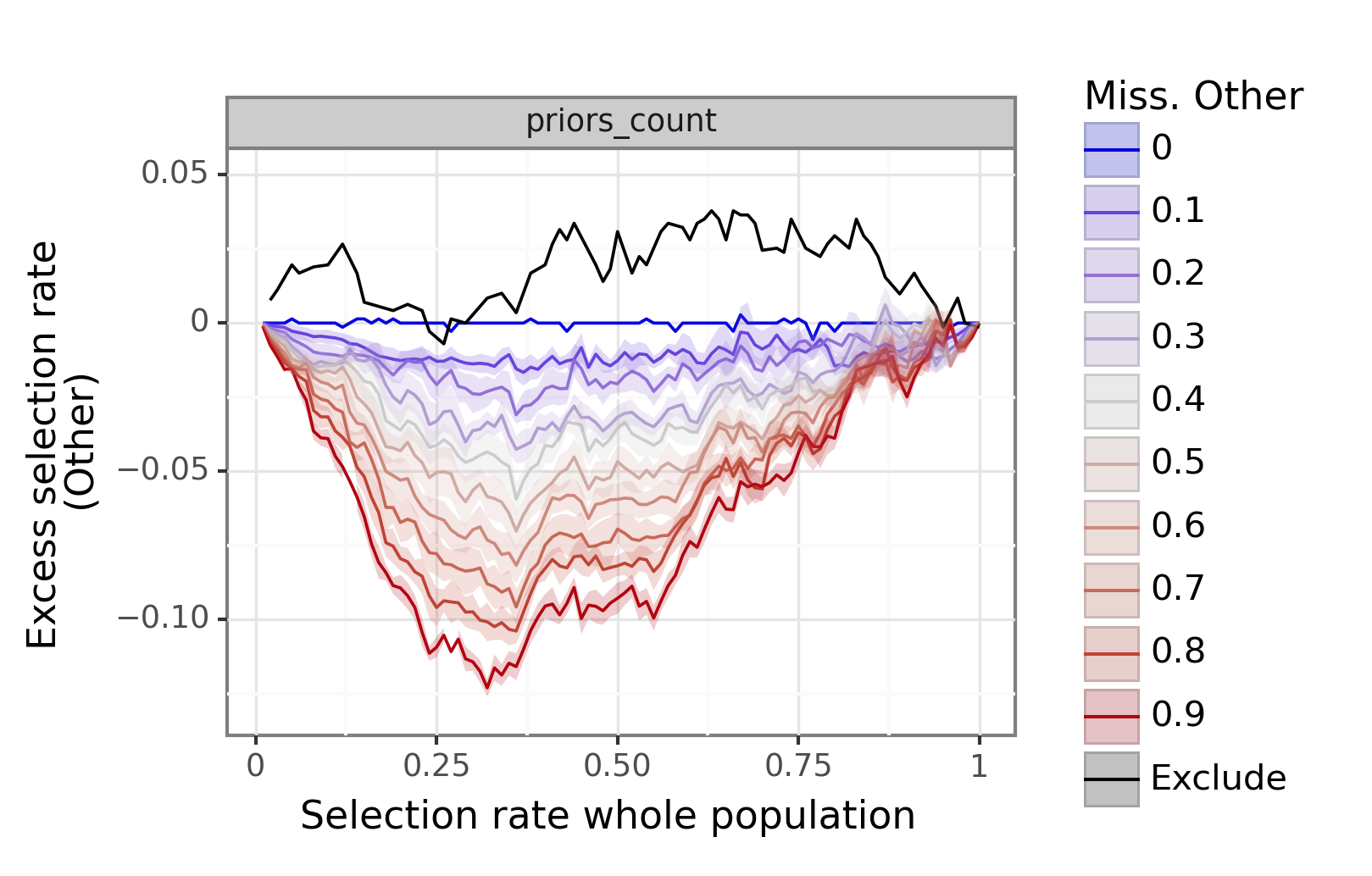}
    \includegraphics[scale=0.38]{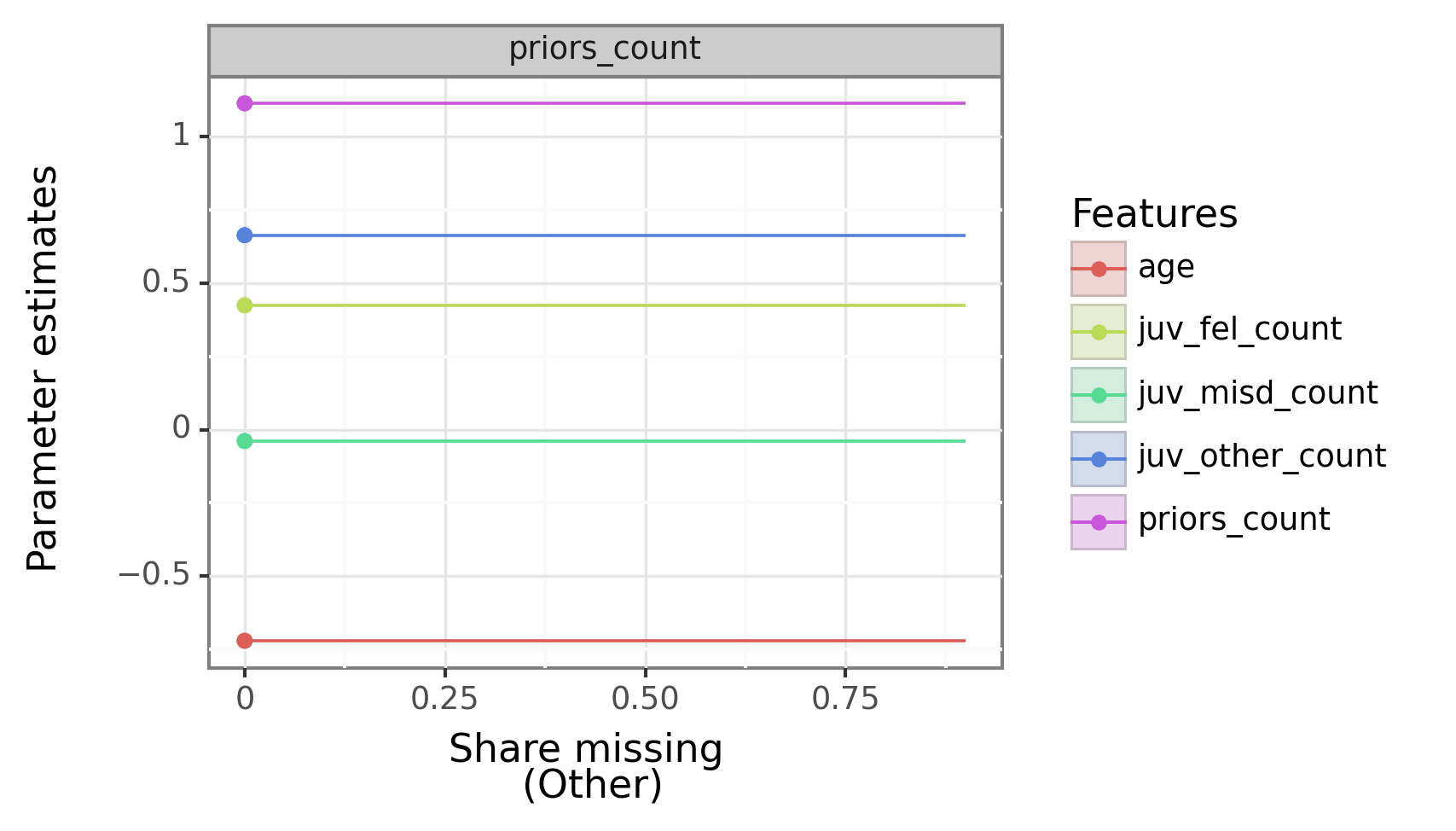}
    \includegraphics[scale=0.38]{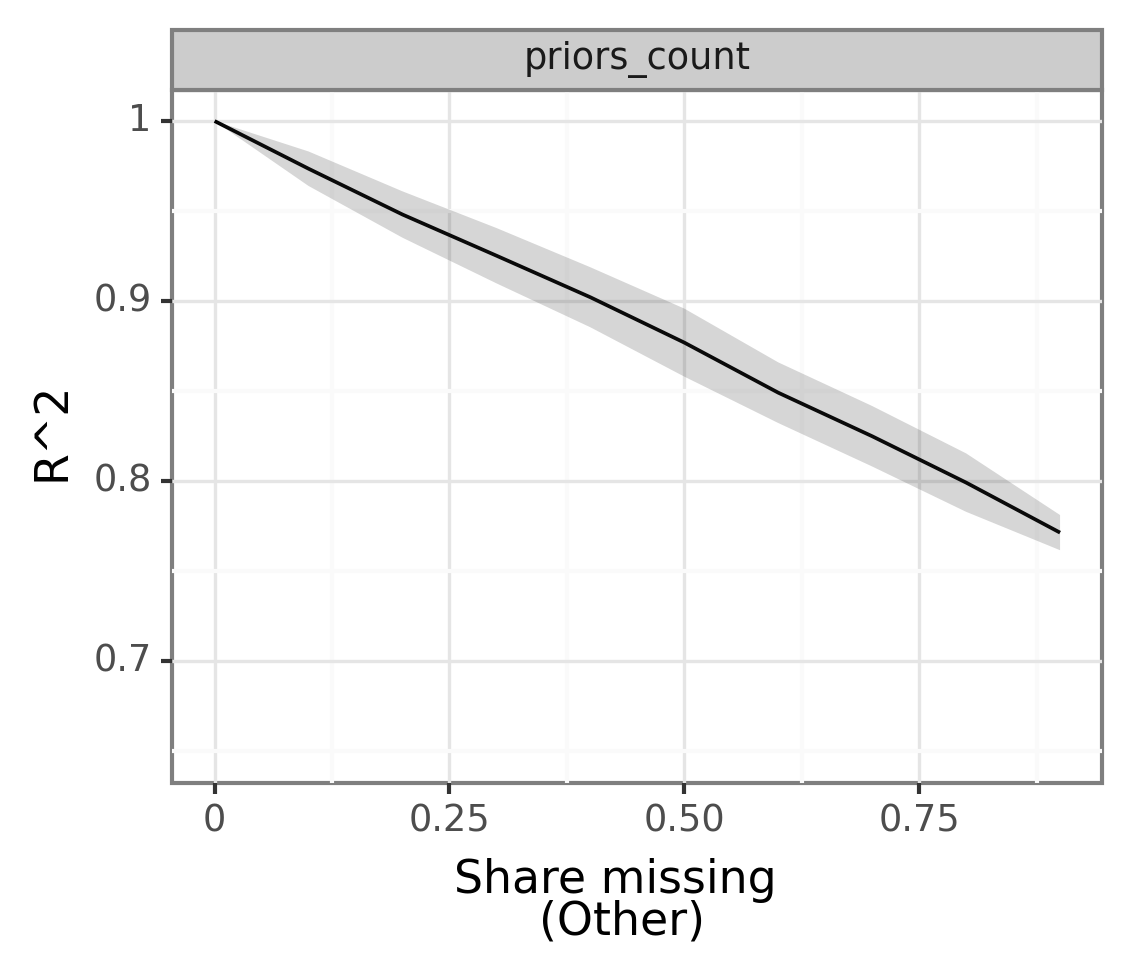}
    \subcaption{Estimation on rows with non-zero entries and prediction with under-reported feature}
    \end{subfigure}
    \begin{subfigure}[b]{1\linewidth}
    \centering
    \includegraphics[scale=0.38]{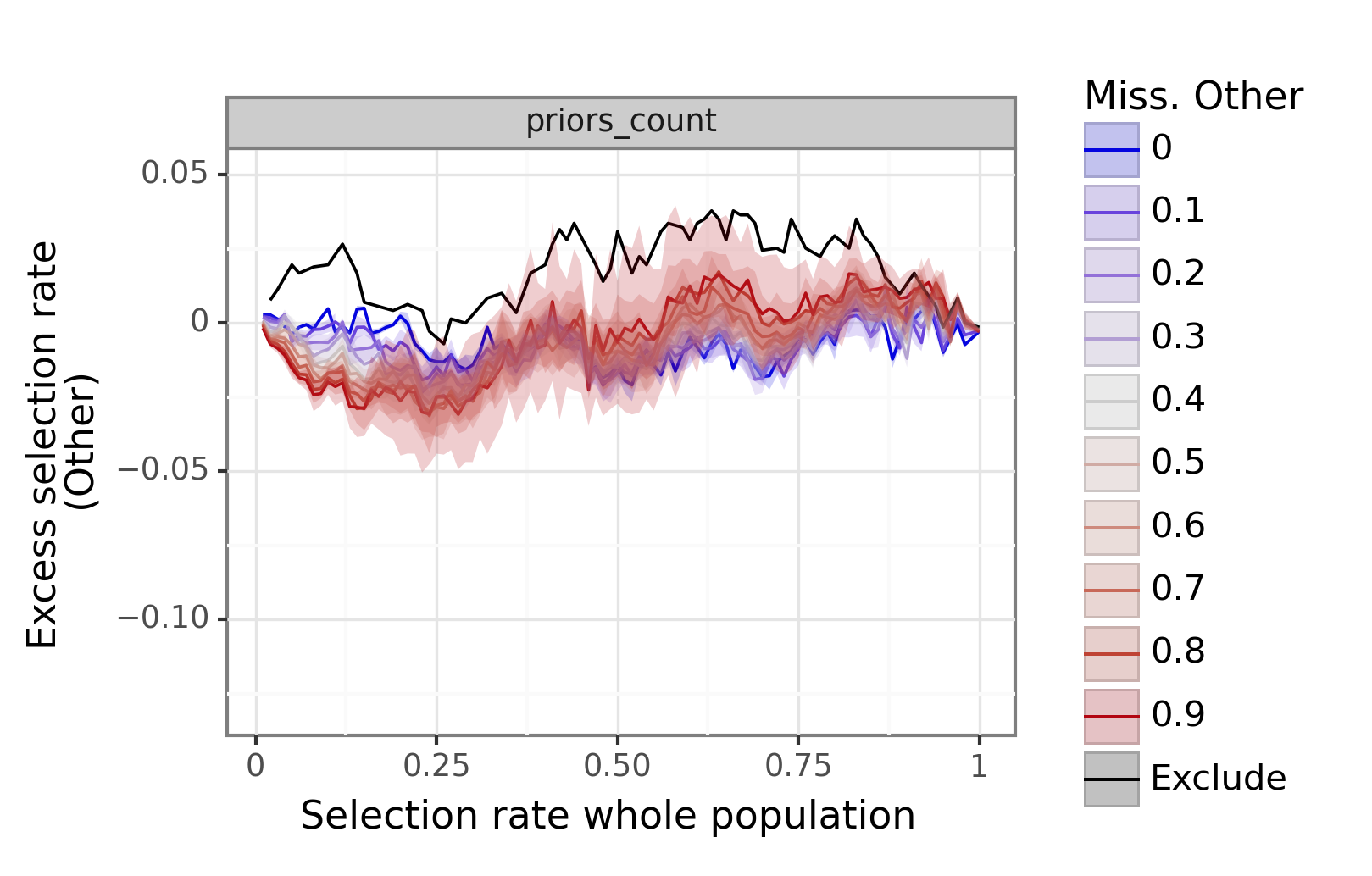}
    \includegraphics[scale=0.38]{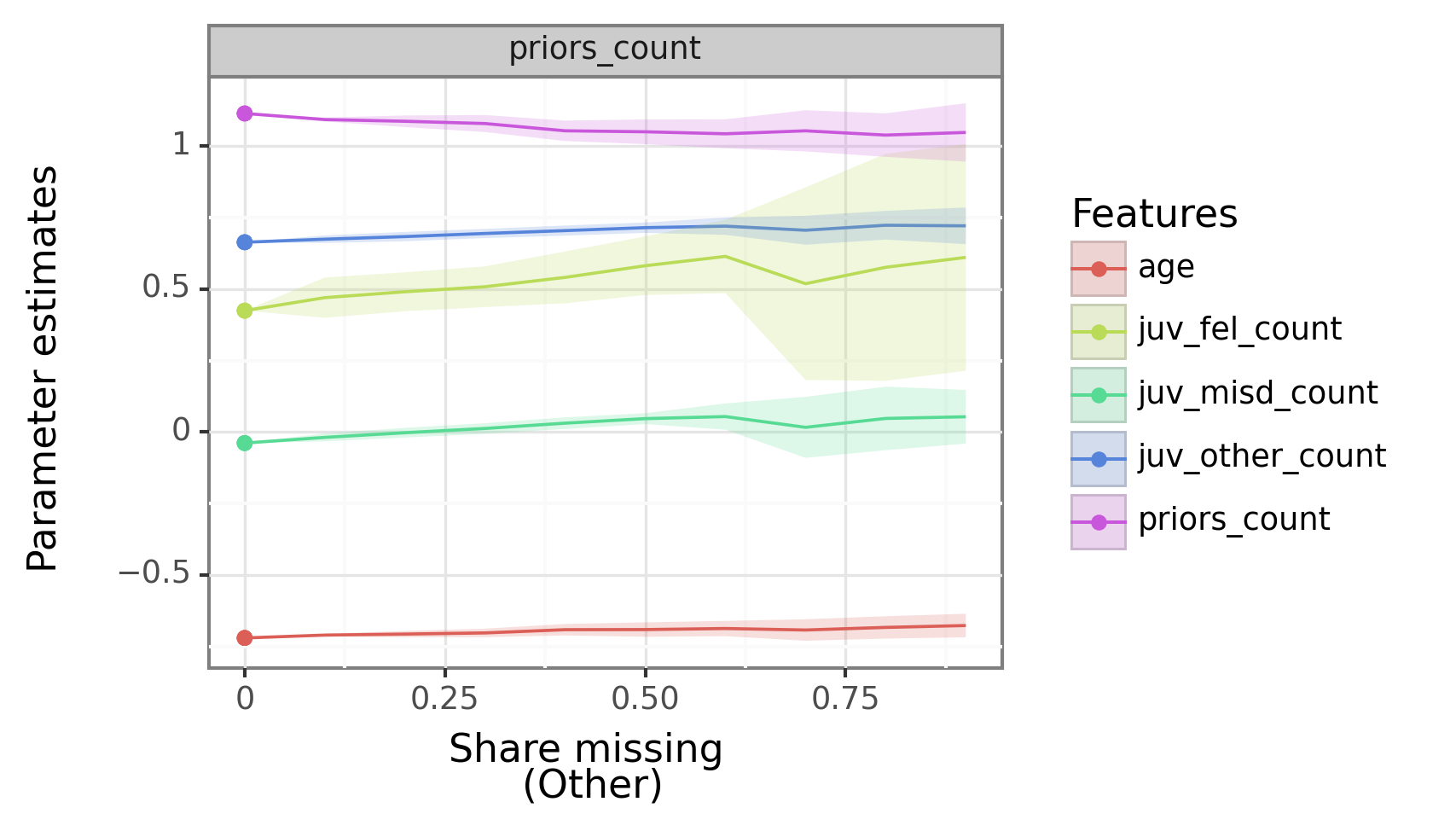}
    \includegraphics[scale=0.38]{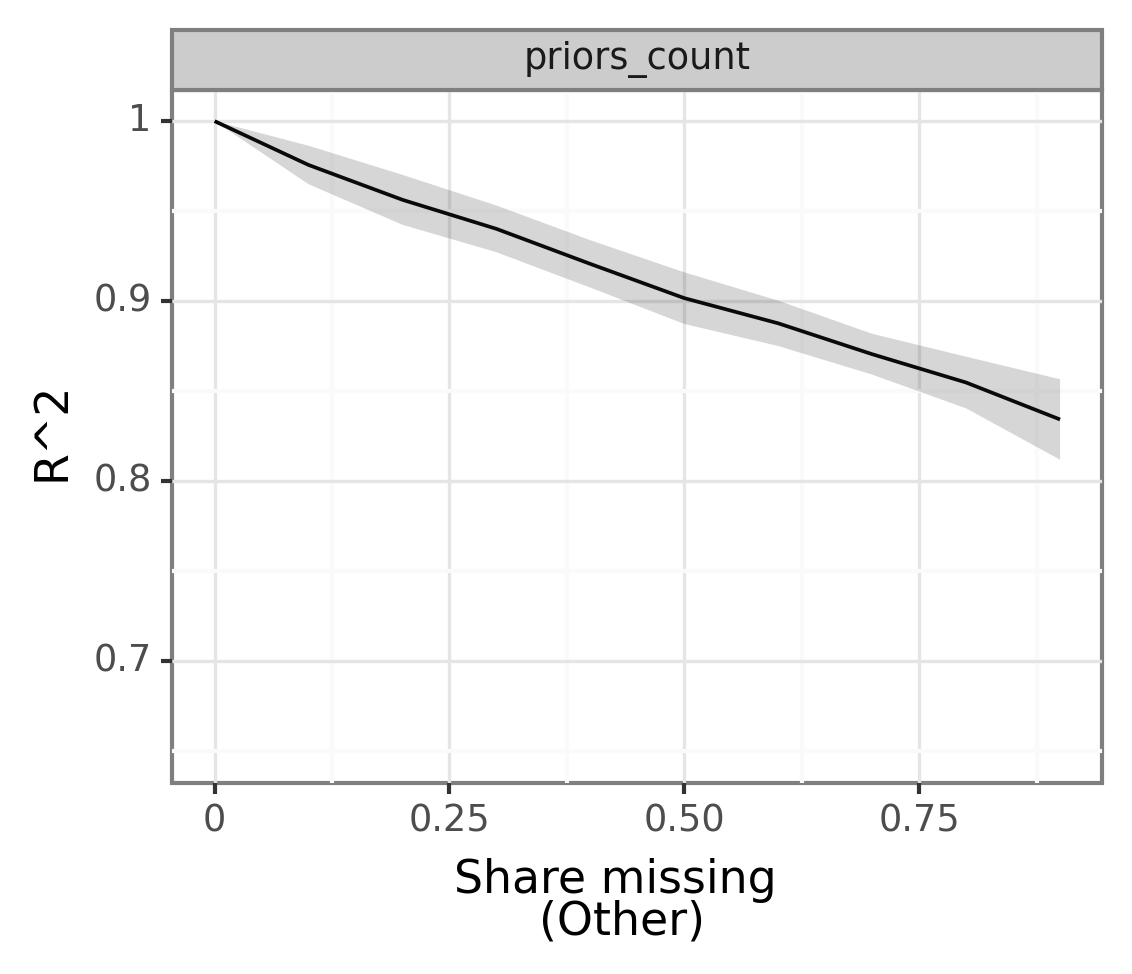}
    \subcaption{Estimation with group-dependent augmented loss and prediction with optimal group-dependent imputation (\textbf{our method})}
    \end{subfigure}
    \caption{Excess selection rates of group Other (i.e. not African-American) (left columns), parameter estimates (middle column), and test set $R^2$ (right columns) when under-reporting is injected into `priors count' in group Other using the COMPAS dataset and synthetic two-year recidivism outcomes. In (a), the model is trained and evaluated using the under-reported feature. For (b), we first train a multiple imputation model and then train and evaluate the prediction model using probabilistic imputations. For (c), the model is trained on only rows without 0-entries in `priors count' and evaluated on the under-reported data. In (d), we train with group-dependent augmented loss and use group-dependent optimal imputation values for prediction.
    Results are reported as averages over 30 runs. Shaded areas correspond to one standard deviation. The solid dots in the middle column correspond to true parameters. Note that in order to preserve readability, parameter estimates are only displayed for continuous features. Figure~\ref{fig:r_sq_facet} provides an overlay plot of the rightmost column for easy comparison.}
    \label{fig:panel}
\end{figure}

\paragraph{COMPAS data}
We focus on results for under-reporting in count features (Figure~\ref{fig:compas}) and point to Appendix~\ref{app:decreasingcompas} for additional results.
The feature `priors count', i.e. the number of previous criminal offenses individuals have been convicted of, emerges as important feature with respect to under-reporting.
Under-reporting in priors count leads to under-selection of the impacted group.
This pattern repeats itself for any of the groups and both of the available prediction outcomes. The more feature under-reporting in a group, the larger the occurring outcome disparity. Similarly to the previous results, this suggests a setting of Case 2 as discussed in Section~\ref{sec:xx}.
As before, parameter estimates suggest an attenuation effect which is displayed in Figure~\ref{fig:panel}.
Under-reporting in priors count could be interpreted as an extreme case of crimes that do not result in arrest. Assuming that one demographic group is more likely to be convicted for committed crimes than the other group, the result implies that the already more frequently targeted group may additionally be flagged as high risk for recidivism at disproportionate rates. Racial disparities in arrest rates and police encounters are well-documented in the US \cite[e.g.][]{Alexander2020-wj,Butcher22,Fogliato21,Pierson2020-xn} which highlights the importance of this finding.

\paragraph{Standard missing data methods}
Our experiments reveal that none of the standard missing data methods reliably mitigate the bias introduced through under-reporting and, instead, may themselves introduce disparities in selection rates.
Omission of the feature `hours worked per week' leads to over-selection of the female group and under-selection of the male group in the ACS Income data. This is because female individuals report to work on average less than male individuals (35.43h/week vs. 40.05h/week) while work hours contribute positively to income (Figure~\ref{fig:paramsacs}). Omitting the feature blinds the model to these differences.
Similar effects occur with the feature `education attainment', and `priors count' in the COMPAS data (Figures~\ref{fig:compas} and \ref{fig:panel}).

For multiple imputation on the COMPAS data, we see that the excess selection rate flips signs and the group with under-reported `priors count' is over-selected (Figures~\ref{fig:panel} and \ref{fig:multipmcompas}).
This is because the feature has a lot of true 0-entries that are wrongfully imputed as positive values. 
The cost incurred by these wrong imputations exceeds the benefit of imputation.
In comparison to training on mismeasured features directly, the parameter estimation bias is considerable even for small amounts of under-reporting.
For high under-reporting rates, the excess selection rate follows a similar pattern as the excess selection rate with feature omission since imputation is conducted using the features already present in the model adding little to no additional information.
Since our models are well-specified, row omission recovers the true parameter estimates as displayed for the COMPAS data and feature `priors count' in Figure~\ref{fig:panel}. Despite access to the ground-truth, we observe that under-reporting bias introduced at prediction time increases the selection rate disparity. 
While the model without row omission is able to shift weight to correlated features as more and more entries for `priors count' are under-reported, the row omission model cannot make use of the feature correlations ultimately leading to the increasing rather than decreasing disparities. With the same reasoning, the test set performance as measured by $R^2$ is decreased as displayed in the figure.

\begin{figure}[t]
    \centering
    \includegraphics[scale = 0.6]{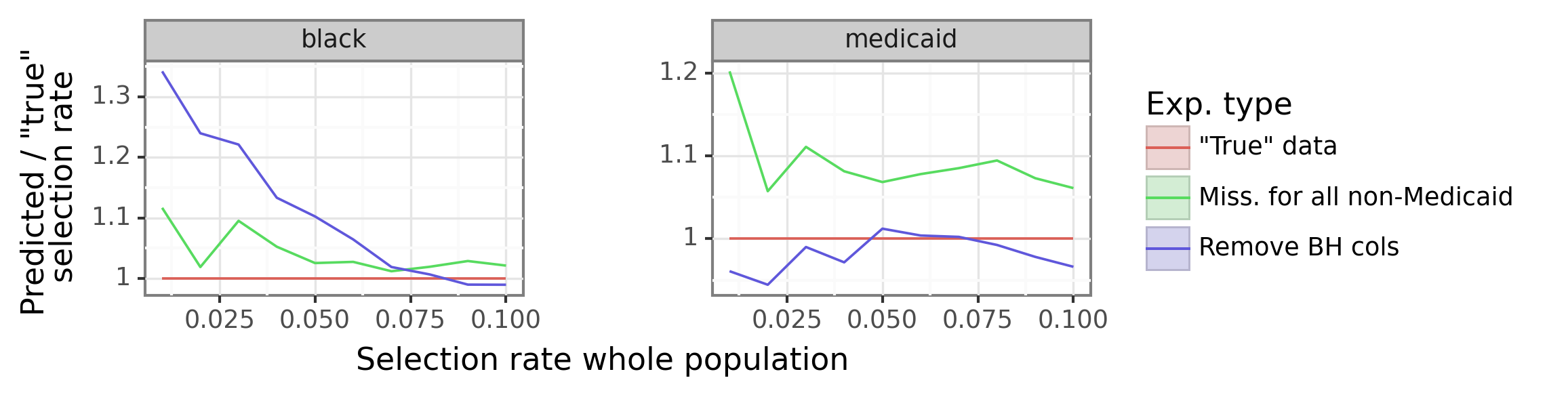}
    \caption{Selection rate fractions of different models. On the left, the results are displayed for the sub-population of Black individuals. On the right, the results are displayed for the sub-population that is insured through Medicaid. The selection rate of the whole population is considered to be 10\% or lower which reflects a realistic range for predictive risk modeling.}
    \label{fig:birthresults}
\end{figure}

\paragraph{Augmented loss and optimal prediction imputation}
We contrast the performance of our method and standard approaches for handling missing features at the example of the `priors count' feature in the COMPAS dataset.
Under-reporting rates are estimated with the procedure described in Section~\ref{sec:estimate_m}. We refer to Appendix~\ref{app:missingness_estimation} for further details on the under-reporting rate estimation.
The results in Figure~\ref{fig:panel} show that selection rate disparities decrease considerably when using group-dependent augmented loss and group-dependent optimal prediction imputation.
In contrast multiple imputation, this fairness improvement comes at no visible cost in performance. In fact, the average test set $R^2$ of the corrected model is very similar to, and even slightly higher than, the test set $R^2$ of the model trained directly on under-reported data (see Figure~\ref{fig:r_sq_facet}). Despite some variability, the average parameter estimates of the corrected model appear more stable across different amounts of under-reporting which suggests that the method successfully diminishes the bias introduced by under-reporting.

\paragraph{County-level birth data}
For the birth data, Figure~\ref{fig:birthresults} suggest that under-reporting of all behavioral health data for the non-Medicaid population leads to over-selection of the Medicaid population.
In the displayed overall selection rate range (<10\%), the Medicaid population is selected about 10\% more often than in the ``true'' data setting. Note that in reality this difference could be even larger because some of the ``true'' data  features were likely already under-reported. As shown in the Figure, some of the resulting disadvantage is still observable when evaluating performance for Black families. This can be explained by the fact that the two group variables are positively correlated in the dataset ($\rho=0.44$).
Excluding behavioral health features altogether leads to a reversal of selection rate disparities in the Medicaid / non-Medicaid groups, and significantly increases selection of the Black sub-population for overall selection rates less than about 7\%.
This provides additional evidence highlighting that omitting features is an unreliable solution for addressing the disparities arising from differential under-reporting. It underscores the point that such an approach lacks precision, potentially leading to arbitrary and inequitable outcomes.

\section{Discussion}

Differential feature under-reporting is a common phenomenon in administrative data. Data records are generally more complete for individuals who rely more consistently on public services (e.g. public health coverage).
In many predictive risk assessment settings, the segment of the population with more complete observations overlaps with sub-populations that are more commonly flagged as high-risk. Critics have argued that differential data availability is a key driver of the observed disparity in selection rates. When being classified as high risk subjects one to greater scrutiny of burden, this may disadvantage those with more complete data \citep{eubanks2018}.
Overall, the results of our study lend further credence to the concern by demonstrating how feature under-reporting generally leads to under-selection of a group that is already less frequently identified as high-risk. While, as we demonstrate, groups with greater data availability can theoretically be under-selected, the feature dependence structure under which this occurs appears to be uncommon in practice.

We illustrate the increased selection rate that individuals who rely on public healthcare coverage may experience at a real world example. Following the idea of Allegheny County's Hello Baby program \cite{hello-baby}, we build a model that predicts the risk that a newborn child will be removed from their family by Child Protective Services (CPS) within three years based on county-level data. The dataset contains behavioral and mental health information on the parents which can be assumed to be more complete for families that rely on public insurance. We note that, for privately insured individuals, some of this information may still be observed, e.g. because the individual was publicly insured previously, or individual information has been collected explicitly, but a lot of the information can be assumed missing.
Our experiments suggest that further under-reporting in behavioral health related information for the privately insured sub-population leads to an increase in high-risk predictions for the publicly insured group. We hypothesize that this effect would be even larger if the dataset was not already missing large portions of the feature observations for the privately insured group. This finding implies an unfair targeting of the publicly insured sub-population as high-risk. Since Black mothers in the data are publicly insured more frequently than mothers from other racial groups, these results also suggest that Black families are predicted to be at high-risk at unfairly inflated rates. Of course, depending on intervention type, a high risk classification may lead to an advantage or disadvantage for the families. In the Hello Baby setting, it is tied to eligibility for voluntary supportive services provided by county-funded service providers.

Our work proposes a technical remedy for the impact of under-reporting as a driver of disparities in selection rates. While standard missing data methods did not lead to more equitable outcomes in our experiments, these new methods reduced disparities considerably with little to no decrease in model accuracy. The applicability and performance in real-world administrative data settings like the Hello baby program remains an interesting and important avenue for future work.



\clearpage
\section*{Research ethics and social impact}

\paragraph{Ethical considerations statement}
The authors did not face ethical concerns that had to be mitigated while conducting this study. Experiments in this paper are based on the commonly used COMPAS and German credit datasets, 2018 US Census data from the American Community Survey, and a private county level data set. Result are aggregated over broad population groups and no identifiable information can be retrieved.  

\paragraph{Researcher positionality statement}
The authors recognize that their societal advantages give them certain benefits not shared by all individuals undergoing public sector risk assessment.
Thus, great care was taken in reflecting on the question: Does this work benefit us or the community at large? Since inflated selection rates can lead to tangible advantages (e.g. qualification for further publicly funded services) and significant disadvantages (e.g. unfavorable bail decisions in criminal risk assessment) depending on the application area, we believe that addressing problems of feature under-reporting ultimately benefits the community at large.

\paragraph{Adverse impact statement}

The authors believe that drawing attention to the problem of differential feature under-reporting has the potential to positively impact public sector risk assessment instruments for all individuals subjected to these systems. However, our work provides only a first step towards finding appropriate solutions to this problem. We propose a potential mitigation method and evaluate the method in a semi-synthetic setting, but
assessment of potential adverse effects of the method in real-world applications are beyond the scope of this paper. We clearly state this at the end of the discussion and call for future work in this direction.

\clearpage

\bibliographystyle{ACM-Reference-Format}
\bibliography{sample-base}


\begin{thebibliography}{62}


\ifx \showCODEN    \undefined \def \showCODEN     #1{\unskip}     \fi
\ifx \showDOI      \undefined \def \showDOI       #1{#1}\fi
\ifx \showISBNx    \undefined \def \showISBNx     #1{\unskip}     \fi
\ifx \showISBNxiii \undefined \def \showISBNxiii  #1{\unskip}     \fi
\ifx \showISSN     \undefined \def \showISSN      #1{\unskip}     \fi
\ifx \showLCCN     \undefined \def \showLCCN      #1{\unskip}     \fi
\ifx \shownote     \undefined \def \shownote      #1{#1}          \fi
\ifx \showarticletitle \undefined \def \showarticletitle #1{#1}   \fi
\ifx \showURL      \undefined \def \showURL       {\relax}        \fi
\providecommand\bibfield[2]{#2}
\providecommand\bibinfo[2]{#2}
\providecommand\natexlab[1]{#1}
\providecommand\showeprint[2][]{arXiv:#2}

\bibitem[Adams et~al\mbox{.}(2019)]%
        {Adams2019}
\bibfield{author}{\bibinfo{person}{Roy Adams}, \bibinfo{person}{Yuelong Ji}, \bibinfo{person}{Xiaobin Wang}, {and} \bibinfo{person}{Suchi Saria}.} \bibinfo{year}{2019}\natexlab{}.
\newblock \showarticletitle{Learning Models from Data with Measurement Error: Tackling Underreporting}. In \bibinfo{booktitle}{\emph{International Conference on Machine Learning (IMCL '20)}}.
\newblock


\bibitem[Ahmad et~al\mbox{.}(2019)]%
        {ahmad2019challenge}
\bibfield{author}{\bibinfo{person}{Muhammad~Aurangzeb Ahmad}, \bibinfo{person}{Carly Eckert}, {and} \bibinfo{person}{Ankur Teredesai}.} \bibinfo{year}{2019}\natexlab{}.
\newblock \showarticletitle{The Challenge of Imputation in Explainable Artificial Intelligence Models}.
\newblock \bibinfo{journal}{\emph{arXiv preprint, arXiv:1907.12669}} (\bibinfo{year}{2019}).
\newblock


\bibitem[Aigner and Cain(1977)]%
        {Aigner1977}
\bibfield{author}{\bibinfo{person}{Dennis~J. Aigner} {and} \bibinfo{person}{Glen~G. Cain}.} \bibinfo{year}{1977}\natexlab{}.
\newblock \showarticletitle{Statistical Theories of Discrimination in Labor Markets}.
\newblock \bibinfo{journal}{\emph{Industrial and Labor Relations Review}} \bibinfo{volume}{30}, \bibinfo{number}{2} (\bibinfo{date}{Jan.} \bibinfo{year}{1977}), \bibinfo{pages}{175}.
\newblock


\bibitem[Alexander(2010)]%
        {Alexander2020-wj}
\bibfield{author}{\bibinfo{person}{Michelle Alexander}.} \bibinfo{year}{2010}\natexlab{}.
\newblock \bibinfo{booktitle}{\emph{The new Jim Crow: Mass Incarceration in the Age of Colorblindness}}.
\newblock \bibinfo{publisher}{New Press}, \bibinfo{address}{New York, NY}.
\newblock


\bibitem[Angrist and Pischke(2008)]%
        {Angrist2008-oo}
\bibfield{author}{\bibinfo{person}{J~D Angrist} {and} \bibinfo{person}{Jorn-Steffen Pischke}.} \bibinfo{year}{2008}\natexlab{}.
\newblock \bibinfo{booktitle}{\emph{Mostly harmless econometrics}}.
\newblock \bibinfo{publisher}{Princeton University Press}, \bibinfo{address}{Princeton, NJ}.
\newblock


\bibitem[Angwin et~al\mbox{.}(2016)]%
        {compas}
\bibfield{author}{\bibinfo{person}{J. Angwin}, \bibinfo{person}{J. Larson}, \bibinfo{person}{S. Mattu}, {and} \bibinfo{person}{L. Kirchner}.} \bibinfo{year}{2016}\natexlab{}.
\newblock \showarticletitle{Machine Bias. There's software used across the country to predict future criminals. And it's biased against blacks.}
\newblock \bibinfo{journal}{\emph{ProPublica}} (\bibinfo{year}{2016}).
\newblock
\urldef\tempurl%
\url{https://www.propublica.org/article/machine-bias-risk-assessments-in-criminal-sentencing}
\showURL{%
\tempurl}


\bibitem[Arpey et~al\mbox{.}(2017)]%
        {Arpey2017-nh}
\bibfield{author}{\bibinfo{person}{Nicholas~C Arpey}, \bibinfo{person}{Anne~H Gaglioti}, {and} \bibinfo{person}{Marcy~E Rosenbaum}.} \bibinfo{year}{2017}\natexlab{}.
\newblock \showarticletitle{How socioeconomic status affects patient perceptions of health care: A qualitative study}.
\newblock \bibinfo{journal}{\emph{J. Prim. Care Community Health}} \bibinfo{volume}{8}, \bibinfo{number}{3} (\bibinfo{date}{July} \bibinfo{year}{2017}), \bibinfo{pages}{169--175}.
\newblock


\bibitem[Bailey et~al\mbox{.}(2005)]%
        {Bailey2005}
\bibfield{author}{\bibinfo{person}{T.C. Bailey}, \bibinfo{person}{M.S. Carvalho}, \bibinfo{person}{T.M. Lapa}, \bibinfo{person}{W.V. Souza}, {and} \bibinfo{person}{M.J. Brewer}.} \bibinfo{year}{2005}\natexlab{}.
\newblock \showarticletitle{Modeling of Under-detection of Cases in Disease Surveillance}.
\newblock \bibinfo{journal}{\emph{Annals of Epidemiology}} \bibinfo{volume}{15}, \bibinfo{number}{5} (\bibinfo{date}{May} \bibinfo{year}{2005}), \bibinfo{pages}{335--343}.
\newblock


\bibitem[Barnes and Hyatt(2012)]%
        {barnes2012classifying}
\bibfield{author}{\bibinfo{person}{Geoffrey~C Barnes} {and} \bibinfo{person}{Jordan~M Hyatt}.} \bibinfo{year}{2012}\natexlab{}.
\newblock \showarticletitle{Classifying adult probationers by forecasting future offending}.
\newblock \bibinfo{journal}{\emph{National Institute of Justice. Retrieved February}}  \bibinfo{volume}{4} (\bibinfo{year}{2012}), \bibinfo{pages}{2020}.
\newblock


\bibitem[Berk et~al\mbox{.}(2018)]%
        {Berk2018}
\bibfield{author}{\bibinfo{person}{Richard Berk}, \bibinfo{person}{Hoda Heidari}, \bibinfo{person}{Shahin Jabbari}, \bibinfo{person}{Michael Kearns}, {and} \bibinfo{person}{Aaron Roth}.} \bibinfo{year}{2018}\natexlab{}.
\newblock \showarticletitle{Fairness in Criminal Justice Risk Assessments: The State of the Art}.
\newblock \bibinfo{journal}{\emph{Sociological Methods {\&} Research}} \bibinfo{volume}{50}, \bibinfo{number}{1} (\bibinfo{date}{July} \bibinfo{year}{2018}), \bibinfo{pages}{3--44}.
\newblock


\bibitem[Black et~al\mbox{.}(2022)]%
        {black2022algorithmic}
\bibfield{author}{\bibinfo{person}{Emily Black}, \bibinfo{person}{Hadi Elzayn}, \bibinfo{person}{Alexandra Chouldechova}, \bibinfo{person}{Jacob Goldin}, {and} \bibinfo{person}{Daniel Ho}.} \bibinfo{year}{2022}\natexlab{}.
\newblock \showarticletitle{Algorithmic fairness and vertical equity: Income fairness with IRS tax audit models}. In \bibinfo{booktitle}{\emph{2022 ACM Conference on Fairness, Accountability, and Transparency}}.
\newblock


\bibitem[Brenner and Loomis(1994)]%
        {Brenner1994}
\bibfield{author}{\bibinfo{person}{Hermann Brenner} {and} \bibinfo{person}{Dana Loomis}.} \bibinfo{year}{1994}\natexlab{}.
\newblock \showarticletitle{Varied Forms of Bias Due to Nondifferential Error in Measuring Exposure}.
\newblock \bibinfo{journal}{\emph{Epidemiology}} \bibinfo{volume}{5}, \bibinfo{number}{5} (\bibinfo{year}{1994}), \bibinfo{pages}{510--517}.
\newblock


\bibitem[Bross(1954)]%
        {Bross1954-vr}
\bibfield{author}{\bibinfo{person}{Irwin Bross}.} \bibinfo{year}{1954}\natexlab{}.
\newblock \showarticletitle{Misclassification in 2 {X} 2 Tables}.
\newblock \bibinfo{journal}{\emph{Biometrics}} \bibinfo{volume}{10}, \bibinfo{number}{4} (\bibinfo{date}{Dec.} \bibinfo{year}{1954}), \bibinfo{pages}{478}.
\newblock


\bibitem[Butcher et~al\mbox{.}(2022)]%
        {Butcher22}
\bibfield{author}{\bibinfo{person}{Bradley Butcher}, \bibinfo{person}{Chris Robinson}, \bibinfo{person}{Miri Zilka}, \bibinfo{person}{Riccardo Fogliato}, \bibinfo{person}{Carolyn Ashurst}, {and} \bibinfo{person}{Adrian Weller}.} \bibinfo{year}{2022}\natexlab{}.
\newblock \showarticletitle{Racial Disparities in the Enforcement of Marijuana Violations in the US}. In \bibinfo{booktitle}{\emph{Proceedings of the 2022 AAAI/ACM Conference on AI, Ethics, and Society}} \emph{(\bibinfo{series}{AIES '22})}. \bibinfo{pages}{130–143}.
\newblock


\bibitem[Char et~al\mbox{.}(2018)]%
        {Char2018}
\bibfield{author}{\bibinfo{person}{Danton~S. Char}, \bibinfo{person}{Nigam~H. Shah}, {and} \bibinfo{person}{David Magnus}.} \bibinfo{year}{2018}\natexlab{}.
\newblock \showarticletitle{Implementing Machine Learning in Health Care {\textemdash} Addressing Ethical Challenges}.
\newblock \bibinfo{journal}{\emph{New England Journal of Medicine}} \bibinfo{volume}{378}, \bibinfo{number}{11} (\bibinfo{date}{March} \bibinfo{year}{2018}), \bibinfo{pages}{981--983}.
\newblock


\bibitem[Chen et~al\mbox{.}(2018)]%
        {chen2018my}
\bibfield{author}{\bibinfo{person}{Irene~Y. Chen}, \bibinfo{person}{Fredrik~D. Johansson}, {and} \bibinfo{person}{David Sontag}.} \bibinfo{year}{2018}\natexlab{}.
\newblock \showarticletitle{Why is My Classifier Discriminatory?}. In \bibinfo{booktitle}{\emph{32nd International Conference on Neural Information Processing Systems}} \emph{(\bibinfo{series}{NIPS'18})}.
\newblock


\bibitem[Chouldechova et~al\mbox{.}(2018)]%
        {chouldechova2018case}
\bibfield{author}{\bibinfo{person}{Alexandra Chouldechova}, \bibinfo{person}{Diana Benavides-Prado}, \bibinfo{person}{Oleksandr Fialko}, {and} \bibinfo{person}{Rhema Vaithianathan}.} \bibinfo{year}{2018}\natexlab{}.
\newblock \showarticletitle{A case study of algorithm-assisted decision making in child maltreatment hotline screening decisions}. In \bibinfo{booktitle}{\emph{Conference on Fairness, Accountability and Transparency (FAT*)}}.
\newblock


\bibitem[Chu et~al\mbox{.}(2006)]%
        {Chu2006}
\bibfield{author}{\bibinfo{person}{Haitao Chu}, \bibinfo{person}{Zhaojie Wang}, \bibinfo{person}{Stephen~R. Cole}, {and} \bibinfo{person}{Sander Greenland}.} \bibinfo{year}{2006}\natexlab{}.
\newblock \showarticletitle{Sensitivity Analysis of Misclassification: A Graphical and a Bayesian Approach}.
\newblock \bibinfo{journal}{\emph{Annals of Epidemiology}} \bibinfo{volume}{16}, \bibinfo{number}{11} (\bibinfo{year}{2006}), \bibinfo{pages}{834--841}.
\newblock


\bibitem[Cismondi et~al\mbox{.}(2013)]%
        {Cismondi2013}
\bibfield{author}{\bibinfo{person}{Federico Cismondi}, \bibinfo{person}{Andr{\'{e}}~S. Fialho}, \bibinfo{person}{Susana~M. Vieira}, \bibinfo{person}{Shane~R. Reti}, \bibinfo{person}{Jo{\~{a}}o~M.C. Sousa}, {and} \bibinfo{person}{Stan~N. Finkelstein}.} \bibinfo{year}{2013}\natexlab{}.
\newblock \showarticletitle{Missing data in medical databases: Impute, delete or classify?}
\newblock \bibinfo{journal}{\emph{Artificial Intelligence in Medicine}} \bibinfo{volume}{58}, \bibinfo{number}{1} (\bibinfo{date}{May} \bibinfo{year}{2013}), \bibinfo{pages}{63--72}.
\newblock


\bibitem[de~Oliveira et~al\mbox{.}(2017)]%
        {deOliveira2017}
\bibfield{author}{\bibinfo{person}{Guilherme~Lopes de Oliveira}, \bibinfo{person}{Rosangela~Helena Loschi}, {and} \bibinfo{person}{Renato~Martins Assun{\c{c}}{\~{a}}o}.} \bibinfo{year}{2017}\natexlab{}.
\newblock \showarticletitle{A random-censoring Poisson model for underreported data}.
\newblock \bibinfo{journal}{\emph{Statistics in Medicine}} \bibinfo{volume}{36}, \bibinfo{number}{30} (\bibinfo{date}{Oct.} \bibinfo{year}{2017}), \bibinfo{pages}{4873--4892}.
\newblock


\bibitem[Ding et~al\mbox{.}(2021)]%
        {folktables}
\bibfield{author}{\bibinfo{person}{Frances Ding}, \bibinfo{person}{Moritz Hardt}, \bibinfo{person}{John Miller}, {and} \bibinfo{person}{Ludwig Schmidt}.} \bibinfo{year}{2021}\natexlab{}.
\newblock \showarticletitle{Retiring Adult: New Datasets for Fair Machine Learning}. In \bibinfo{booktitle}{\emph{Advances in Neural Information Processing Systems (Neurips '21)}}.
\newblock


\bibitem[Dosemeci et~al\mbox{.}(1990)]%
        {DOSEMECI1990}
\bibfield{author}{\bibinfo{person}{Mustafa Dosemeci}, \bibinfo{person}{Sholom Wacholder}, {and} \bibinfo{person}{Jay~H. Lubin}.} \bibinfo{year}{1990}\natexlab{}.
\newblock \showarticletitle{Does Nondifferential Misclassification of Exposure Always Bias a True Effect Toward the Null Value?}
\newblock \bibinfo{journal}{\emph{American Journal of Epidemiology}} \bibinfo{volume}{132}, \bibinfo{number}{4} (\bibinfo{year}{1990}), \bibinfo{pages}{746--748}.
\newblock


\bibitem[Edwards et~al\mbox{.}(2013)]%
        {Edwards2013-jl}
\bibfield{author}{\bibinfo{person}{Jessie~K Edwards}, \bibinfo{person}{Stephen~R Cole}, \bibinfo{person}{Melissa~A Troester}, {and} \bibinfo{person}{David~B Richardson}.} \bibinfo{year}{2013}\natexlab{}.
\newblock \showarticletitle{Accounting for misclassified outcomes in binary regression models using multiple imputation with internal validation data}.
\newblock \bibinfo{journal}{\emph{Am. J. Epidemiol.}} \bibinfo{volume}{177}, \bibinfo{number}{9} (\bibinfo{date}{May} \bibinfo{year}{2013}), \bibinfo{pages}{904--912}.
\newblock


\bibitem[Elkan and Noto(2008)]%
        {Elkan2008}
\bibfield{author}{\bibinfo{person}{Charles Elkan} {and} \bibinfo{person}{Keith Noto}.} \bibinfo{year}{2008}\natexlab{}.
\newblock \showarticletitle{Learning classifiers from only positive and unlabeled data}. In \bibinfo{booktitle}{\emph{Proceedings of the 14th {ACM} {SIGKDD} international conference on Knowledge discovery and data mining}}. \bibinfo{publisher}{{ACM}}.
\newblock


\bibitem[Engin and Treleaven(2019)]%
        {engin2019algorithmic}
\bibfield{author}{\bibinfo{person}{Zeynep Engin} {and} \bibinfo{person}{Philip Treleaven}.} \bibinfo{year}{2019}\natexlab{}.
\newblock \showarticletitle{Algorithmic government: Automating public services and supporting civil servants in using data science technologies}.
\newblock \bibinfo{journal}{\emph{Comput. J.}} \bibinfo{volume}{62}, \bibinfo{number}{3} (\bibinfo{year}{2019}), \bibinfo{pages}{448--460}.
\newblock


\bibitem[Eubanks(2018)]%
        {eubanks2018}
\bibfield{author}{\bibinfo{person}{Virginia Eubanks}.} \bibinfo{year}{2018}\natexlab{}.
\newblock \bibinfo{title}{A Child Abuse Prediction Model Fails Poor Families}.
\newblock
\newblock
\urldef\tempurl%
\url{https://www.wired.com/story/excerpt-from-automating-inequality/}
\showURL{%
\tempurl}


\bibitem[Fernando et~al\mbox{.}(2021)]%
        {Fernando2021}
\bibfield{author}{\bibinfo{person}{Mart\'inez-Plumed Fernando}, \bibinfo{person}{Ferri C{\`{e}}sar}, \bibinfo{person}{Nieves David}, {and} \bibinfo{person}{Hern{\'{a}}ndez-Orallo Jos{\'{e}}}.} \bibinfo{year}{2021}\natexlab{}.
\newblock \showarticletitle{Missing the missing values: The ugly duckling of fairness in machine learning}.
\newblock \bibinfo{journal}{\emph{International Journal of Intelligent Systems}} \bibinfo{volume}{36}, \bibinfo{number}{7} (\bibinfo{date}{March} \bibinfo{year}{2021}), \bibinfo{pages}{3217--3258}.
\newblock


\bibitem[Fogliato et~al\mbox{.}(2021)]%
        {Fogliato21}
\bibfield{author}{\bibinfo{person}{Riccardo Fogliato}, \bibinfo{person}{Alice Xiang}, \bibinfo{person}{Zachary Lipton}, \bibinfo{person}{Daniel Nagin}, {and} \bibinfo{person}{Alexandra Chouldechova}.} \bibinfo{year}{2021}\natexlab{}.
\newblock \showarticletitle{On the Validity of Arrest as a Proxy for Offense: Race and the Likelihood of Arrest for Violent Crimes}. In \bibinfo{booktitle}{\emph{Proceedings of the 2021 AAAI/ACM Conference on AI, Ethics, and Society}} \emph{(\bibinfo{series}{AIES '21})}. \bibinfo{pages}{100–111}.
\newblock


\bibitem[Fricke(2020)]%
        {Fricke2020}
\bibfield{author}{\bibinfo{person}{Christian Fricke}.} \bibinfo{year}{2020}\natexlab{}.
\newblock \showarticletitle{Missing Fairness: The Discriminatory Effect of Missing Values inDatasets on Fairness in Machine Learning}.
\newblock \bibinfo{journal}{\emph{Master thesis}} (\bibinfo{year}{2020}).
\newblock


\bibitem[Fuller(1987)]%
        {fuller1987}
\bibfield{author}{\bibinfo{person}{Wayne~A Fuller}.} \bibinfo{year}{1987}\natexlab{}.
\newblock \bibinfo{booktitle}{\emph{Measurement Error Models}}.
\newblock \bibinfo{publisher}{John Wiley \& Sons}, \bibinfo{address}{Nashville, TN}.
\newblock


\bibitem[Gelman et~al\mbox{.}(2013)]%
        {Gelman2013-da}
\bibfield{author}{\bibinfo{person}{Andrew Gelman}, \bibinfo{person}{John~B Carlin}, \bibinfo{person}{Hal~S Stern}, \bibinfo{person}{David~B Dunson}, \bibinfo{person}{Aki Vehtari}, {and} \bibinfo{person}{Donald~B Rubin}.} \bibinfo{year}{2013}\natexlab{}.
\newblock \bibinfo{booktitle}{\emph{Bayesian Data Analysis} (\bibinfo{edition}{3} ed.)}.
\newblock \bibinfo{publisher}{Chapman \& Hall/CRC}, \bibinfo{address}{Philadelphia, PA}.
\newblock


\bibitem[Gianfrancesco et~al\mbox{.}(2018)]%
        {Gianfrancesco2018}
\bibfield{author}{\bibinfo{person}{Milena~A. Gianfrancesco}, \bibinfo{person}{Suzanne Tamang}, \bibinfo{person}{Jinoos Yazdany}, {and} \bibinfo{person}{Gabriela Schmajuk}.} \bibinfo{year}{2018}\natexlab{}.
\newblock \showarticletitle{Potential Biases in Machine Learning Algorithms Using Electronic Health Record Data}.
\newblock \bibinfo{journal}{\emph{{JAMA} Internal Medicine}} \bibinfo{volume}{178}, \bibinfo{number}{11} (\bibinfo{date}{Nov.} \bibinfo{year}{2018}), \bibinfo{pages}{1544}.
\newblock


\bibitem[Greenland(2014)]%
        {Greenland2014}
\bibfield{author}{\bibinfo{person}{Sander Greenland}.} \bibinfo{year}{2014}\natexlab{}.
\newblock \showarticletitle{Sensitivity Analysis and Bias Analysis}.
\newblock In \bibinfo{booktitle}{\emph{Handbook of Epidemiology}}. \bibinfo{publisher}{Springer New York}, \bibinfo{pages}{685--706}.
\newblock


\bibitem[Groenwold and Dekkers(2020)]%
        {Groenwold2020}
\bibfield{author}{\bibinfo{person}{Rolf H~H Groenwold} {and} \bibinfo{person}{Olaf~M Dekkers}.} \bibinfo{year}{2020}\natexlab{}.
\newblock \showarticletitle{Missing data: the impact of what is not there}.
\newblock \bibinfo{journal}{\emph{European Journal of Endocrinology}} \bibinfo{volume}{183}, \bibinfo{number}{4} (\bibinfo{date}{Oct.} \bibinfo{year}{2020}), \bibinfo{pages}{E7--E9}.
\newblock


\bibitem[Hausman(2001)]%
        {Hausman01}
\bibfield{author}{\bibinfo{person}{Jerry Hausman}.} \bibinfo{year}{2001}\natexlab{}.
\newblock \showarticletitle{Mismeasured Variables in Econometric Analysis: Problems from the Right and Problems from the Left}.
\newblock \bibinfo{journal}{\emph{Journal of Economic Perspectives}} \bibinfo{volume}{15}, \bibinfo{number}{4} (\bibinfo{year}{2001}), \bibinfo{pages}{57--67}.
\newblock


\bibitem[Houser and Sanders(2016)]%
        {houser2016use}
\bibfield{author}{\bibinfo{person}{Kimberly~A Houser} {and} \bibinfo{person}{Debra Sanders}.} \bibinfo{year}{2016}\natexlab{}.
\newblock \showarticletitle{The use of big data analytics by the IRS: Efficient solutions or the end of privacy as we know it}.
\newblock \bibinfo{journal}{\emph{Vand. J. Ent. \& Tech. L.}}  \bibinfo{volume}{19} (\bibinfo{year}{2016}), \bibinfo{pages}{817}.
\newblock


\bibitem[Jeanselme et~al\mbox{.}(2022)]%
        {Jeanselme_De-Arteaga_Zhang_Barrett_Tom_2022}
\bibfield{author}{\bibinfo{person}{Vincent Jeanselme}, \bibinfo{person}{Maria De-Arteaga}, \bibinfo{person}{Zhe Zhang}, \bibinfo{person}{Jessica Barrett}, {and} \bibinfo{person}{Brian Tom}.} \bibinfo{year}{2022}\natexlab{}.
\newblock \showarticletitle{Imputation Strategies Under Clinical Presence: Impact on Algorithmic Fairness}.
\newblock \bibinfo{journal}{\emph{Machine Learning for Health (ML4H '22)}}.
\newblock


\bibitem[Khani and Liang(2020)]%
        {khani2020feature}
\bibfield{author}{\bibinfo{person}{Fereshte Khani} {and} \bibinfo{person}{Percy Liang}.} \bibinfo{year}{2020}\natexlab{}.
\newblock \showarticletitle{Feature Noise Induces Loss Discrepancy Across Groups}. In \bibinfo{booktitle}{\emph{International Conference on Machine Learning}}. PMLR, \bibinfo{pages}{5209--5219}.
\newblock


\bibitem[King et~al\mbox{.}(2001)]%
        {King2001}
\bibfield{author}{\bibinfo{person}{G. King}, \bibinfo{person}{J. Honaker}, \bibinfo{person}{A. Joseph}, {and} \bibinfo{person}{K. Scheve}.} \bibinfo{year}{2001}\natexlab{}.
\newblock \showarticletitle{Analyzing incomplete political science data: an alternative algorithm for multipleimputation}.
\newblock \bibinfo{journal}{\emph{Am. Polit. Sci. Rev.95}} (\bibinfo{year}{2001}), \bibinfo{pages}{49--69}.
\newblock


\bibitem[Kithulgoda et~al\mbox{.}(2022)]%
        {kithulgoda2022predictive}
\bibfield{author}{\bibinfo{person}{Chamari~I Kithulgoda}, \bibinfo{person}{Rhema Vaithianathan}, {and} \bibinfo{person}{Dennis~P Culhane}.} \bibinfo{year}{2022}\natexlab{}.
\newblock \showarticletitle{Predictive risk modeling to identify homeless clients at risk for prioritizing services using routinely collected data}.
\newblock \bibinfo{journal}{\emph{Journal of Technology in Human Services}} \bibinfo{volume}{40}, \bibinfo{number}{2} (\bibinfo{year}{2022}), \bibinfo{pages}{134--156}.
\newblock


\bibitem[Levy et~al\mbox{.}(2021)]%
        {levy2021algorithms}
\bibfield{author}{\bibinfo{person}{Karen Levy}, \bibinfo{person}{Kyla~E Chasalow}, {and} \bibinfo{person}{Sarah Riley}.} \bibinfo{year}{2021}\natexlab{}.
\newblock \showarticletitle{Algorithms and decision-making in the public sector}.
\newblock \bibinfo{journal}{\emph{Annual Review of Law and Social Science}}  \bibinfo{volume}{17} (\bibinfo{year}{2021}), \bibinfo{pages}{309--334}.
\newblock


\bibitem[Mayson(2019)]%
        {mayson2019bias}
\bibfield{author}{\bibinfo{person}{Sandra~G Mayson}.} \bibinfo{year}{2019}\natexlab{}.
\newblock \showarticletitle{Bias in, bias out}.
\newblock \bibinfo{journal}{\emph{The Yale Law Journal}} \bibinfo{volume}{128}, \bibinfo{number}{8} (\bibinfo{year}{2019}), \bibinfo{pages}{2218--2300}.
\newblock


\bibitem[McCarthy et~al\mbox{.}(2015)]%
        {mccarthy2015predictive}
\bibfield{author}{\bibinfo{person}{John~F McCarthy}, \bibinfo{person}{Robert~M Bossarte}, \bibinfo{person}{Ira~R Katz}, \bibinfo{person}{Caitlin Thompson}, \bibinfo{person}{Janet Kemp}, \bibinfo{person}{Claire~M Hannemann}, \bibinfo{person}{Christopher Nielson}, {and} \bibinfo{person}{Michael Schoenbaum}.} \bibinfo{year}{2015}\natexlab{}.
\newblock \showarticletitle{Predictive modeling and concentration of the risk of suicide: implications for preventive interventions in the US Department of Veterans Affairs}.
\newblock \bibinfo{journal}{\emph{American journal of public health}} \bibinfo{volume}{105}, \bibinfo{number}{9} (\bibinfo{year}{2015}), \bibinfo{pages}{1935--1942}.
\newblock


\bibitem[McKnight et~al\mbox{.}(2007)]%
        {McKnight2007}
\bibfield{author}{\bibinfo{person}{P.E. McKnight}, \bibinfo{person}{K.M. McKnight}, \bibinfo{person}{S. Sidani}, {and} \bibinfo{person}{A.~J. Figueredo}.} \bibinfo{year}{2007}\natexlab{}.
\newblock \bibinfo{booktitle}{\emph{Missing Data: A Gentle Introduction}}.
\newblock


\bibitem[Natarajan et~al\mbox{.}(2013)]%
        {Natarajan13}
\bibfield{author}{\bibinfo{person}{Nagarajan Natarajan}, \bibinfo{person}{Inderjit~S Dhillon}, \bibinfo{person}{Pradeep~K Ravikumar}, {and} \bibinfo{person}{Ambuj Tewari}.} \bibinfo{year}{2013}\natexlab{}.
\newblock \showarticletitle{Learning with Noisy Labels}. In \bibinfo{booktitle}{\emph{Advances in Neural Information Processing Systems}}, \bibfield{editor}{\bibinfo{person}{C.J. Burges}, \bibinfo{person}{L.~Bottou}, \bibinfo{person}{M.~Welling}, \bibinfo{person}{Z.~Ghahramani}, {and} \bibinfo{person}{K.Q. Weinberger}} (Eds.), Vol.~\bibinfo{volume}{26}. \bibinfo{publisher}{Curran Associates, Inc.}
\newblock


\bibitem[Nemirovski et~al\mbox{.}(2009)]%
        {Nemirovski2009}
\bibfield{author}{\bibinfo{person}{A. Nemirovski}, \bibinfo{person}{A. Juditsky}, \bibinfo{person}{G. Lan}, {and} \bibinfo{person}{A. Shapiro}.} \bibinfo{year}{2009}\natexlab{}.
\newblock \showarticletitle{Robust Stochastic Approximation Approach to Stochastic Programming}.
\newblock \bibinfo{journal}{\emph{SIAM Journal on Optimization}} \bibinfo{volume}{19}, \bibinfo{number}{4} (\bibinfo{year}{2009}), \bibinfo{pages}{1574--1609}.
\newblock


\bibitem[Phelps(1972)]%
        {Phelps1972}
\bibfield{author}{\bibinfo{person}{Edmund Phelps}.} \bibinfo{year}{1972}\natexlab{}.
\newblock \showarticletitle{The Statistical Theory of Racism and Sexism}.
\newblock \bibinfo{journal}{\emph{American Economic Review}} \bibinfo{volume}{62}, \bibinfo{number}{4} (\bibinfo{year}{1972}), \bibinfo{pages}{659--61}.
\newblock


\bibitem[Pierson et~al\mbox{.}(2020)]%
        {Pierson2020-xn}
\bibfield{author}{\bibinfo{person}{Emma Pierson}, \bibinfo{person}{Camelia Simoiu}, \bibinfo{person}{Jan Overgoor}, \bibinfo{person}{Sam Corbett-Davies}, \bibinfo{person}{Daniel Jenson}, \bibinfo{person}{Amy Shoemaker}, \bibinfo{person}{Vignesh Ramachandran}, \bibinfo{person}{Phoebe Barghouty}, \bibinfo{person}{Cheryl Phillips}, \bibinfo{person}{Ravi Shroff}, {and} \bibinfo{person}{Sharad Goel}.} \bibinfo{year}{2020}\natexlab{}.
\newblock \showarticletitle{A large-scale analysis of racial disparities in police stops across the United States}.
\newblock \bibinfo{journal}{\emph{Nat. Hum. Behav.}} \bibinfo{volume}{4}, \bibinfo{number}{7} (\bibinfo{date}{July} \bibinfo{year}{2020}), \bibinfo{pages}{736--745}.
\newblock


\bibitem[Rahardja and Young(2021)]%
        {Rahardja2021}
\bibfield{author}{\bibinfo{person}{Dewi Rahardja} {and} \bibinfo{person}{Dean~M. Young}.} \bibinfo{year}{2021}\natexlab{}.
\newblock \showarticletitle{Confidence Intervals for the Risk Ratio Using Double Sampling with Misclassified Binomial Data}.
\newblock \bibinfo{journal}{\emph{Journal of Data Science}} \bibinfo{volume}{9}, \bibinfo{number}{4} (\bibinfo{year}{2021}), \bibinfo{pages}{529--548}.
\newblock


\bibitem[Rajkomar et~al\mbox{.}(2018)]%
        {Rajkomar2018}
\bibfield{author}{\bibinfo{person}{Alvin Rajkomar}, \bibinfo{person}{Michaela Hardt}, \bibinfo{person}{Michael~D. Howell}, \bibinfo{person}{Greg Corrado}, {and} \bibinfo{person}{Marshall~H. Chin}.} \bibinfo{year}{2018}\natexlab{}.
\newblock \showarticletitle{Ensuring Fairness in Machine Learning to Advance Health Equity}.
\newblock \bibinfo{journal}{\emph{Annals of Internal Medicine}} \bibinfo{volume}{169}, \bibinfo{number}{12} (\bibinfo{date}{Dec.} \bibinfo{year}{2018}), \bibinfo{pages}{866}.
\newblock


\bibitem[Repository(1994)]%
        {German}
\bibfield{author}{\bibinfo{person}{UCI Machine~Learning Repository}.} \bibinfo{year}{1994}\natexlab{}.
\newblock \showarticletitle{German Credit data}.
\newblock  (\bibinfo{year}{1994}).
\newblock
\urldef\tempurl%
\url{https://archive.ics.uci.edu/ml/datasets/statlog+(german+credit+data)}
\showURL{%
\tempurl}


\bibitem[Richardson et~al\mbox{.}(2019)]%
        {richardson2019dirty}
\bibfield{author}{\bibinfo{person}{Rashida Richardson}, \bibinfo{person}{Jason~M Schultz}, {and} \bibinfo{person}{Kate Crawford}.} \bibinfo{year}{2019}\natexlab{}.
\newblock \showarticletitle{Dirty data, bad predictions: How civil rights violations impact police data, predictive policing systems, and justice}.
\newblock \bibinfo{journal}{\emph{NYUL Rev. Online}}  \bibinfo{volume}{94} (\bibinfo{year}{2019}), \bibinfo{pages}{15}.
\newblock


\bibitem[Rubin(1976)]%
        {RUBIN1976}
\bibfield{author}{\bibinfo{person}{Donald~B. Rubin}.} \bibinfo{year}{1976}\natexlab{}.
\newblock \showarticletitle{Inference and missing data}.
\newblock \bibinfo{journal}{\emph{Biometrika}} \bibinfo{volume}{63}, \bibinfo{number}{3} (\bibinfo{year}{1976}), \bibinfo{pages}{581--592}.
\newblock


\bibitem[Sechidis et~al\mbox{.}(2017)]%
        {Sechidis2017}
\bibfield{author}{\bibinfo{person}{Konstantinos Sechidis}, \bibinfo{person}{Matthew Sperrin}, \bibinfo{person}{Emily~S. Petherick}, \bibinfo{person}{Mikel Luj{\'{a}}n}, {and} \bibinfo{person}{Gavin Brown}.} \bibinfo{year}{2017}\natexlab{}.
\newblock \showarticletitle{Dealing with under-reported variables: An information theoretic solution}.
\newblock \bibinfo{journal}{\emph{International Journal of Approximate Reasoning}}  \bibinfo{volume}{85} (\bibinfo{year}{2017}), \bibinfo{pages}{159--177}.
\newblock


\bibitem[Stoner et~al\mbox{.}(2019)]%
        {Stoner_2019}
\bibfield{author}{\bibinfo{person}{Oliver Stoner}, \bibinfo{person}{Theo Economou}, {and} \bibinfo{person}{Gabriela Drummond~Marques da Silva}.} \bibinfo{year}{2019}\natexlab{}.
\newblock \showarticletitle{A Hierarchical Framework for Correcting Under-Reporting in Count Data}.
\newblock \bibinfo{journal}{\emph{J. Amer. Statist. Assoc.}} \bibinfo{volume}{114}, \bibinfo{number}{528} (\bibinfo{date}{apr} \bibinfo{year}{2019}), \bibinfo{pages}{1481--1492}.
\newblock


\bibitem[Vaithianathan et~al\mbox{.}([n.\,d.])]%
        {hello-baby}
\bibfield{author}{\bibinfo{person}{Rhema Vaithianathan}, \bibinfo{person}{Diana Benavides-Prado}, {and} \bibinfo{person}{Emily Putnam-Hornstein}.} \bibinfo{year}{[n.\,d.]}\natexlab{}.
\newblock \bibinfo{title}{Hello baby methodology - alleghenycountyanalytics.us}.
\newblock
\newblock
\urldef\tempurl%
\url{https://www.alleghenycountyanalytics.us/wp-content/uploads/2020/12/Hello-Baby-Methodology-v6.pdf}
\showURL{%
\tempurl}


\bibitem[Vaithianathan et~al\mbox{.}(2017)]%
        {vaithianathan2017developing}
\bibfield{author}{\bibinfo{person}{Rhema Vaithianathan}, \bibinfo{person}{Emily Putnam-Hornstein}, \bibinfo{person}{Nan Jiang}, \bibinfo{person}{Parma Nand}, {and} \bibinfo{person}{Tim Maloney}.} \bibinfo{year}{2017}\natexlab{}.
\newblock \showarticletitle{Developing predictive models to support child maltreatment hotline screening decisions: Allegheny County methodology and implementation}.
\newblock \bibinfo{journal}{\emph{Center for Social data Analytics}} (\bibinfo{year}{2017}).
\newblock


\bibitem[Van~Bekkum and Borgesius(2021)]%
        {van2021digital}
\bibfield{author}{\bibinfo{person}{Marvin Van~Bekkum} {and} \bibinfo{person}{Frederik~Zuiderveen Borgesius}.} \bibinfo{year}{2021}\natexlab{}.
\newblock \showarticletitle{Digital welfare fraud detection and the Dutch SyRI judgment}.
\newblock \bibinfo{journal}{\emph{European Journal of Social Security}} \bibinfo{volume}{23}, \bibinfo{number}{4} (\bibinfo{year}{2021}), \bibinfo{pages}{323--340}.
\newblock


\bibitem[Wang and Singh(2021)]%
        {Wang2021}
\bibfield{author}{\bibinfo{person}{Yanchen Wang} {and} \bibinfo{person}{Lisa Singh}.} \bibinfo{year}{2021}\natexlab{}.
\newblock \showarticletitle{Analyzing the impact of missing values and selection bias on fairness}.
\newblock \bibinfo{journal}{\emph{International Journal of Data Science and Analytics}} \bibinfo{volume}{12}, \bibinfo{number}{2} (\bibinfo{date}{May} \bibinfo{year}{2021}), \bibinfo{pages}{101--119}.
\newblock


\bibitem[Wu et~al\mbox{.}(2023)]%
        {Zou23}
\bibfield{author}{\bibinfo{person}{Kevin Wu}, \bibinfo{person}{Dominik Dahlem}, \bibinfo{person}{Christopher Hane}, \bibinfo{person}{Eran Halperin}, {and} \bibinfo{person}{James Zou}.} \bibinfo{year}{2023}\natexlab{}.
\newblock \showarticletitle{Collecting data when missingness is unknown: a method for improving model performance given under-reporting in patient populations}. In \bibinfo{booktitle}{\emph{Proceedings of the Conference on Health, Inference, and Learning}}, Vol.~\bibinfo{volume}{209}. \bibinfo{publisher}{PMLR}, \bibinfo{pages}{229--242}.
\newblock


\bibitem[Zhang and Long(2021)]%
        {Zhang21Neurips}
\bibfield{author}{\bibinfo{person}{Yiliang Zhang} {and} \bibinfo{person}{Qi Long}.} \bibinfo{year}{2021}\natexlab{}.
\newblock \showarticletitle{Assessing Fairness in the Presence of Missing Data}. In \bibinfo{booktitle}{\emph{Advances in Neural Information Processing Systems (Neurips '21)}}.
\newblock


\bibitem[Zhou et~al\mbox{.}(2022)]%
        {zhou22}
\bibfield{author}{\bibinfo{person}{Helen Zhou}, \bibinfo{person}{Balakrishnan Sivaraman}, {and} \bibinfo{person}{Zachary~C. Lipton}.} \bibinfo{year}{2022}\natexlab{}.
\newblock \showarticletitle{Domain Adaptation Under Missingness Shift}.
\newblock \bibinfo{journal}{\emph{arXiv preprint, arXiv:2211.02093}} (\bibinfo{year}{2022}).
\newblock


\end{thebibliography}

\appendix
\counterwithin{theorem}{section}
\setcounter{figure}{0}
\setcounter{table}{0}
\renewcommand\thefigure{\thesection.\arabic{figure}} 
\renewcommand\thetable{\thesection.\arabic{table}} 

\section{Summary of assumptions}
\label{app:summary_assumptions}

\begin{table}[H]
    \centering
    \renewcommand{\arraystretch}{1.5}
    \footnotesize{
    \begin{tabular}{|P{2cm}|P{1.5cm}|P{1.5cm}|P{1.5cm}|P{1.5cm}|P{1.5cm}|P{1.6cm}|}
        \hline
        \textbf{Result / Assumption} & \textbf{Missingness indicators} & \textbf{Default value $m=0$} &  \textbf{Linear ground truth} & \textbf{$G\perp Z$, i.e. MCAR} & \textbf{$Z_{[2:d]}$ uncorrelated} & \textbf{$Z\sim \mathcal{N}$ jointly Gaussian}\\
        \hline
        Lemma~\ref{lemma:attenutationbiasonedim} &&x&x&x&&\\
        \hline
        Proposition~\ref{prop:propertieshatbeta1} \& \ref{prop:propertieshatbetak}&&x&x&x&x&\\
        \hline  
        Proposition~\ref{prop:whobenefits_short} \& \ref{prop:whobenefits}&&x&&x&&x\\
        \hline  
        Corollary~\ref{cor:maincorollary}&&x&x&x&x&x\\
        \hline  
        Lemma~\ref{lem:augmented_loss} \& \ref{lem:group_dep_augmented_loss}&&x&&(x)&&\\
        \hline  
        Lemma~\ref{lem:optimal_prediction_imp_value} \& \ref{lem:group_optimam_prediction_imp_value}&&x&x&(x)&&\\
        \hline
        Under-reporting rate estimation&&x&&(x)&&\\
        \hline
    \end{tabular}
    }
    \caption{Summary of assumptions. Rows represent paper segments, columns indicate sufficient assumptions for corresponding findings.}
\end{table}

\section{Additional theorems}
\begin{proposition}
\label{prop:params}
In the $d$-dimensional case setting described in Section~\ref{sec:linreg}, the parameter estimates from Equation~\ref{eq:paramestimates} take the form
\begin{align}
\begin{split}
\label{eq:param_estim_bias}
    \hat{\beta}_1 &= \beta_1 \frac{1}{1-R^2}\sqrt{\frac{\V{Z_1}}{\V{X_1}}}\left(\rho(X_1,Z_1)-\sum_{i=2}^d\rho(X_1,Z_i)\rho(Z_1,Z_i)\right),\\
    \hat{\beta}_k &
    =\beta_1\sqrt{\frac{\V{Z_1}}{\V{Z_k}}}\left(\rho(Z_k,Z_1)-\frac{1}{1-R^2}\rho(X_1,Z_k)\left(\rho(X_1,Z_1)-\sum_{i=2}^d\rho(X_1,Z_i)\rho(Z_1,Z_i)\right)\right) + \beta_k\\
\end{split}
\end{align}
for $k\in[2:d]$. Here, $R^2=\sum_{i=2}^d\rho(X_1,Z_i)^2\in[0,1)$.
\end{proposition}

Here, $R^2$ is the squared coefficient of multiple correlation between $Z_1\xi_1$ and $Z_{[2:d]}=[Z_2,\ldots,Z_d]$ which can be interpreted as the fraction of variance in $Z_1\xi_1$ that can be explained by the independent variables $Z_{[2:d]}$.
If all features are observed, the factor $\rho(X_1,Z_1)-\sum_{i=2}^d\rho(X_1,Z_i)\rho(Z_1,Z_i)$ collapses to $(1-R^2)$, and the estimates are unbiased.
With under-reporting the bias introduced into the parameter estimates depends on the strength of correlations between features, as well as how this correlation changes with the mismeasurement of $Z_1$.
The bias in Equation~\ref{eq:param_estim_bias} can be conceptualized as a generalization of omitted variable bias \cite{Angrist2008-oo} which is further explored in Appendix~\ref{app:omitted_variable_bias}.

\begin{proposition}
\label{prop:whobenefits}
Define the threshold turning point $T$ as 
$$
    T=\hat{\alpha} + \hat{\beta}_{[2:d]}^T\mu_{[2:d]} + \frac{\text{sd}\left(\hat{\beta}^T_{[2:d]}Z_{[2:d]}\right)}{\text{sd}\left(\hat{\beta}^T_{[2:d]}Z_{[2:d]}\right)-\text{sd}\left(\hat{\beta}^TZ\right)}\hat{\beta}_1\mu_1.
$$
Then, for a high threshold $\tilde{y}$ with $\tilde{y}>T$, the group with more under-reporting is
\begin{itemize}[topsep=2ex,itemsep=1ex]
    \item \textbf{Case 1:} Over-selected if $\V{\hat{\alpha}+\hat{\beta}_{[2:d]}Z_{[2:d]}}>\V{\hat{\alpha}+\hat{\beta}Z}$, or
    \item \textbf{Case 2:} Under-selected if $\V{\hat{\alpha}+\hat{\beta}_{[2:d]}Z_{[2:d]}}<\V{\hat{\alpha}+\hat{\beta}Z}$.
\end{itemize}
For low thresholds $\tilde{y}<T$, the cases are reversed.
\end{proposition}

In practical applications, thresholds are usually set such that only a small portion of predictions exceeds the threshold. For example, we can only decide to screen a small portion of calls in the child welfare setting. In particular, realistic thresholds are generally well above the average $\hat{Y}$. On a high level, the turning point $T$ in Proposition~\ref{prop:whobenefits} represent an adjusted mean predicted value where the influence of the feature with under-reporting is weighed depending on a ratio determined by prediction variances with and without the feature. 

\begin{lemma}[Group-dependent augmented loss]
    \label{lem:group_dep_augmented_loss}
    Assume fixed $f\in\mathcal{F},z\in\mathcal{Z},y\in\mathcal{Y},g\in\{0,1\}$ and $X\in\R^d$ defined by $X_1=Z_1\xi_1^g$ and $X_{[2:d]}=z_{[2:d]}$. Define
    \begin{align*}
        \tilde{l}(f,X,y,g) &:=\frac{1}{m_1^g}l(f,X,y) - \frac{1-m_1^g}{m_1^g}l(f,[0,X_{[2:d]}]^T,y)
    \end{align*}
    Then, we have that $\mathds{E}_{\xi_1^g}\left[\tilde{l}(f,X,y,g)\right]=l(f,z,y)$.
\end{lemma}

\begin{lemma}[Group-dependent optimal prediction imputation values]
    \label{lem:group_optimam_prediction_imp_value}
    Assume $f(Z)=\alpha + \beta^TZ$ is the ground truth model, $X$ a random vector of observed features, and $G$ the group membership. We set
    \begin{align*}
        X'=\begin{cases}
            X\text{ if } X_1\neq 0,\\
            [x_1'^0,X_{[2:d]}] \text{ if } X_1=0 \text{ and } G=0,\\
            [x_1'^1,X_{[2:d]}] \text{ if } X_1=0 \text{ and } G=1,\\
        \end{cases}
    \end{align*}
    where $x_1'^0,x_1'^1$ are group-dependent fixed imputation values.
    Then, 
    $$
        \arg\min_{x_1'^g}\mathds{E}_X\left[(f(X')-Y)^2\right] = \E{Z_1\mid X_1=0, G=g}
    $$
    are the optimal group-dependent prediction imputation values for $g\in\{0,1\}$.
\end{lemma}
Similar to before, the optimal imputation values can be written as
$$
    \E{Z_1\mid X_1 = 0,G=g} = \frac{\frac{1}{m_1^g}\E{X_1,G=g}-P(X_1\neq 0\mid G=g)\E{X_1\mid X_1\neq 0,G=g}}{P(X_1=0\mid G=g)},
$$
which can be estimated directly from observed data.

\section{Connection between Proposition~\ref{prop:params} and omitted variable bias}
\label{app:omitted_variable_bias}

Econometrics literature uses the term omitted variable bias to refer to the model estimation bias that is introduced when omitting an independent variable that influences both other independent variables and the dependent outcome \cite{Angrist2008-oo}. In the setting of Proposition~\ref{prop:params}, omitting the first feature entirely corresponds to a setting in which all feature entries are under-reported, i.e. default to 0. 
The $k$-th parameter estimate in this case can be written as
$$
    \hat{\beta}_k = \beta_1 \frac{\Cov{Z_k}{Z_1}}{\V{Z_k}} + \beta_k
$$
which is known as omitted variable bias formula \cite{Angrist2008-oo}.
Here, $\gamma_{Z_1,Z_k}=\Cov{Z_1}{Z_k}/\V{Z_k}$ corresponds to the population regression coefficient of a linear regression of $Z_1$ on $Z_k$ which can be written as 
$$
    Z_1=\alpha_{Z_1,Z_k}+\gamma_{Z_1,Z_k}Z_k,
$$
where $\alpha_{Z_1,Z_k}$ is an intercept. Omitting $Z_1$ from the regression induces a confounding relationship where the effects of $Z_1$ on $Z_k$ become intertwined. Instead of isolating the effect of $Z_k$ on $Y$, $\hat{\beta}_k$ also includes a partial effect of $Z_1$ on $Y$. This effect is scaled by $\gamma_{Z_1,Z_k}$ to account for the linear relationship between $Z_1$ and $Z_k$.

In the setting of this paper, we are interested in cases in which some but not necessarily all of the feature entries are missing. Maintaining the same notation as before, $\hat{\beta}_1$ from Equation~\ref{eq:param_estim_bias} in this general case can be written as
\begin{align*}
   \hat{\beta_1} &= \beta_1 \frac{1}{1-R^2} \left(\frac{\Cov{X_1}{Z_1}}{\V{X_1}}-\sum_{i=2}^d\frac{\Cov{X_1}{Z_i}\Cov{Z_1}{Z_i}}{\V{X_1}\V{Z_i}}\right)\\
   &=\beta_1 \left(\frac{\gamma_{Z_1,X_1}-\sum_{i=2}^d\gamma_{Z_i,X_1}\gamma_{Z_1,Z_i}}{1-R^2}\right).
\end{align*}
Here, the numerator of the biasing factor reflects how much information about $Z_1$ remains encoded in $X_1$ without drawing on associations through the other features $Z_2,\ldots,Z_d$ (i.e, arrows of the form $X_1\to Z_i\to Z_1$).
The denominator measures how much of the variance in $X_1$ is explained by $Z_2,\ldots,Z_d$.
For $k\in[2:d]$, we receive
\begin{align*}
    \hat{\beta}_k &= \beta_k + \beta_1 \frac{\Cov{Z_k}{Z_1}}{\V{Z_k}}-\beta_1\frac{1}{1-R^2}\frac{\Cov{X_1}{Z_k}}{\V{Z_k}}\left(\frac{\Cov{X_1}{Z_1}}{\V{X_1}}-\sum_{i=2}^d\frac{\Cov{X_1}{Z_i}\Cov{Z_1}{Z_i}}{\V{Z_i}\V{X_1}}\right)\\
    &=\beta_k + \beta_1 \gamma_{Z_1,Z_k}-\beta_1\frac{1}{1-R^2}\gamma_{X_1,Z_k}\left(\gamma_{Z_1,X_1}-\sum_{i=2}^d\gamma_{Z_i,X_1}\gamma_{Z_1,Z_i}\right)\\
    &=\beta_k + \underbrace{\beta_1 \gamma_{Z_1,Z_k}}_{\text{omitted variable bias}}-\underbrace{\hat{\beta}_1\gamma_{X_1,Z_k}.}_{\text{Correction since partially observed}}
\end{align*}
Instead of just encoding the effect of $Z_k$ on $Y$ and partial effect of $Z_1$ on $Y$ like before, the estimate $\hat{\beta}_k$ now also corrects for the fact that $Z_1$ is partially observed. The magnitude of the correction depends on the parameter estimate for the partially observed variable as well as the linear relationship between $X_1$ and $Z_k$.

\section{Proofs}
In this section, we provide the full proofs for the results in the main text.

\paragraph{Lemma~\ref{lemma:attenutationbiasonedim}} We have $\xi\perp Z$ and $\E{\xi^2}=\E{\xi}$. Since $\E{\xi}\in[0,1]$, we have 

\begin{align*}
    \mid\hat{\beta}\mid
    = \mid\frac{\Cov{X}{Z}}{\Cov{X}{X}}\beta\mid
    = \mid\frac{\E{\xi Z^2}-\E{\xi Z}\E{Z}}{\E{\xi^2 Z^2}-\E{\xi Z}^2}\beta\mid
    = \frac{\E{\xi}\V{Z}}{\E{\xi}(\E{Z^2}-\E{\xi}\E{Z}^2)}\mid\beta\mid \leq \mid\beta\mid.
\end{align*}

\paragraph{Proposition~\ref{prop:params}}
In order to derive the equations for $\hat{\beta}_i$ for $i\in[1:d]$, we start by inverting the covariance matrix
\begin{align*}
    \Sigma_X =
    \begin{pmatrix}
    \V{Z_1\xi_1} & \Cov{Z_1\xi_1}{Z_2} & \Cov{Z_1\xi_1}{Z_3}&\cdots& \Cov{Z_1\xi_1}{Z_d}\\
    \Cov{Z_2}{Z_1\xi_1}&\V{Z_2}&\Cov{Z_2}{Z_3}&\cdots&\Cov{Z_2}{Z_d}\\
    \Cov{Z_3}{Z_1\xi_1}&\Cov{Z_3}{Z_2}&\V{Z_3}&\cdots&\Cov{Z_3}{Z_d}\\
    \vdots &&& \ddots &\vdots\\
    \Cov{Z_d}{Z_1\xi_1} & \Cov{Z_d}{Z_2}&\cdots &&\V{Z_d}
    \end{pmatrix}.
\end{align*}
For this, we separate the matrix into the blocks $A = (\Sigma_X)_{11}$, $B = ((\Sigma_X)_{1j})_{j\in[2:d]}$,  $C = ((\Sigma_X)_{i1})_{i\in[2:d]}$, and  $D = ((\Sigma_X)_{ij})_{i,j\in[2:d]}$.
Note that $A\in\R^{1\times 1}$, $B=C^T\in\R^{1\times (d-1)}$, and $D\in\R^{(d-1)\times (d-1)}$. Using matrix inversion theorem, the inverse of $\Sigma_X$ can be written as
\begin{align}
\begin{split}
\label{eq:appendixschur}
    (\Sigma_X)^{-1}=
    \begin{pmatrix}
        A^{-1} + A^{-1}B(D-CA^{-1}B)^{-1}CA^{-1} & -A^{-1}B(D-CA^{-1}B)^{-1}\\
        -(D-CA^{-1}B)^{-1}CA^{-1} & (D-CA^{-1}B)^{-1}
    \end{pmatrix},
\end{split}
\end{align}
where $D-CA^{-1}B=D-CA^{-1}C^T$ is the Schur complement of $A$ in $\Sigma_{X}$.
Recall that, by assumption, $\Cov{Z_i}{Z_j}=0$ for $i,j>1$ with $i\neq j$ which means that $D$ is a diagonal matrix. We also note that $\text{rank}(CC^T)=1$. Denoting $g=\text{trace}(-A^{-1}CC^TD^{-1})$, the inverse of the Schur complement can be written as 
\begin{align*}
    (D-CA^{-1}C^T)^{-1} = (D-A^{-1}CC^T)^{-1} = D^{-1}+\frac{1}{1+g}D^{-1}A^{-1}CC^TD^{-1}.
\end{align*}
Here, $D^{-1}$ is a diagonal matrix with values $1/\V{Z_i}$ for $i\in[2:d]$ on the diagonal, $A^{-1}=1/\V{Z_1\xi_1}$, and the diagonal of $CC^T$ is $\Cov{Z_1\xi_1}{Z_{i}}^2$ for $i\in[2:d]$.
It follows that
\begin{align*}
    g = -\sum_{i=2}^d \rho(Z_1\xi_1,Z_i)^2
\end{align*}
which corresponds to the negative of the $R^2$ between $Z_1\xi$ and $Z_2,\ldots,Z_d$. We hence write $R^2$ for $-g$ in the following.

Now we can calculate the top left block of the inverse matrix in Equation~\ref{eq:appendixschur} as
\begin{align*}
    A^{-1} + A^{-1}B(D-CA^{-1}B)^{-1}CA^{-1} = \frac{1}{\V{Z_1\xi_1}}\frac{1}{1-R^2}.
\end{align*}
The top right bock corresponds to the row vector
\begin{align*}
    -AB(D-CA^{-1}B)^{-1} = \left(\left(-\frac{\Cov{Z_1\xi_1}{Z_i}}{\V{Z_i}\V{Z_1\xi_1}}\frac{1}{1-R^2}\right)_i\right)_{i\in [2:d]},
\end{align*}
while the bottom left block is the same transposed. 
Lastly, the bottom left block of Equation~\ref{eq:appendixschur} can be computed as
\begin{align*}
    (D-CA^{-1}B)^{-1}= \left(\text{diag}(1/\V{Z_i}) + \frac{1}{1-R^2}\left(\frac{\Cov{Z_1\xi_1}{Z_i}\Cov{Z_1\xi_1}{Z_j}}{\V{Z_1\xi_1}\V{Z_i}\V{Z_j}}\right)_{ij}\right)_{i,j\in[2:d]}.
\end{align*}

Inserting these values into Equation~\ref{eq:paramestimates} yields the desired parameter estimates
\begin{align*}
    \hat{\beta}_1 &= \beta_1 \frac{1}{1-R^2}\sqrt{\frac{\V{Z_1}}{\V{X_1}}}\left(\rho(X_1,Z_1)-\sum_{i=2}^d\rho(X_1,Z_i)\rho(Z_1,Z_i)\right),
\end{align*}
and
\begin{align*}
    \hat{\beta}_k&=\beta_1\sqrt{\frac{\V{Z_1}}{\V{Z_k}}}\left(\rho(Z_k,Z_1)-\frac{1}{1-R^2}\rho(X_1,Z_k)\left(\rho(X_1,Z_1)-\sum_{i=2}^d\rho(X_1,Z_i)\rho(Z_1,Z_i)\right)\right) + \beta_k.
\end{align*}

\paragraph{Proposition~\ref{prop:propertieshatbeta1}}
The estimates from Proposition~\ref{prop:params} simplify to
\begin{align*}
    \hat{\beta}_1 = \frac{1}{1-R^2}\left(\frac{\E{\xi_1}\V{Z_1}}{\V{Z_1\xi_1}}-\frac{R^2}{\E{\xi_1}}\right)\beta_1
\end{align*}
and
\begin{align*}
    \hat{\beta}_k = \beta_1 \frac{1}{1-R^2}\left(\frac{\Cov{Z_1}{Z_k}(\V{Z_1\xi_1}-\V{Z_1}\E{\xi_1}^2)}{\V{Z_k}\V{Z_1\xi_1}}\right) + \beta_k,
\end{align*}
for $k\in[2:d]$.

For the first claim, recall that $\V{\xi}=\E{\xi_1}-\E{\xi_1}^2$ and $\V{Z_1\xi_1}=\V{Z_1}\E{\xi_1}^2+\V{\xi_1}\E{Z_1^2}$.
Note that $Z\perp \xi$ allows us to rewrite
\begin{align*}
    R^2&=\sum_{i=2}^{d}\rho(Z_1\xi_1,Z_i)^2\\
    &=\sum_{i=2}^{d}\frac{\E{\xi_1}^2\Cov{Z_1}{Z_i}^2\V{Z_1}}{\V{Z_1\xi_1}\V{Z_i}\V{Z_1}}\\
    &=\frac{\E{\xi_1}^2\V{Z_1}}{\V{Z_1\xi_1}}\sum_{i=2}^{d}\rho(Z_1,Z_i)^2\\
\end{align*}
and thus 
\begin{align*}
    &\frac{1}{1-R^2}\left(\frac{\E{\xi_1}\V{Z_1}}{\V{Z_1\xi_1}}-\frac{R^2}{\E{\xi_1}}\right)\\
    =&\frac{\E{\xi_1}\V{Z_1}\E{\xi_1}-R^2\V{Z_1\xi_1}}{(1-R^2)\V{Z_1\xi_1}\E{\xi_1}}\\
    =&\frac{\E{\xi_1}\V{Z_1}\E{\xi_1}-\frac{\E{\xi_1}^2\V{Z_1}}{\V{Z_1\xi_1}}\sum_{i=2}^{d}\rho(Z_1,Z_i)^2\V{Z_1\xi_1}}{(1-\frac{\E{\xi_1}^2\V{Z_1}}{\V{Z_1\xi_1}}\sum_{i=2}^{d}\rho(Z_1,Z_i)^2)\V{Z_1\xi_1}\E{\xi_1}}\\
    =&\frac{\E{\xi_1}\V{Z_1}\E{\xi_1}-\E{\xi_1}^2\V{Z_1}\sum_{i=2}^{d}\rho(Z_1,Z_i)^2}{(\V{Z_1}\E{\xi_1}^2+(\E{\xi_1}-\E{\xi_1}^2)\E{Z_1^2})\E{\xi_1}-\E{\xi_1}^2\V{Z_1}\sum_{i=2}^{d}\rho(Z_1,Z_i)^2\E{\xi_1}}\\
    =&\frac{\V{Z_1}(1-\sum_{i=2}^{d}\rho(Z_1,Z_i)^2)}{\V{Z_1}\E{\xi_1}(1-\sum_{i=2}^{d}\rho(Z_1,Z_i)^2)+(1-\E{\xi_1})\E{Z_1^2}},\\
\end{align*}
which is positive as long as $Z_1$ is not a linear combination of other features which was explicitly excluded from consideration. The claim follows. 

For the second claim, we show that
\begin{align*}
    &\frac{\V{Z_1}(1-\sum_{i=2}^{d}\rho(Z_1,Z_i)^2)}{\V{Z_1}\E{\xi_1}(1-\sum_{i=2}^{d}\rho(Z_1,Z_i)^2)+(1-\E{\xi_1})\E{Z_1^2}} <1\\
    \Leftrightarrow &(1-\sum_{i=2}^{d}\rho(Z_1,Z_i)^2)\V{Z_1}(1-\E{\xi_1}) <(1-\E{\xi_1})\E{Z_1^2}\\
    \Leftrightarrow &(1-\sum_{i=2}^{d}\rho(Z_1,Z_i)^2)(\E{Z_1^2}-\E{Z_1}^2) <\E{Z_1^2}\\
    \Leftrightarrow &(1-\sum_{i=2}^{d}\rho(Z_1,Z_i)^2)\left(1-\frac{\E{Z_1}^2}{\E{Z_1}^2}\right) <1.
\end{align*}
Since $Z_1$ is not a linear combination of other features, we know that $1-\sum_{i=2}^{d}\rho(Z_1,Z_i)^2\in(0,1]$ and this inequatily is always true. The claim follows with the first part of the proposition.

For the third claim, recall that $\E{\xi_1}=rm^1_1+(1-r)m^0_1$. Assume we have two sets of parameters $(m_1^0,m_1^1)$ and $(m_1^{0'},m_1^{1'})$. If $m_1^0<m_1^{0'}$ and $m_1^1\leq m_1^{1'}$, we have
$$
\E{\xi} = rm^1_1+(1-r)m^0_1 < rm^{1'}_1+(1-r)m^{0'}_1 = \E{\xi'}.
$$
The same holds true if $m_1^0\leq m_1^{0'}$ and $m_1^1< m_1^{1'}$ which shows that the expected share of observed features $\E{\xi_1}$ is decreasing if and only if we are increasing under-reporting in either (or both) of the groups while leaving everything else fixed. Thus, instead of changes in $m_1^g$, we argue directly about changes in $\E{\xi_1}$ in the following. 

Denote $S^2=\sum_{i=2}^{d}\rho(Z_1,Z_i)^2$ and consider the function
\begin{align*}
    f:(0,1]&\to \R\\
    \E{\xi_1}=x&\mapsto \frac{\V{Z_1}(1-S^2)}{\V{Z_1}x(1-S^2)+(1-x)\E{Z_1^2}}.
\end{align*}
We show that $f$ is monotonically increasing from which the claim follows directly. 
It holds that
\begin{align*}
    \frac{d}{dx}f(x) = \frac{-\V{Z_1}(1-S^2)(\V{Z_1}(1-S^2)-\E{Z_1^2})}{\left(\V{Z_1}x(1-S^2)+(1-x)\E{Z_1^2}\right)^2}.
\end{align*}
Since $1-S^2\in[0,1)$ and $\V{Z_1}>0$, the numerator is positive iff 
\begin{align*}
    &-\V{Z_1}(1-S^2)(\V{Z_1}(1-S^2)-\E{Z_1^2})>0\\
    \Leftrightarrow &\V{Z_1}(1-S^2) < \E{Z_1^2}\\
    \Leftrightarrow &\left(1-\frac{\E{Z_1}^2}{\E{Z_1^2}}\right)(1-S^2) < 1,
\end{align*}
which is a true statement. We conclude that $f$ is monotonically increasing in $x$ and the claim follows.

\paragraph{Proposition~\ref{prop:propertieshatbetak}}
Given a $k\in[2:d]$, we know that
\begin{align*}
    \hat{\beta}_k = \beta_1 \frac{1}{1-R^2}\left(\frac{\Cov{Z_1}{Z_k}(\V{Z_1\xi_1}-\V{Z_1}\E{\xi_1}^2)}{\V{Z_k}\V{Z_1\xi_1}}\right) + \beta_k.
\end{align*}
The first claim is obvious from this expression. 

For the second claim, we follow similar steps as for the third claim in the proof of Proposition~\ref{prop:propertieshatbeta1}.
We know from the previous proof that
\begin{align*}
    R^2 = \frac{\E{\xi_1}^2\V{Z_1}}{\V{Z_1\xi_1}} \sum_{i=2}^d\rho(Z_1,Z_i)^2
\end{align*}
and thus, denoting $S^2 = \sum_{i=2}^d\rho(Z_1,Z_i)^2$,
\begin{align*}
    &\frac{1}{1-R^2}\left(\frac{\Cov{Z_1}{Z_k}(\V{Z_1\xi_1}-\V{Z_2}\E{\xi}^2)}{\V{Z_k}\V{Z_1\xi_1}}\right)\\
    =&\frac{\Cov{Z_1}{Z_k}\V{\xi_1}\E{Z_1^2}}{(1-\frac{\E{\xi_1}^2\V{Z_1}}{\V{Z_1\xi_1}}S^2)\V{Z_k}\V{Z_1\xi_1}}\\
    =&\frac{\Cov{Z_1}{Z_k}\V{\xi_1}\E{Z_1^2}}{\V{Z_k}\V{Z_1\xi_1}-\E{\xi_1}^2\V{Z_1}S^2\V{Z_k}}\\
    =&\frac{\Cov{Z_1}{Z_k}(\E{\xi_1}-\E{\xi_1}^2)\E{Z_1^2}}{\V{Z_k}(\V{Z_1}\E{\xi_1}^2+(\E{\xi_1}-\E{\xi_1}^2)\E{Z_1^2})-\E{\xi_1}^2\V{Z_1}S^2\V{Z_k}}\\
    =&\frac{\Cov{Z_1}{Z_k}(1-\E{\xi_1})\E{Z_1^2}}{(1-S^2)\V{Z_k}\V{Z_1}\E{\xi_1}+\V{Z_k}(1-\E{\xi_1})\E{Z_1^2}},
\end{align*}
since $\V{Z_1\xi_1}=\V{Z_1}\E{\xi_1}^2+\V{\xi_1}\E{Z_1^2}$ and $\V{\xi_1}=\E{\xi_1}-\E{\xi_1}^2$.

Now, consider the function
\begin{align*}
    g:(0,1]&\to \R\\
    \E{\xi_1}=x&\mapsto \frac{\Cov{Z_1}{Z_k}(1-x)\E{Z_1^2}}{(1-S^2)\V{Z_k}\V{Z_1}x+\V{Z_k}(1-x)\E{Z_1^2}}.
\end{align*}
We compute
\begin{align*}
    \frac{d}{dx}g(x) = &\frac{-\Cov{Z_1}{Z_k}\E{Z_1^2}((1-S^2)\V{Z_k}\V{Z_1}x+\V{Z_k}(1-x)\E{Z_1^2})}{\left((1-S^2)\V{Z_k}\V{Z_1}x+\V{Z_k}(1-x)\E{Z_1^2}\right)^2}\\
    &-\frac{(1-x)\Cov{Z_1}{Z_k}\E{Z_1^2}((1-S^2)\V{Z_k}
    \V{Z_1} - \V{Z_k}\E{Z_1^2})}{\left((1-S^2)\V{Z_k}\V{Z_1}x+\V{Z_k}(1-x)\E{Z_1^2}\right)^2}.
\end{align*}
Further, 
\begin{align*}
    &\frac{d}{dx}g(x) > 0 \\
    \Leftrightarrow & -\Cov{Z_1}{Z_k}\E{Z_1^2}((1-S^2)\V{Z_k}\V{Z_1}x+\V{Z_k}(1-x)\E{Z_1^2})\\& >  (1-x)\Cov{Z_1}{Z_k}\E{Z_1^2}((1-S^2)\V{Z_k}
    \V{Z_1} - \V{Z_k}\E{Z_1^2})\\
    \Leftrightarrow & -\Cov{Z_1}{Z_k}(1-S^2)\V{Z_k}\V{Z_1}x >  (1-x)\Cov{Z_1}{Z_k}(1-S^2)\V{Z_k}
    \V{Z_1}\\
    \Leftrightarrow & 0 >  \Cov{Z_1}{Z_k}.
\end{align*}
This shows that factor determining the influence of $\beta_1$ on $\hat{\beta}_k$ is increasing with decreasing under-reporting if $\Cov{Z_1}{Z_k}<0$ and decreasing with decreasing under-reporting otherwise. The claim follows. 

\paragraph{Proposition~\ref{prop:whobenefits}}
Predictions are obtained from the model $\hat{Y}=\hat{\alpha}+\hat{\beta}^TX$. Since $Z\sim\mathcal{N}(\mu,\Sigma)$ is jointly Gaussian, we know that
\begin{align*}
    \hat{\beta}^TZ\sim\mathcal{N}\left(\hat{\beta}^T\mu,\hat{\beta}^T\Sigma \hat{\beta}\right)=\mathcal{N}\left(\sum_{i=1}^d\hat{\beta}_i\mu_i,\sum_{i=1}^d\hat{\beta}_i^2\sigma_i^2 + 
    \sum_{i=1}^d\sum_{j=i+1}^d2\hat{\beta}_i\hat{\beta}_j\Cov{Z_i}{Z_j}\right)
\end{align*}
and
\begin{align*}
    \hat{\beta}_{[2:d]}^TZ_{[2:d]}\sim\mathcal{N}\left(\hat{\beta}_{[2:d]}^T\mu_{[2:d]},\hat{\beta}_{[2:d]}^T\Sigma_{[2:d,2:d]} \hat{\beta}_{[2:d]}\right)=\mathcal{N}\left(\sum_{i=2}^d\hat{\beta}_i\mu_i,\sum_{i=2}^d\hat{\beta}_i^2\sigma_i^2 + 
    \sum_{i=2}^d\sum_{j=i+1}^d2\hat{\beta}_i\hat{\beta}_j\Cov{Z_i}{Z_j}\right)
\end{align*}
where $\sigma_i^2 = \V{Z_i}$ for $i\in[1:d]$.

The cdf of predictions $\hat{Y}$ in group $g\in\{0,1\}$ can be written as
\begin{align*}
    F_{\hat{Y}\mid G=g}(x) = &P\left(\hat{\beta}_1Z_1\xi_1^g+\hat{\beta}_{[2:d]}^TZ_{[2:d]}\leq x-\hat{\alpha}\right)\\
    =&(1-m_1^g)P\left(\hat{\beta}_{[2:d]}^TZ_{[2:d]}\leq x-\hat{\alpha}\right) + m_1^gP\left(\hat{\beta}^TZ\leq x-\hat{\alpha}\right).
\end{align*}
Let $C\in[0,1]$ and denote $\tilde{y}=F_{\hat{Y}}^{-1}(1-C)$. Without loss of generality, assume that $m_1^0<m_1^1$. If $m_1^0=m_1^1$ the selection rate disparity is 0, if $m_1^0>m_1^1$ the following calculation can easily be adjusted. The inequality $m_1^0<m_1^1$ means that group 0 has the same or more expected under-reporting in feature $Z_1$ than group 1. Group 0 is over-selected at threshold $C$ according to Definition~\ref{def:overselectionindependent} if and only if
\begin{align*}
    &1-F_{\hat{Y}\mid G=1}(\tilde{y}) < 1-F_{\hat{Y}\mid G=0}(\tilde{y})\\ 
    \Leftrightarrow & (1-m_1^1)P\left(\hat{\beta}_{[2:d]}^TZ_{[2:d]}\leq \tilde{y}-\hat{\alpha}\right) + m_1^1P\left(\hat{\beta}^TZ\leq \tilde{y}-\hat{\alpha}\right) > (1-m_1^0)P\left(\hat{\beta}_{[2:d]}^TZ_{[2:d]}\leq \tilde{y}-\hat{\alpha}\right) + m_1^0P\left(\hat{\beta}^TZ\leq \tilde{y}-\hat{\alpha}\right)\\
    \Leftrightarrow & (m_1^1-m_1^0)P\left(\hat{\beta}^TZ\leq \tilde{y}-\hat{\alpha}\right) > (m_1^1-m^0_1)P\left(\hat{\beta}_{[2:d]}^TZ_{[2:d]}\leq \tilde{y}-\hat{\alpha}\right)\\
    \Leftrightarrow & P\left(\hat{\beta}^TZ\leq \tilde{y}-\hat{\alpha}\right) > P\left(\hat{\beta}_{[2:d]}^TZ_{[2:d]}\leq \tilde{y}-\hat{\alpha}\right).
\end{align*}

Expanding on this in the jointly Gaussian case, we can see that
\begin{align*}
    & P\left(\hat{\beta}^TZ\leq \tilde{y}-\hat{\alpha}\right) > P\left(\hat{\beta}_{[2:d]}^TZ_{[2:d]}\leq \tilde{y}-\hat{\alpha}\right)\\
    \Leftrightarrow & \Phi\left(\frac{\tilde{y}-\hat{\alpha}-\hat{\beta}^T\mu}{\sqrt{\hat{\beta}^T\Sigma\hat{\beta}}}\right) > \Phi\left(\frac{\tilde{y}-\hat{\alpha}-\hat{\beta}_{[2:d]}^T\mu_{[2:d]}}{\sqrt{\hat{\beta}_{[2:d]}^T\Sigma_{[2:d,2:d]}\hat{\beta}_{[2:d]}}}\right)\\
    \Leftrightarrow &\left(\tilde{y}-\hat{\alpha}-\hat{\beta}^T\mu\right)\sqrt{\hat{\beta}_{[2:d]}^T\Sigma_{[2:d,2:d]}\hat{\beta}_{[2:d]}}> \left(\tilde{y}-\hat{\alpha}-\hat{\beta}_{[2:d]}^T\mu_{[2:d]}\right)\sqrt{\hat{\beta}^T\Sigma\hat{\beta}}\\
    \Leftrightarrow &\tilde{y}\left(\sqrt{\hat{\beta}_{[2:d]}^T\Sigma_{[2:d,2:d]}\hat{\beta}_{[2:d]}}-\sqrt{\hat{\beta}^T\Sigma\hat{\beta}}\right)>\left(\hat{\alpha}+\hat{\beta}_{[2:d]}^T\mu_{[2:d]}\right)\left(\sqrt{\hat{\beta}_{[2:d]}^T\Sigma_{[2:d,2:d]}\hat{\beta}_{[2:d]}}-\sqrt{\hat{\beta}^T\Sigma\hat{\beta}}\right) +\left(\hat{\beta}_1\mu_1\right)\sqrt{\hat{\beta}_{[2:d]}^T\Sigma_{[2:d,2:d]}\hat{\beta}_{[2:d]}}.
\end{align*}
Here, $\Phi$ is the standard normal cdf.
If 
$$
\sqrt{\hat{\beta}_{[2:d]}^T\Sigma_{[2:d,2:d]}\hat{\beta}_{[2:d]}}-\sqrt{\hat{\beta}^T\Sigma\hat{\beta}}>0,
$$
group 0 is over-selected if and only if $C$ implies a threshold $\tilde{y}$ with
$$
\tilde{y}>\left(\hat{\alpha}+\hat{\beta}_{[2:d]}^T\mu_{[2:d]}\right)+\left(\hat{\beta}_1\mu_1\right)\frac{\sqrt{\hat{\beta}_{[2:d]}^T\Sigma_{[2:d,2:d]}\hat{\beta}_{[2:d]}}}{\sqrt{\hat{\beta}_{[2:d]}^T\Sigma_{[2:d,2:d]}\hat{\beta}_{[2:d]}}-\sqrt{\hat{\beta}^T\Sigma\hat{\beta}}}.
$$
If 
$$
\sqrt{\hat{\beta}_{[2:d]}^T\Sigma_{[2:d,2:d]}\hat{\beta}_{[2:d]}}-\sqrt{\hat{\beta}^T\Sigma\hat{\beta}}<0,
$$
group 0 is over-selected if and only if $C$ implies a threshold $\tilde{y}$ with
$$
\tilde{y}<\left(\hat{\alpha}+\hat{\beta}_{[2:d]}^T\mu_{[2:d]}\right)+\left(\hat{\beta}_1\mu_1\right)\frac{\sqrt{\hat{\beta}_{[2:d]}^T\Sigma_{[2:d,2:d]}\hat{\beta}_{[2:d]}}}{\sqrt{\hat{\beta}_{[2:d]}^T\Sigma_{[2:d,2:d]}\hat{\beta}_{[2:d]}}-\sqrt{\hat{\beta}^T\Sigma\hat{\beta}}}.
$$
The proposition follows.

\paragraph{Corollary~\ref{cor:maincorollary}}
We combine Proposition~\ref{prop:whobenefits} with the parameter estimates given in the proof of Proposition~\ref{prop:propertieshatbeta1}.
For a high threshold $\tilde{y}$, the group with more under-reporting is over-selected if
\begin{align*}
    &\V{\hat{\beta}_{[2:d]}^TZ_{[2:d]}} > \V{\hat{\beta}Z}\\
    \Leftrightarrow & \sum_{i=2}^d\hat{\beta}_i^2\sigma_i^2 +\sum_{i=2}^d\sum_{j=i+1}^d2\hat{\beta}_i\hat{\beta}_j\Cov{Z_i}{Z_j}>\sum_{i=1}^d\hat{\beta}_i^2\sigma_i^2 +\sum_{i=1}^d\sum_{j=i+1}^d2\hat{\beta}_i\hat{\beta}_j\Cov{Z_i}{Z_j}\\
    \Leftrightarrow & \hat{\beta}_1^2\sigma_1^2 + \sum_{j=2}^d2\hat{\beta}_1\hat{\beta}_j\Cov{Z_1}{Z_j} < 0. 
\end{align*}
Recall that
\begin{align*}
    R^2 = \frac{\E{\xi_1}^2\V{Z_1}}{\V{Z_1\xi_1}} \sum_{i=2}^d\rho(Z_1,Z_i)^2
\end{align*}
and denote $S^2=\sum_{i=2}^d\rho(Z_1,Z_i)^2$.
Then
\begin{align*}
    \frac{1}{1-R^2} = \frac{\V{Z_1\xi_1}}{\V{Z_1\xi_1}-\E{\xi_1}^2\V{Z_1}S^2}.
\end{align*}
Using Proposition~\ref{prop:propertieshatbeta1} and 
inserting the parameter estimates gives
\begin{align*}
    & \hat{\beta}_1^2\sigma_1^2 + \sum_{j=2}^d2\hat{\beta}_1\hat{\beta}_j\Cov{Z_1}{Z_j} < 0\\
    \Leftrightarrow & \left(\frac{1}{1-R^2}\right)^2\left(\frac{\E{\xi_1}\V{Z_1}}{\V{Z_1\xi_1}}-\frac{R^2}{\E{\xi_1}}\right)^2\beta_1^2\V{Z_1}\\ 
    &+ \frac{1}{1-R^2}\left(\frac{\E{\xi_1}\V{Z_1}}{\V{Z_1\xi_1}}-\frac{R^2}{\E{\xi_1}}\right)\beta_1\sum_{j=2}^d2\left(\beta_1 \frac{1}{1-R^2}\left(\frac{\Cov{Z_1}{Z_j}(\V{Z_1\xi_1}-\V{Z_1}\E{\xi_1}^2)}{\V{Z_j}\V{Z_1\xi_1}}\right) + \beta_j\right)\Cov{Z_1}{Z_j} < 0\\
    \Leftrightarrow & \left(\frac{1}{1-R^2}\right)\left(\frac{\E{\xi_1}\V{Z_1}}{\V{Z_1\xi_1}}-\frac{R^2}{\E{\xi_1}}\right)\beta_1^2\V{Z_1}\\ 
    &+ 2\frac{1}{1-R^2}\beta_1^2\sum_{j=2}^d \left(\frac{\Cov{Z_1}{Z_j}(\V{Z_1\xi_1}-\V{Z_1}\E{\xi_1}^2)}{\V{Z_j}\V{Z_1\xi_1}}\right)\Cov{Z_1}{Z_j}+ 2\beta_1\sum_{j=2}^d \beta_j\Cov{Z_1}{Z_j} < 0\\
    \Leftrightarrow & \frac{1}{1-R^2}\left(\frac{\E{\xi_1}\V{Z_1}}{\V{Z_1\xi_1}}-\frac{R^2}{\E{\xi_1}}\right)\beta_1^2\V{Z_1}\\ 
    &+ 2\beta_1^2\frac{1}{1-R^2}\frac{(\V{Z_1\xi_1}-\V{Z_1}\E{\xi_1}^2)\V{Z_1}}{\V{Z_1\xi_1}}\sum_{j=2}^d \left(\frac{\Cov{Z_1}{Z_j}^2}{\V{Z_j}\V{Z_1}}\right)+ 2\beta_1\sum_{j=2}^d \beta_j\Cov{Z_1}{Z_j} < 0\\
    \Leftrightarrow & \frac{\E{\xi_1}\V{Z_1}^2}{\V{Z_1\xi_1}-\E{\xi_1}^2\V{Z_1}S^2}(1-S^2)\beta_1^2+ 2\beta_1^2\frac{(\V{Z_1\xi_1}-\V{Z_1}\E{\xi_1}^2)S^2\V{Z_1}}{\V{Z_1\xi_1}-\E{\xi_1}^2\V{Z_1}S^2}+ 2\beta_1\sum_{j=2}^d \beta_j\Cov{Z_1}{Z_j} < 0\\
    \Leftrightarrow & \beta_1^2\V{Z_1}\frac{\V{Z_1}(1-S^2)+2(1-\E{\xi})\E{Z_1^2}S^2}{\V{Z_1}(1-S^2)\E{\xi_1}+(1-\E{\xi_1})\E{Z_1^2}}+ 2\beta_1\sum_{j=2}^d \beta_j\Cov{Z_1}{Z_j} < 0\\
\end{align*}
where we used that $\V{Z_1\xi_1}=\V{Z_1}\E{\xi_1}^2+\V{\xi_1}\E{Z_1^2}$ and $\V{\xi_1}=\E{\xi_1}-\E{\xi_1}^2$.
Note that the first term on the left side is always positive. Thus the inequality is fulfilled if and only if
\begin{align*}
    \text{sign}\left(\beta_1\sum_{j=2}^d \beta_j\Cov{Z_1}{Z_j}\right) = -1 
\end{align*}
and
\begin{align*}
    & 2\beta_1\sum_{j=2}^d \beta_j\Cov{Z_1}{Z_j} <- \beta_1^2\V{Z_1}\frac{\V{Z_1}(1-S^2)+2(1-\E{\xi_1})\E{Z_1^2}S^2}{\V{Z_1}(1-S^2)\E{\xi_1}+(1-\E{\xi_1})\E{Z_1^2}}\\
    \Leftrightarrow & \frac{1}{\beta_1}\sum_{j=2}^d \beta_j\Cov{Z_1}{Z_j} <- \frac{\V{Z_1}^2(1-S^2)+2(1-\E{\xi_1})\E{Z_1^2}\V{Z_1}S^2}{2\V{Z_1}(1-S^2)\E{\xi_1}+2(1-\E{\xi_1})\E{Z_1^2}}\\
    \Leftrightarrow & \frac{1}{\beta_1}\sum_{j=2}^d \beta_j\Cov{Z_1}{Z_j} <- \frac{\V{Z_1}^2(1-S^2)+2(1-\E{\xi_1})(\V{Z_1}+\E{Z_1}^2)\V{Z_1}S^2}{2\V{Z_1}(1-S^2)\E{\xi_1}+2(1-\E{\xi_1})(\V{Z_1}+\E{Z_1}^2)}.
\end{align*}
Since we know that the fraction on the right side is always positive, this can be rewritten as presented in the corollary. 
If the inequality is not fulfilled, the group with more under-reporting is under-selected at a high threshold.

\paragraph{Lemma~\ref{lem:augmented_loss}}
In the setting of the Lemma, we can write
\begin{align*}
    \mathds{E}_\xi\left[l(f,X,y)\right] &= \mathds{E}_\xi\left[\frac{1}{m}l(f,X,y) - \frac{1-m}{m}l(f,[0,X_{[2:d]}]^T,y)\right]\\
    &= \frac{1}{m}\mathds{E}_\xi\left[l(f,X,y)\right] - \frac{1-m}{m}l(f,[0,z_{[2:d]}]^T,y)\\
    &= \frac{1}{m}\left(P(\xi_1=1)l(f,z,y) + P(\xi_1=0)l(f,[0,z_{[2:d]}]^T,y)\right) - \frac{1-m}{m}l(f,[0,z_{[2:d]}]^T,y)\\
    &= l(f,z,y) + \frac{1-m}{m}l(f,[0,z_{[2:d]}]^T,y) - \frac{1-m}{m}l(f,[0,z_{[2:d]}]^T,y)\\
    &= l(f,z,y).
\end{align*}
Here, the first equality holds since only the first feature has under-reporting and the second equality holds because $Z\bot G$ which implies $Z\bot \xi$.

\paragraph{Lemma~\ref{lem:group_dep_augmented_loss}}
Follows the same as Lemma~\ref{lem:augmented_loss}. Instead of under-reporting completely at random, the under-reporting is completely at random within group $g$.

\paragraph{Lemma~\ref{lem:optimal_prediction_imp_value}}
Since $f$ is linear and under-repoting only occurs in the first feature, the expected prediction error for an imputation value $x_1'$ can be written as
\begin{align*}
    R(f) &= \mathds{E}_X\left[(f(X')-Y)^2\right]\\
    &=\beta_1^2\mathds{E}_X\left[(X_1'-Z_1)^2\right]\\
    &=\beta_1^2P(X_1=0)\mathds{E}_X\left[(X_1'-Z_1)^2\mid X_1=0\right] + \beta_1^2P(X_1\neq 0)\mathds{E}_X\left[(X_1'-Z_1)^2\mid X_1\neq 0\right]\\
    &= \beta_1^2P(X_1=0)\mathds{E}_Z\left[(x_1'-Z_1)^2\mid X_1=0\right] + \beta_1^2P(X_1\neq 0)\mathds{E}_Z\left[(Z_1-Z_1)^2\mid X_1\neq 0 \right]\\
    &=\beta_1^2P(X_1=0)\left(x_1'^2 - 2x_1'\mathds{E}\left[Z_1\mid X_1=0\right] + \mathds{E}\left[Z_1^2\mid X_1=0\right]\right).
\end{align*}
Then
\begin{align*}
    &\frac{dR(f)}{dx_1'} = \beta_1^2P(X_1=0)(2x_1' - 2\E{Z_1\mid X_1=0}) \overset{!}{=}0\\
    \Leftrightarrow &x_1' = \E{Z_1\mid X_1=0}.
\end{align*}
We implicitly assume that $\beta_1\neq 0$ and the probability of 0-entries is positive.

\paragraph{Lemma~\ref{lem:group_optimam_prediction_imp_value}}
Recall that $G\sim \text{Bern}(r)$.
Similar to the proof of Lemma~\ref{lem:optimal_prediction_imp_value}, the expected prediction error can be written as
\begin{align*}
    R(f) &= \mathds{E}_X\left[(f(X')-Y)^2\right]\\
    &= \beta_1^2 P(X_1=0)\mathds{E}_Z\left[(X_1'-Z_1)^2\mid X_1=0\right]\\
    &=\beta_1^2 P(X_1=0)\left(r\mathds{E}_Z\left[(X_1'-Z_1)^2\mid X_1=0,G=1\right]+(1-r)\mathds{E}_Z\left[(X_1'-Z_1)^2\mid X_1=0,G=0\right]\right)\\
    &=\beta_1^2 P(X_1=0)\left(r\mathds{E}_Z\left[(x_1'^1-Z_1)^2\mid X_1=0,G=1\right]+(1-r)\mathds{E}_Z\left[(x_1'^0-Z_1)^2\mid X_1=0,G=0\right]\right).
\end{align*}
This prediction error is minimal when
\begin{align*}
    &\frac{dR(f)}{dx_1'^g} = \beta_1^2P(X_1=0)P(G=g)(2x_1'^g-2\E{Z_1\mid X_1=0,G=g})\overset{!}{=} 0 \\
    \Leftrightarrow &
    x_1'^g = \E{Z_1\mid X_1=0,G=g}
\end{align*}
for $g\in\{0,1\}$.

\section{Supplementary Figures}

\begin{table}[H]
\small{
    \centering
    \begin{tabular}{|P{4cm}|C{0.85cm}|C{0.85cm}|P{4.5cm}|P{3.2cm}|}
    \hline
        \textbf{Name} & \textbf{\#Obs.} & \textbf{\#Feat.} & \textbf{Groups} & \textbf{Binary outcomes} \\
         \hline
         COMPAS \cite{compas} & 7,214 & 6 & Race (51\% African-American, 49\% other), Gender (81\% male, 19\% female) & Two-year recidivism, violent recidivism \\
         \hline
         German credit \cite{German} & 1,000 & 19 & Gender (69\% male, 31\% female) & Good credit \\
         \hline
         ACS Income (CA, 2018) \cite{folktables}& 195,665 & 6 & Race (62\% White, 38\% other), Gender (53\% male, 47\% female) & Yearly income over \$50,000 \\
         \hline
         Birth data & 39,365 & 51 & Medicaid (no 72\%, yes 28\%), Race (African-American 21\%, other 79\%) & Child placed in foster care within 3 years \\
         \hline
    \end{tabular}
    }
    \caption{Statistics of the datasets used in experiments. Data is split randomly into 80\% for training and 20\% for testing. For the first three datasets, we iterate over all outcome types, groups, and numerical features for missingness injection.}
    \label{table:datasets}
\end{table}

\begin{figure}[H]
    \centering
    \includegraphics[scale = 0.6]{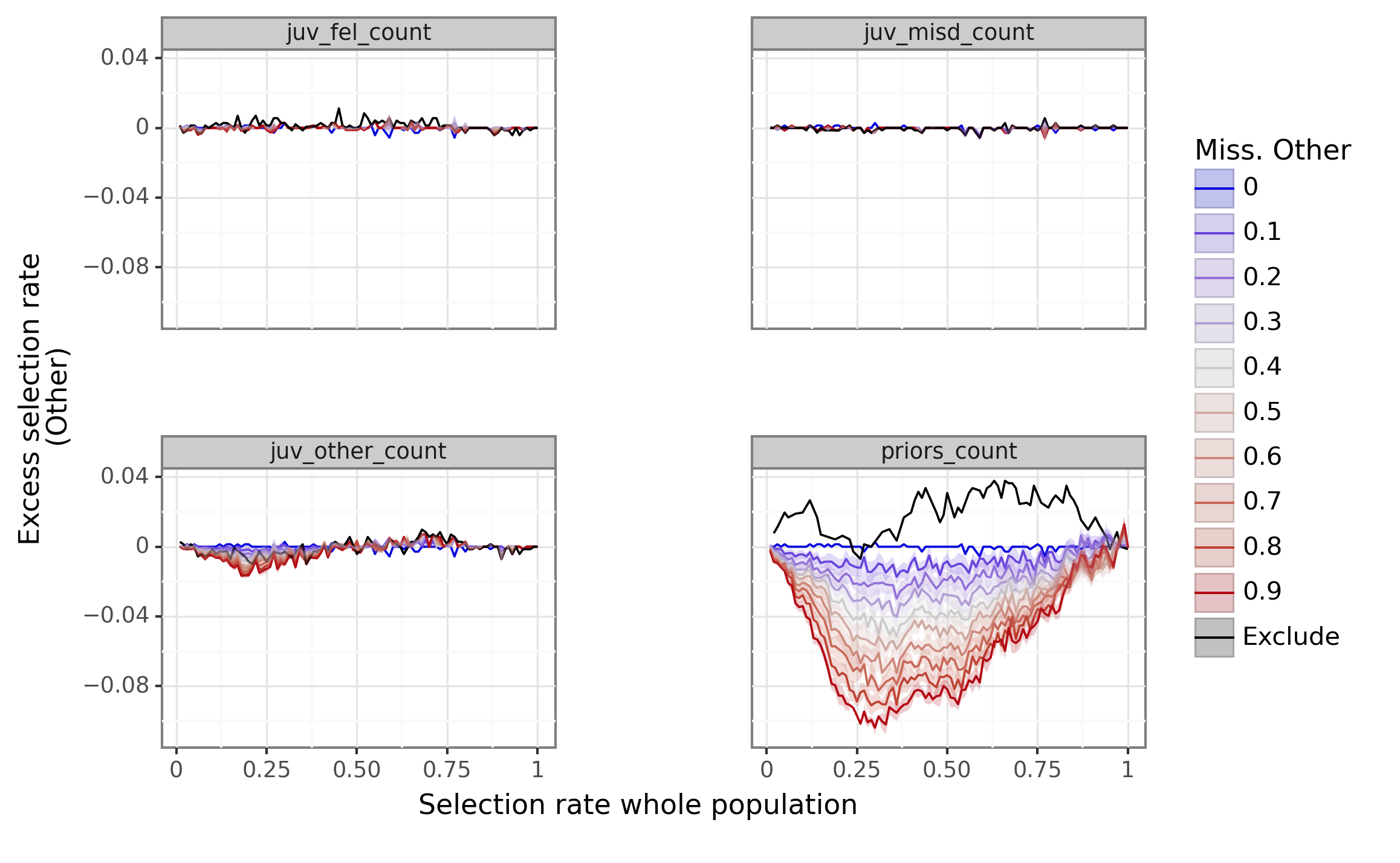}
    \caption{Excess selection rate of racial group Other (i.e. not African-American) at different selection rates of the whole population with synthetic two-year recidivism outcomes using the COMPAS dataset. Each panel represents a feature that has been corrupted by under-reporting in independent runs of the experiment. Feature under-reporting is added to the Other group with 0-90\% missing in 10 percentage point increments. The black curves show performance when excluding the whole feature column from modeling. Results are reported as averages over 30 runs on a test set. Shaded areas correspond to one standard deviation in each direction of the mean.}
    \label{fig:compas}
\end{figure}

\begin{figure}[H]
    \centering
    \begin{subfigure}[b]{0.49\linewidth}
        \includegraphics[scale=0.6]{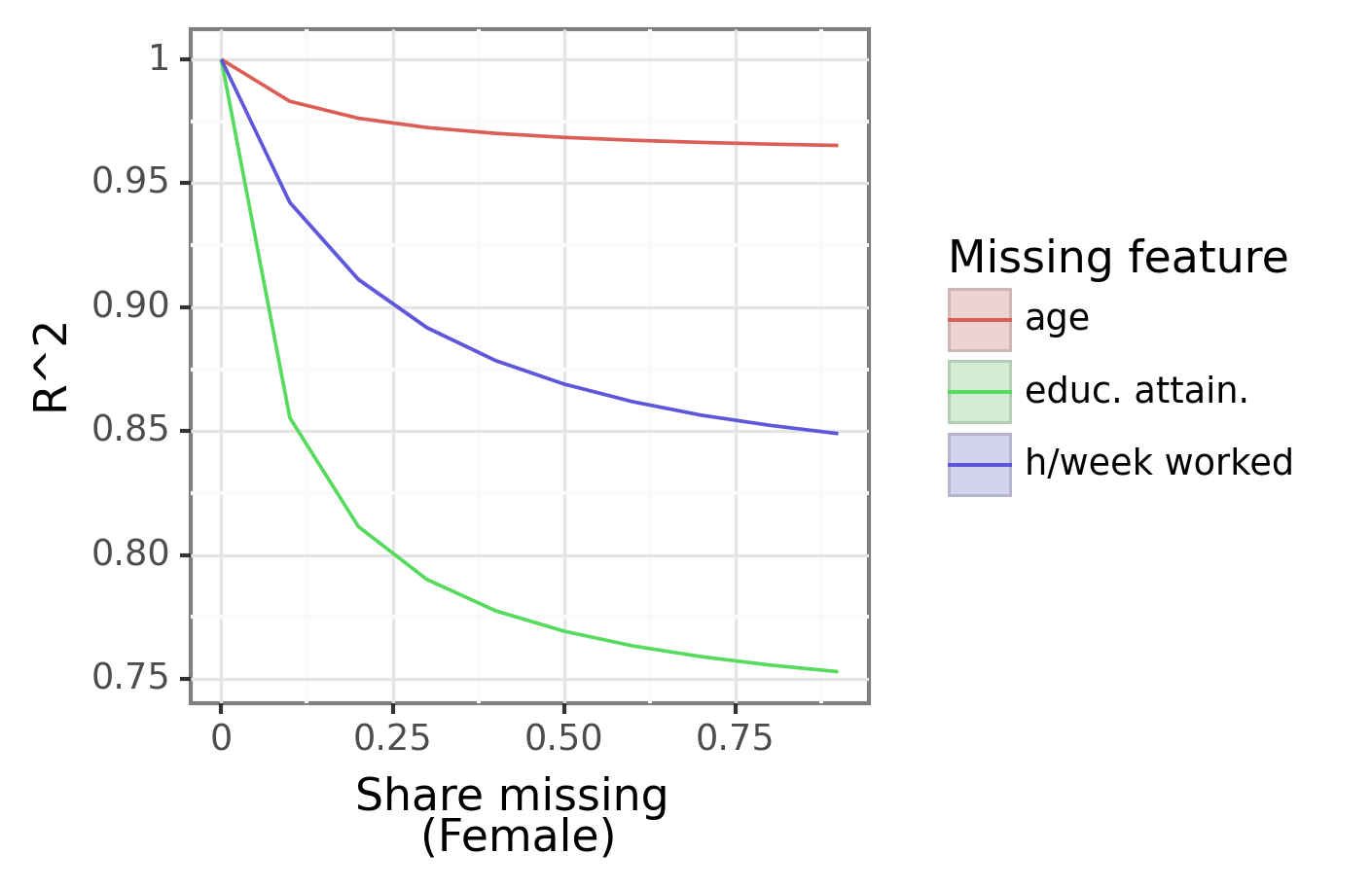}
        \subcaption{Female group}
    \end{subfigure}
    \begin{subfigure}[b]{0.49\linewidth}
        \includegraphics[scale=0.6]{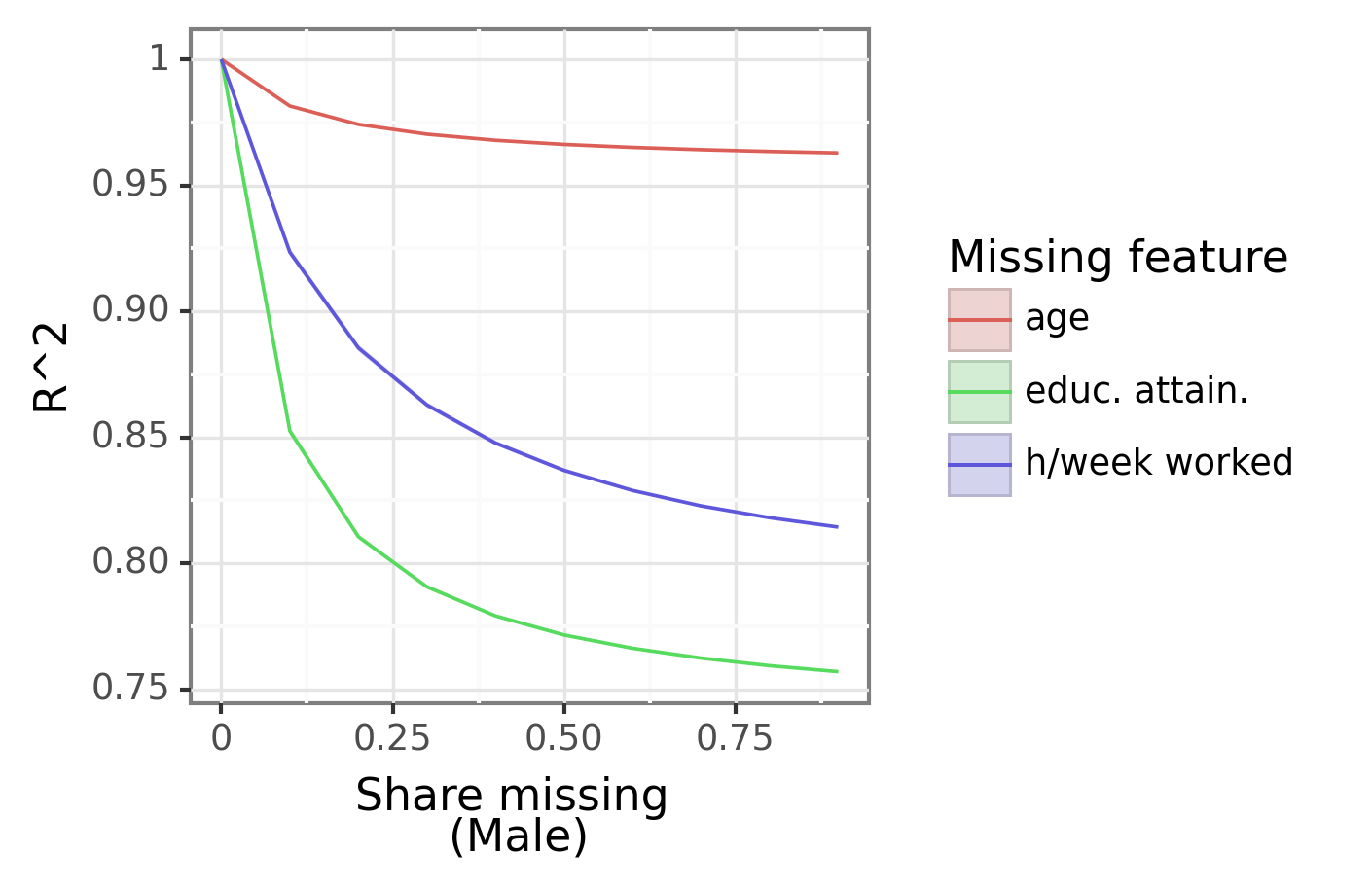}
        \subcaption{Male group}
    \end{subfigure}
    \caption{Test set $R^2$ over varying levels of feature under-reporting in the ACS Income dataset. Results are reported as averages over 50 runs on the test set. Variation in results was minimal.}
    \label{fig:acsincome_test_r2}
\end{figure}

\begin{figure}[H]
    \centering
    \includegraphics[scale = 0.6]{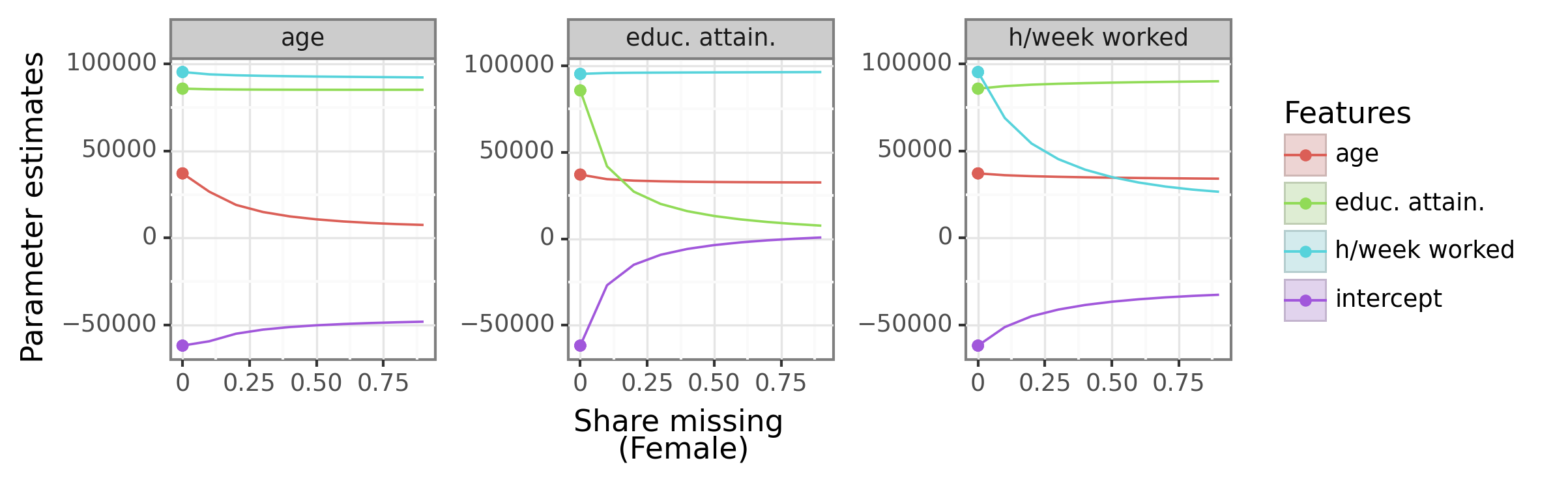}
    \caption{Parameter estimates over varying levels of feature under-reporting in the ACS Income dataset. Each panel indicates a different feature selected for under-reporting injection. Points indicate the true parameters from the semi-synthetic ground truth model. Results are reported as averages over 50 runs. Variation in estimates over runs was minimal. Note that only estimates for continuous features are displayed and the estimated parameters for the levels of the categorical variables worker class, marital status, and relationship to reference person are omitted for readability.}
    \label{fig:paramsacs}
\end{figure}

\begin{figure}[H]
    \centering
    \includegraphics[scale = 0.6]{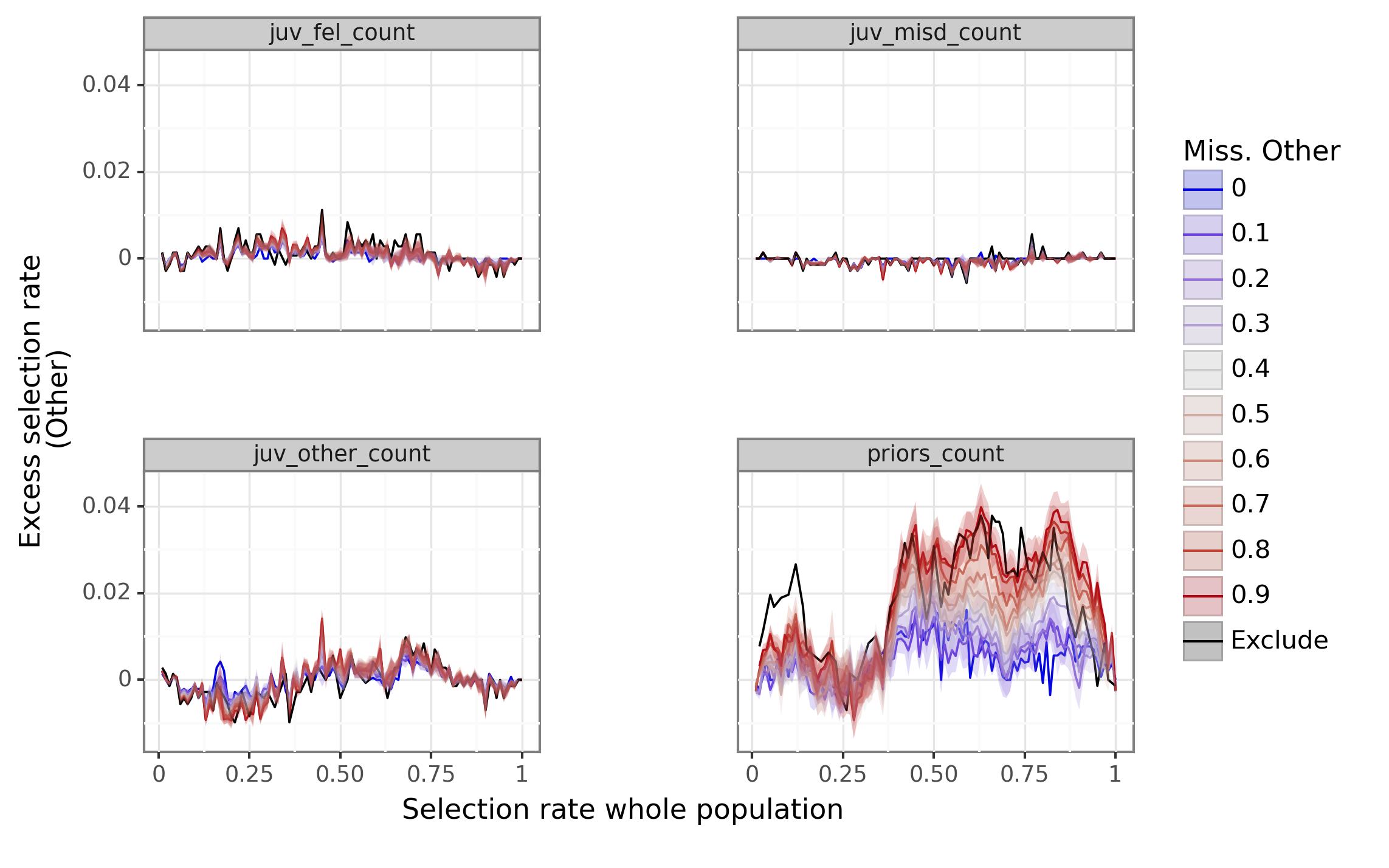}
    \caption{Multiple imputation excess selection rate of racial group Other (i.e. not African-American) at different selection rates of the whole population with synthetic two-year recidivism outcomes using the COMPAS dataset. Results are reported as averages over 30 runs on a test set. Shaded areas correspond to one standard deviation in each direction of the mean.}
    \label{fig:multipmcompas}
\end{figure}

\begin{figure}[H]
    \centering
    \includegraphics[scale = 0.6]{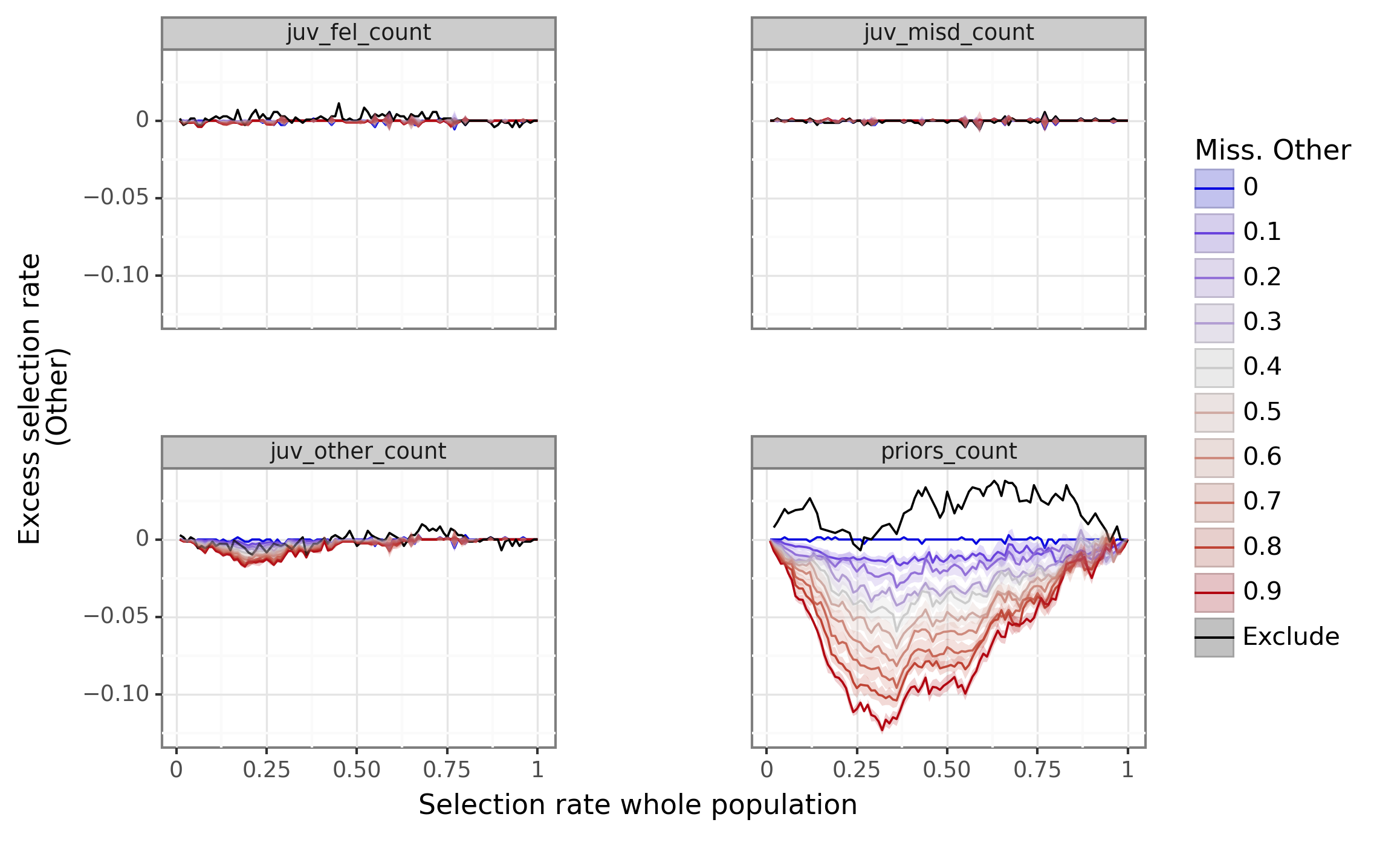}
    \caption{Excess selection rate of racial group Other (i.e. not African-American) when training on rows without 0-entries at different selection rates of the whole population with synthetic two-year recidivism outcomes using the COMPAS dataset. Results are reported as averages over 30 runs on a test set. Shaded areas correspond to one standard deviation in each direction of the mean.}
    \label{fig:onlyzeroscompas}
\end{figure}

\begin{figure}[H]
    \centering
    \includegraphics[scale = 0.6]{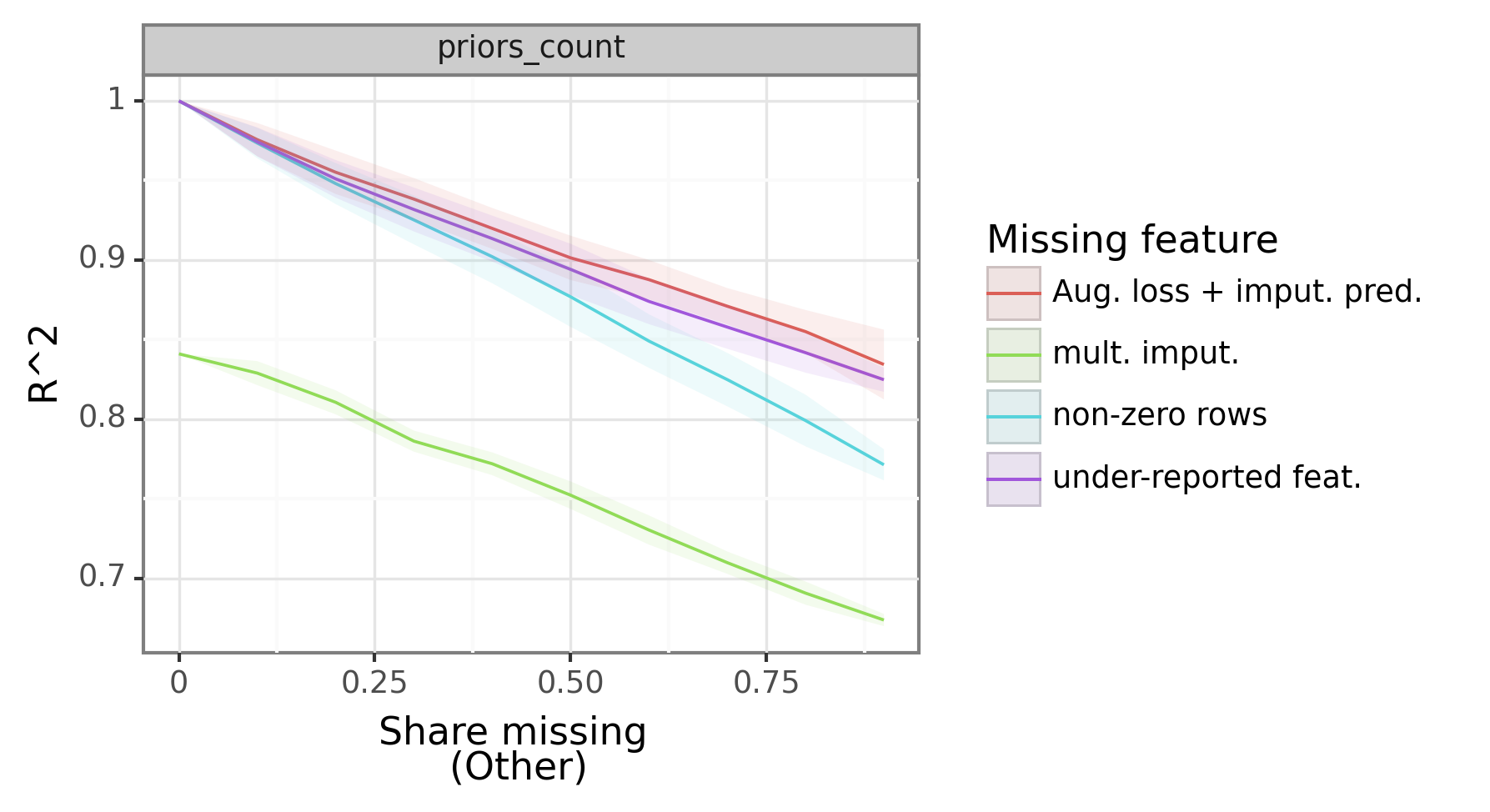}
    \caption{Test set $R^2$ of different solution approaches using the COMPAS dataset with synthetic two-year recidivism outcomes. Under-reporting is injected into the feature `priors count' of group Other (i.e. not African-American). Results are reported as averages over 30 runs. Shaded areas correspond to one standard deviation.}
    \label{fig:r_sq_facet}
\end{figure}

\section{Under-reporting rate estimation}
\label{app:missingness_estimation}

\subsection{Estimation procedure}
We draw on PU-learning literature to estimate under-reporting rates. \citet{Elkan2008} assume a classification setting with a latent indicator $s\in\{0,1\}$ that encodes whether an example is labeled or not. Only positive examples are labeled, i.e. $s=1$ implies that $y=1$, but when the example is unlabeled $s=0$, we don't know if $y=0$ or $y=1$. 
In our case, the outcome $y$ translates to the indicator $\mathds{1}(Z_1\neq 0)$ while the labeling indicator $s$ translates to $\mathds{1}(X_1\neq 0)$. If $X_1\neq 0$, we know that the example is not under-reported and $Z_1\neq 0$. But, if $X_1=0$, we don't have any information about the value of $\mathds{1}(Z_1\neq 0)$ because the example may not be `labeled'. Similar to the assumption in \cite{Elkan2008}, our setting fulfills an under-reported completely at random assumption which can be expressed as
$$
 P(X_1\neq 0\mid X_{[2:d]},Y,Z_1\neq 0)=P(X_1\neq 0\mid Z_1\neq 0).
$$
In our analysis, this assumption is fulfilled either over the whole population or within the group considered for estimation of $m_g$.
Given this notation, the rate of correctly observed feature entries $m$ can be written as
$$
m = P(\xi_1\neq 0) = P(\xi_1Z_1\neq 0\mid Z_1\neq 0) = P(X_1\neq 0 \mid Z_1\neq 0).
$$
This implies that we can estimate $m$ without having to consider correctly recorded 0-entries in the feature vector. As described in the main text, we assume access to a data set $V=\{(x,y)_{i=1}^n\}$ that is split into a training portion $V_{\text{train}}$ and evaluation portion $V_{\text{eval}}$. Let $P_{\text{eval}}$ denote the subset of examples from $V_{\text{eval}}$ for which $x_1\neq 0$. We now fit a model $h$ on $V_{\text{train}}$ to estimate $P(X_1\neq 0\mid X_{[2:d]},Y)$ and evaluate $h$ on $P_{\text{eval}}$. The estimator for the share of observed values $m$ is given by
$$
\hat{m} = \frac{1}{\mid P_{\text{eval}}\mid} \sum_{(x,y)\in P_{\text{eval}}} h(x_{[2:d]},y).
$$
Assuming $h(x_{[2:d]},y)=P(X_1\neq 0\mid X_{[2:d]}=x_{[2:d]},Y=y)$, i.e. no error is introduced through the estimation of $h$, this provides an unbiased estimate of $m$. To see this, we show that $h(x_{[2:d]},y)=m$ for all $(x,y)\in P_{\text{eval}}$.
We can write
\begin{align*}
    h(x_{[2:d]},y)=&P(X_1\neq 0\mid X_{[2:d]}=x_{[2:d]},Y=y)\\
    =&P(X_1\neq 0\mid X_{[2:d]}=x_{[2:d]},Y=y,Z_1\neq 0)P(Z_1\neq 0\mid X_{[2:d]}=x_{[2:d]},Y=y)\\
    &+P(X_1\neq 0\mid X_{[2:d]}=x_{[2:d]},Y=y,Z_1=0)P(Z_1=0\mid X_{[2:d]}=x_{[2:d]},Y=y)\\
    =&P(X_1\neq 0\mid X_{[2:d]}=x_{[2:d]},Y=y,Z_1\neq 0)\\
    =&P(X_1\neq 0\mid Z_1\neq 0) = m.\\
\end{align*}
Here, the third equality follows because $(x,y)\in P_{\text{eval}}$.

\subsection{Estimation results}
\begin{figure}[H]
    \centering
    \includegraphics[scale = 0.6]{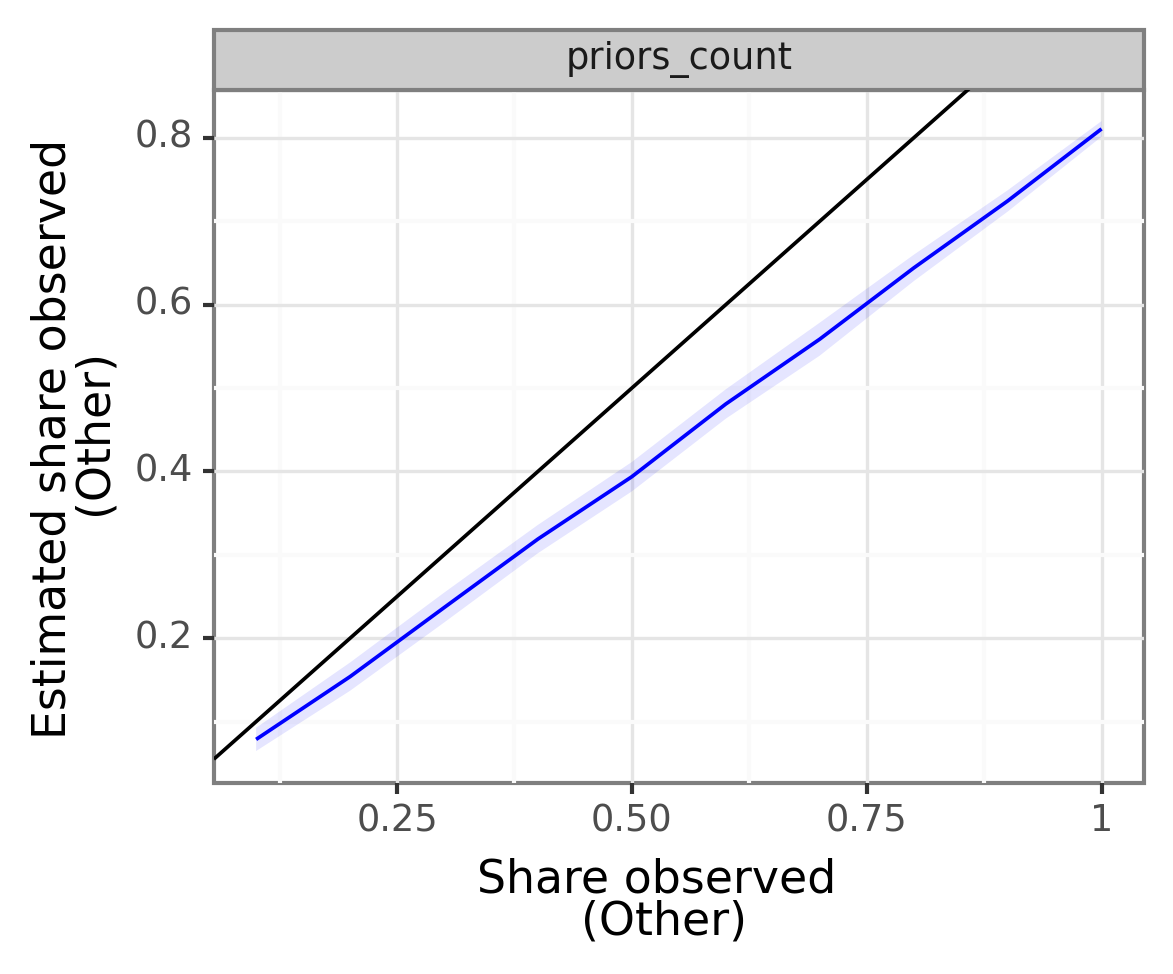}
    \caption{True vs. estimated observed rate (i.e. 1 - under-reporting rate) of feature priors count in racial group Other (i.e. not African-American) with sythetic two-year recidivism outcomes using the COMPAS dataset. Results are reported as averages over 30 runs. Shaded areas correspond to one standard deviation in each direction of the mean. The black line shows $y=x$ for comparison.}
    \label{fig:miss_est}
\end{figure}

We estimate under-reporting rates using the procedure described above where $V$ is taken to be the 80\% training data fold conditioned on the group with under-reporting. Half of the data is used for training of $h$ while the other half is used as evaluation data to compute $\hat{m}_g$. As model class for $h$, we use XGBoost classifiers with 100 trees of maximum depth 3, and learning rate 0.1. 
Figure~\ref{fig:miss_est} depicts the results of the estimation procedure for under-reporting in the feature priors count for the racial group Other and synthetic two-year recidivism outcomes using the COMPAS dataset. We see that estimation of $m_g$ works particularly well when under-reporting in the feature is high. When the feature is fully observed, i.e. $m_g=1$, the estimator returns $\hat{m}_g=0.811$ on average. We hypothesize that this is due to the high share of true 0-entries occurring in the priors count feature which impacts the estimator more when fewer values are missing due to under-reporting.

\section{Additional experiments and results}
\subsection{German credit data}
\label{app:german}

\begin{figure}[H]
    \centering
    \includegraphics[scale = 0.6]{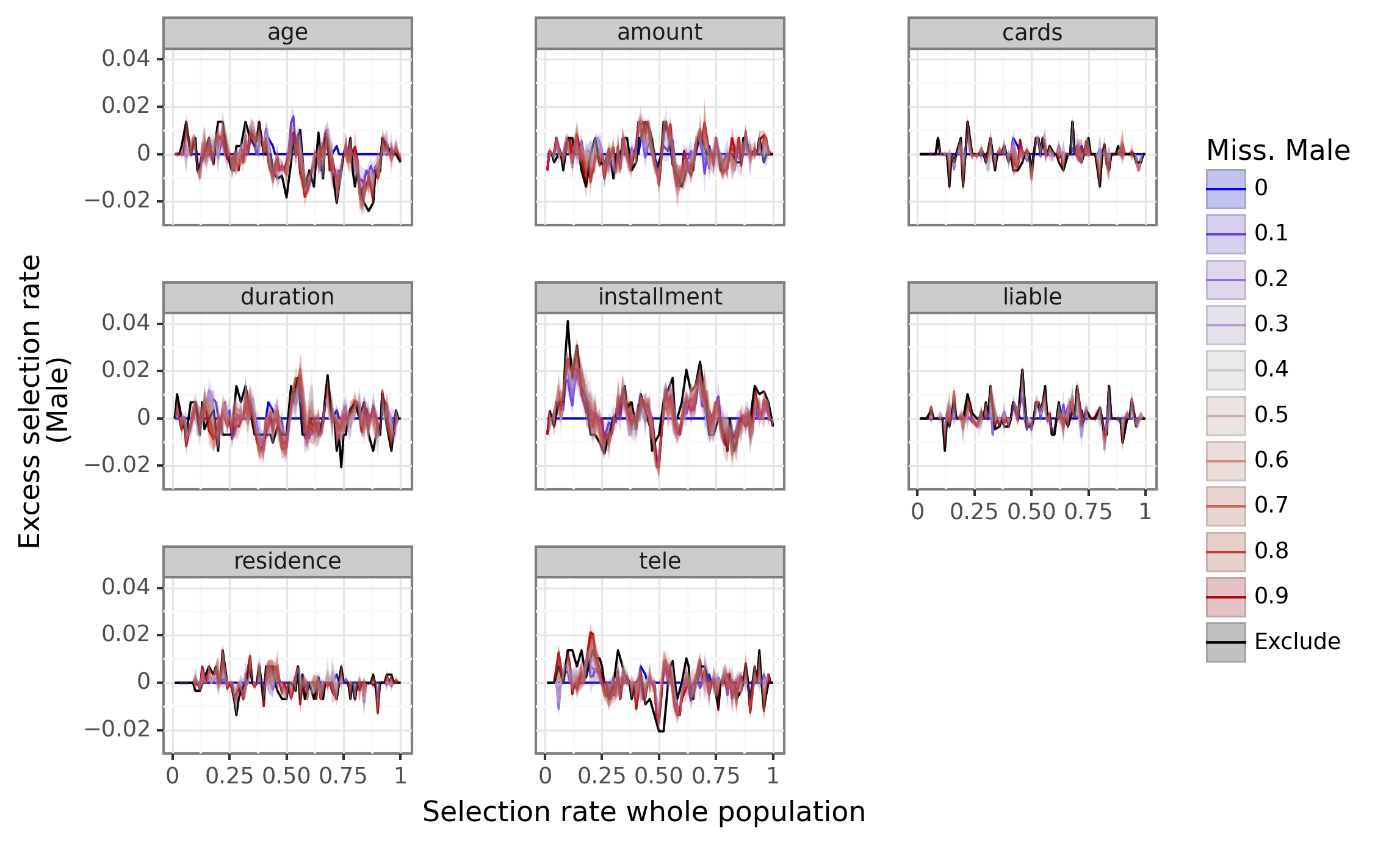}
    \caption{Excess selection rate of male group at different selection rates of the whole population with synthetic outcomes using the German credit dataset. Each panel represents a feature that has been corrupted by under-reporting in independent runs of the experiment. Feature under-reporting is added to the male group with 0-90\% missing in 10 percentage point increments. The black curves show performance when excluding the whole feature column from modeling. Results are reported as averages over 50 runs on a test set. Shaded areas correspond to one standard deviation in each direction of the mean. Note that feature under-reporting is only injected into continuous features and models are estimated using the displayed features as well as the available categorical features checking account status, credit history, purpose, savings, employment, marital status, type of owned property, other installment plans, housing type, and job type.}
    \label{fig:german}
\end{figure}

Our experiments suggest that addition of feature under-reporting to one of the two gender groups in the German credit dataset has only marginal fairness implications.
Figure~\ref{fig:german} depicts the results for synthetic outcomes and addition of different amounts of under-reporting to the features of the male group. We can see that, for any of the considered features, the amount of under-reporting injected has little to no effect on the excess selection rate of the male group. 
However, when selecting rates of around 10-15\% from the whole population \emph{any} amount of under-reporting in the installment feature appears to results in a slight over-selection of the male group. 
The installment feature in the German credit dataset is discretized into four values with lower values indicating a higher installment rate. Incorrectly observed 0-values may thus suggest a high installment rate which is indicative of good credit.

\subsection{Beyond the noise-free setting}
\label{app:noise}

\paragraph{Motivation}
The experiments on the publicly available datasets discussed in Sections~\ref{sec:experiments} and \ref{sec:experimentresults} rely on semi-synthetic outcomes that are computed as deterministic linear functions of correctly measured features.
The implicit simplifying assumptions are that, without feature under-reporting, a linear model on the data can retrieve the true data generating model $f(X)=\alpha + \beta^TZ$ and the exact outcomes $Y$ as recorded in the data.
This modeling choice facilitates isolation of the effect of feature under-reporting by explicitly excluding potential effects of model misspecification and regression noise.
In real-life applications, we can generally not predict outcomes exactly even if correctly measured features are available. In the following additional set of experiments, we loosen the assumption of a noise-free regression setting to allow for more general settings.

\paragraph{Experimental setup}
We follow a similar experimental setup as described in Section~\ref{sec:setup}. Instead of relying on noise-free regression labels $Y = \alpha + \beta^TZ$, we add some noise back into the system by setting
$$
    Y = \alpha + \beta^TZ + \varepsilon.
$$
Here, $\varepsilon\sim\mathcal{N}(0,\sigma^2)$ is i.i.d. and assumed to have mean zero.
Like before, fitting a linear regression of $Y$ on features $Z$ with sufficient data yields the correct parameter estimates $\hat{\beta}=\beta$ and $\hat{\alpha}=\alpha$. However, in contrast to the noise-free setting, the prediction model $\hat{Y}_Z=\alpha + \beta^TZ$ can only retrieve labels $Y$ up to random noise. 
The $R^2$ of this prediction model can be controlled via the variance $\sigma^2$ by setting
$$
    \sigma^2 = \frac{1-R^2}{R^2}\E{\left((\alpha+\beta^TZ) - \E{\alpha+\beta^TZ}\right)^2}.
$$
We experiment with $R^2$ values between 0.1 and 1.0 in increments of 0.1. Instead of comparing predictions $\hat{Y}_Z$ to predictions under feature under-reporting $\hat{Y}_X$, we compare thresholded versions of outcomes $Y$ and $\hat{Y}_X$ directly to measure both the impact of under-reporting and regression noise. 

\begin{figure}[t]
    \centering
    \begin{subfigure}[b]{0.32\linewidth}
    \centering
    \includegraphics[scale=0.4]{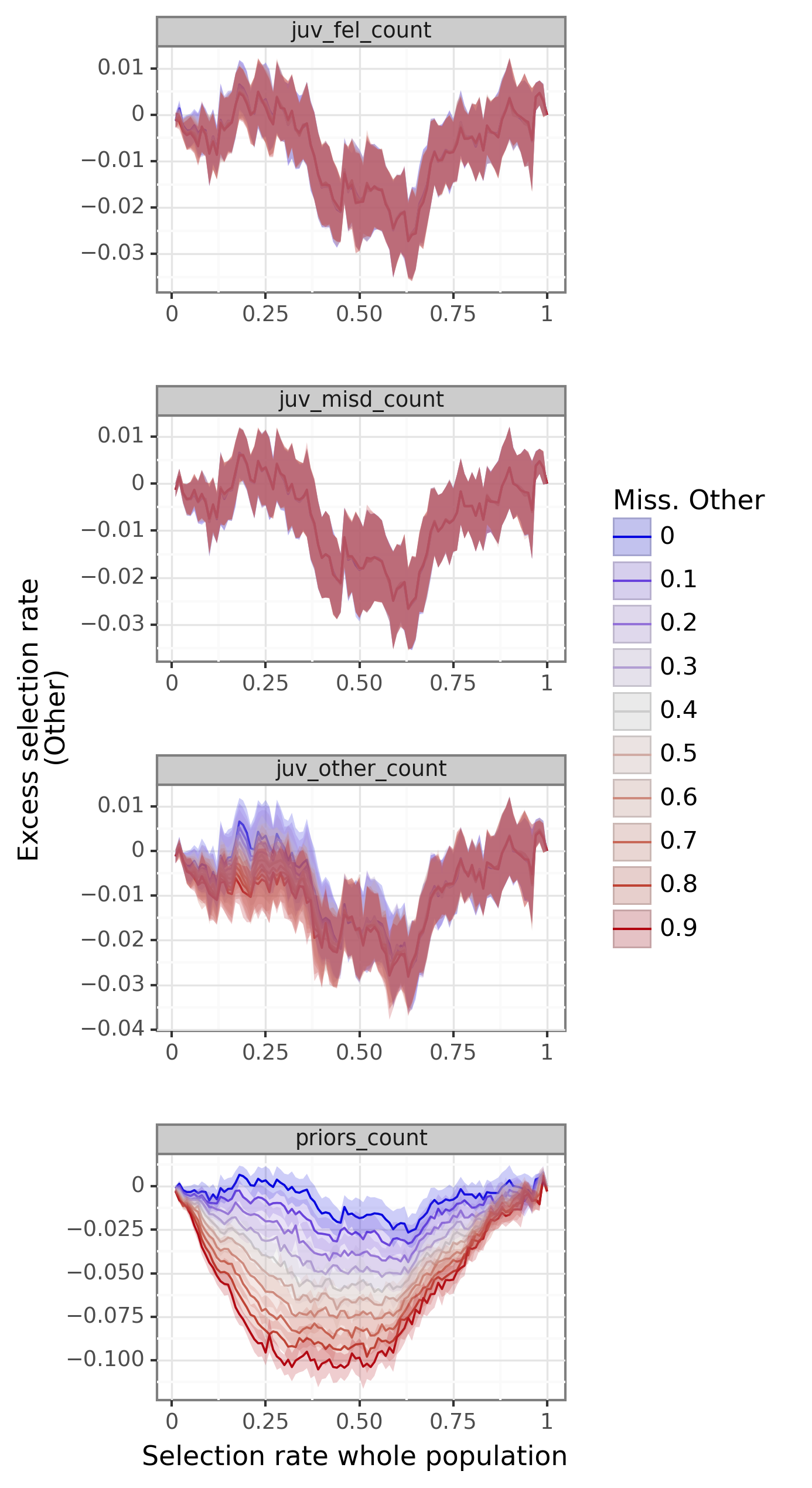}
    \subcaption{Model on $Z$: $R^2=0.9$.}
    \end{subfigure}
    \begin{subfigure}[b]{0.32\linewidth}
    \centering
    \includegraphics[scale=0.4]{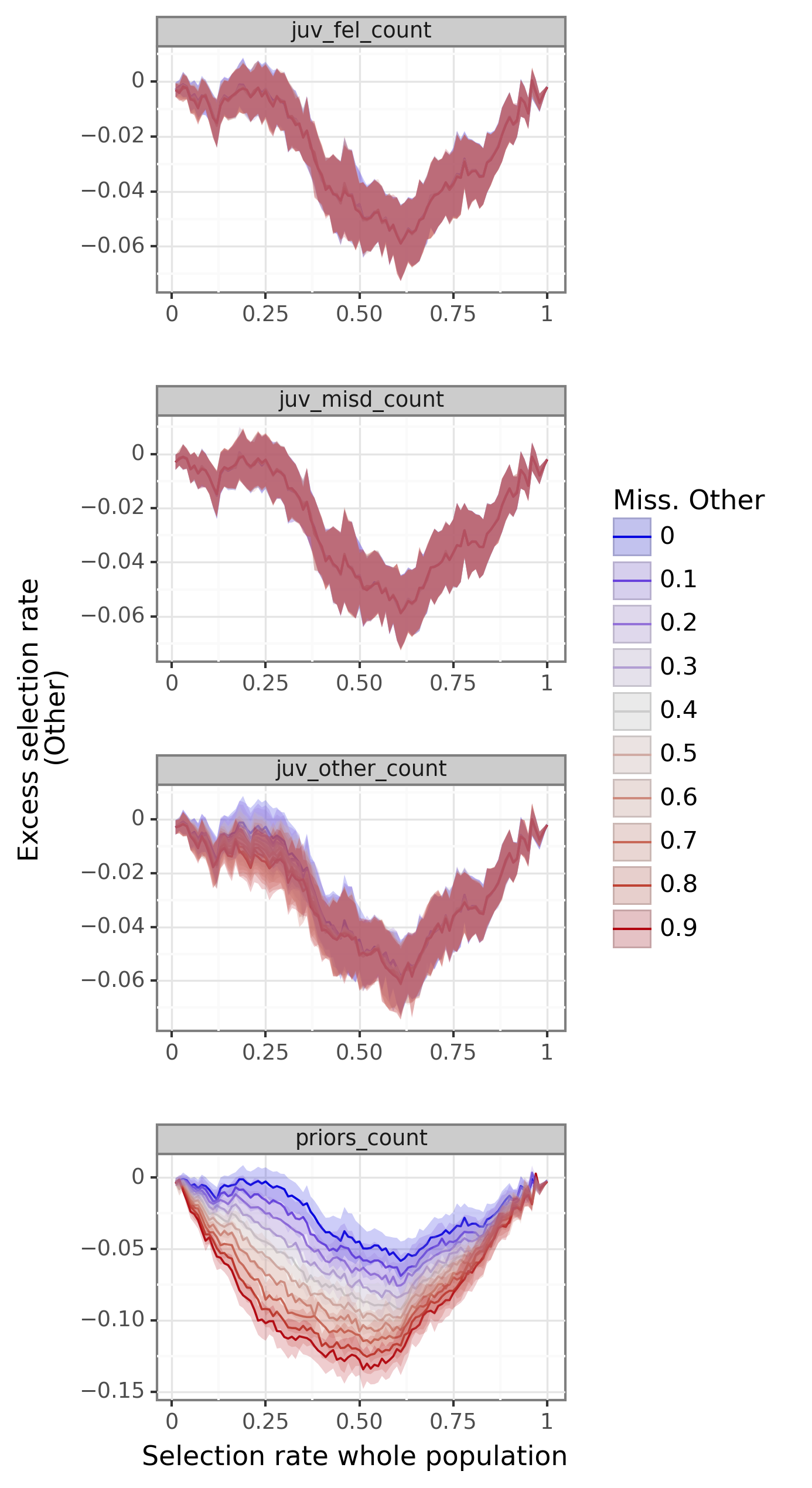}
    \subcaption{Model on $Z$: $R^2=0.6$.}
    \end{subfigure}
    \begin{subfigure}[b]{0.32\linewidth}
    \centering
    \includegraphics[scale=0.4]{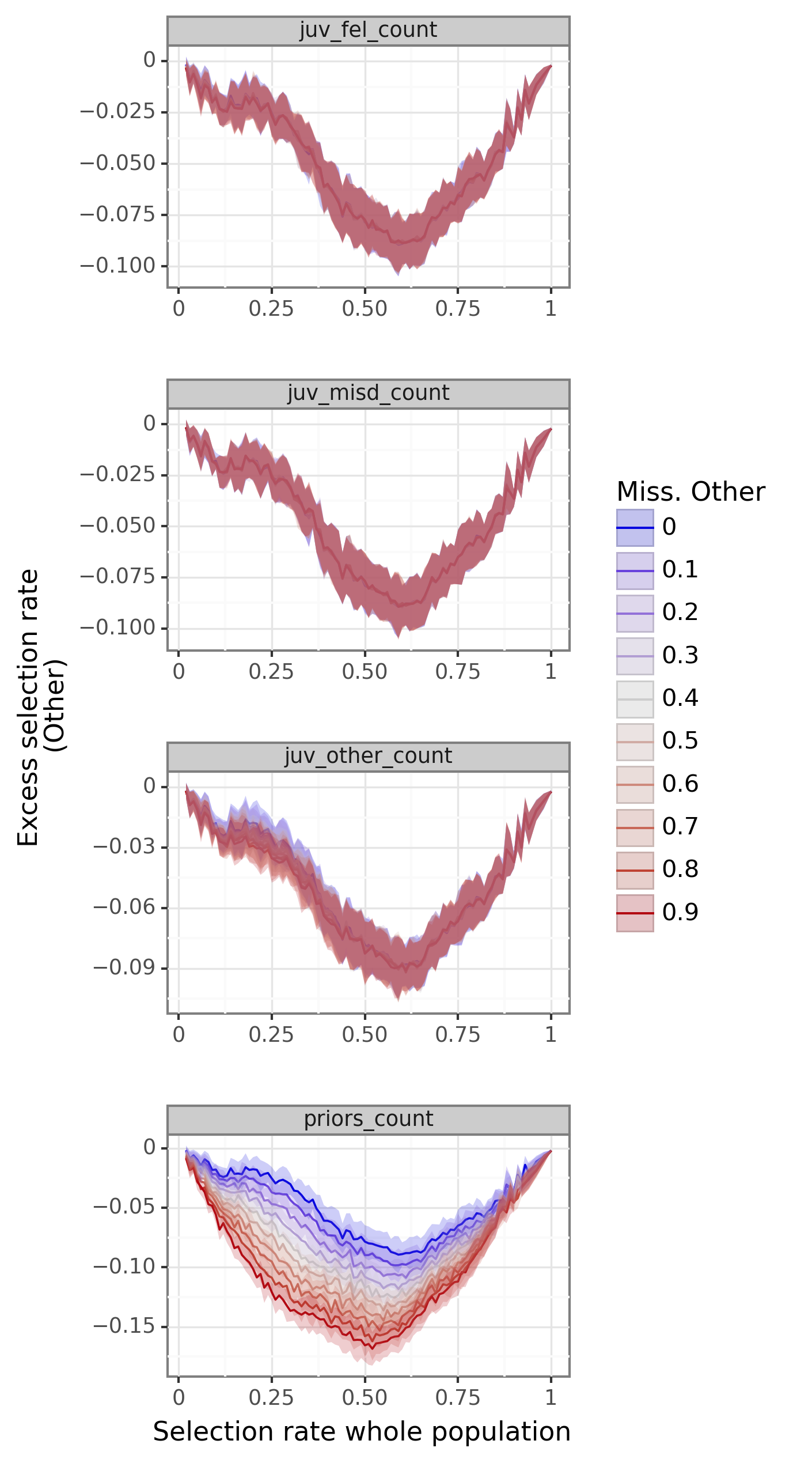}
    \subcaption{Model on $Z$: $R^2=0.3$.}
    \end{subfigure}
    \caption{Excess selection rate  under under-reporting over true labels $Y$ with different levels of $R^2$ for the model on correctly measured features $Z$. Low $R^2$ indicates a high level of noise and vice versa. Under-reporting injected into the features of group Other (i.e. not African-American) in the COMPAS dataset. Results are reported as averages over 30 simulation runs with shaded areas representing one standard deviation in each direction.}
    \label{fig:compasnoise}
\end{figure}

\paragraph{Results}
Figure~\ref{fig:compasnoise} depicts a subset of the results for the COMPAS dataset. Comparing against the results of the noise-free setting summarized in Figure~\ref{fig:compas}, we observe that the group Other is under-selected to a greater extend with additional noise. Under-selection occurs even if no feature under-reporting is added (dark blue curves) and increases with increasing noise, i.e. decreasing $R^2$ of the model on $Z$.
On a high level, this occurs because the predictions $\hat{Y}_Z$ concentrate more closely around their group-level means as compared to the true values $Y$. The mean of $\hat{Y}_Z$ is smaller for the group Other than the group African-American which leads to under-selection of the group Other as compared to the true $Y$ at many thresholds. We note that the group-level variances in outcomes $Y$ and predictions $\hat{Y}_Z$ play a role in this dynamic as well.
The isolated effect of feature under-reporting in the studied setting appears to be similar to the effect in the noise-free setting. As under-reporting is introduced into the group Other via the feature `priors count', the group Other is further under-selected. The more under-reporting is injected, the more the group is under-selected. The magnitude of under-selection due to under-reporting is comparable across different levels of regression noise.
Overall, the results give us some insight into what to expect in more realistic settings of feature under-reporting.
Instead of selection rate disparities that are exclusively due to differential feature under-reporting, disparities in the studied setting also depend on regression noise which, together, leads to increased disparities overall.

\subsection{Possibility of decreasing disparities}
\label{app:decreasingcompas}
We conduct our main experiments on three publicly available datasets, i.e. COMPAS data \cite{compas}, German credit data \cite{German}, and ACS Income data \cite{folktables}, where each numerical feature is considered for the effect of under-reporting. As discussed in Section~\ref{sec:experimentresults}, the results suggest that, if an effect is present, feature under-reporting generally leads to under-selection of the group with under-reporting which aligns with Case 2 from the theoretical derivations in Section~\ref{sec:xx}. If the group with under-reporting aligns with the group that is less frequently selected in the ground truth model, this implies that differential feature under-reporting leads to increased selection rate disparities.

All three datasets have a numerical age feature which was considered for under-reporting but omitted for the discussion of results in the main text. In contrast to most other features (e.g. the counts in the COMPAS data), the default value of 0 is somewhat unintuitive for age and lies outside of the feature’s support in each of the datasets. Studying the effect of fitting a model on differentially available data directly is less compelling in this setting since we essentially have indicators for under-reporting and could hope to use missing data methods like imputation directly. Nevertheless, we discuss the results for feature under-reporting in age for the COMPAS dataset in the following as it presents the only empirical example for decreasing disparities we encounter in our experiments. 

\begin{figure}
    \centering
    \includegraphics[scale=0.5]{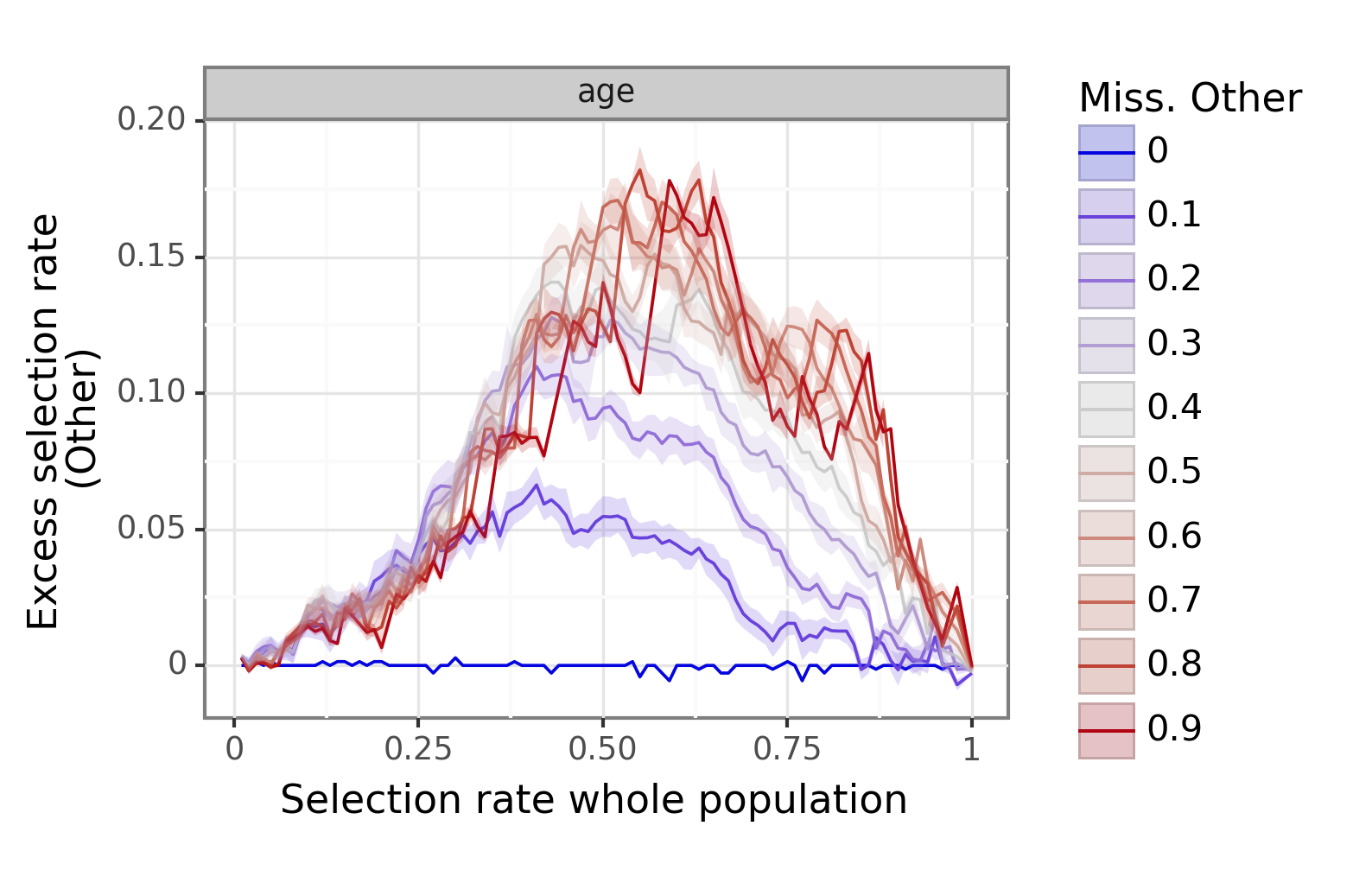}
    \includegraphics[scale=0.5]{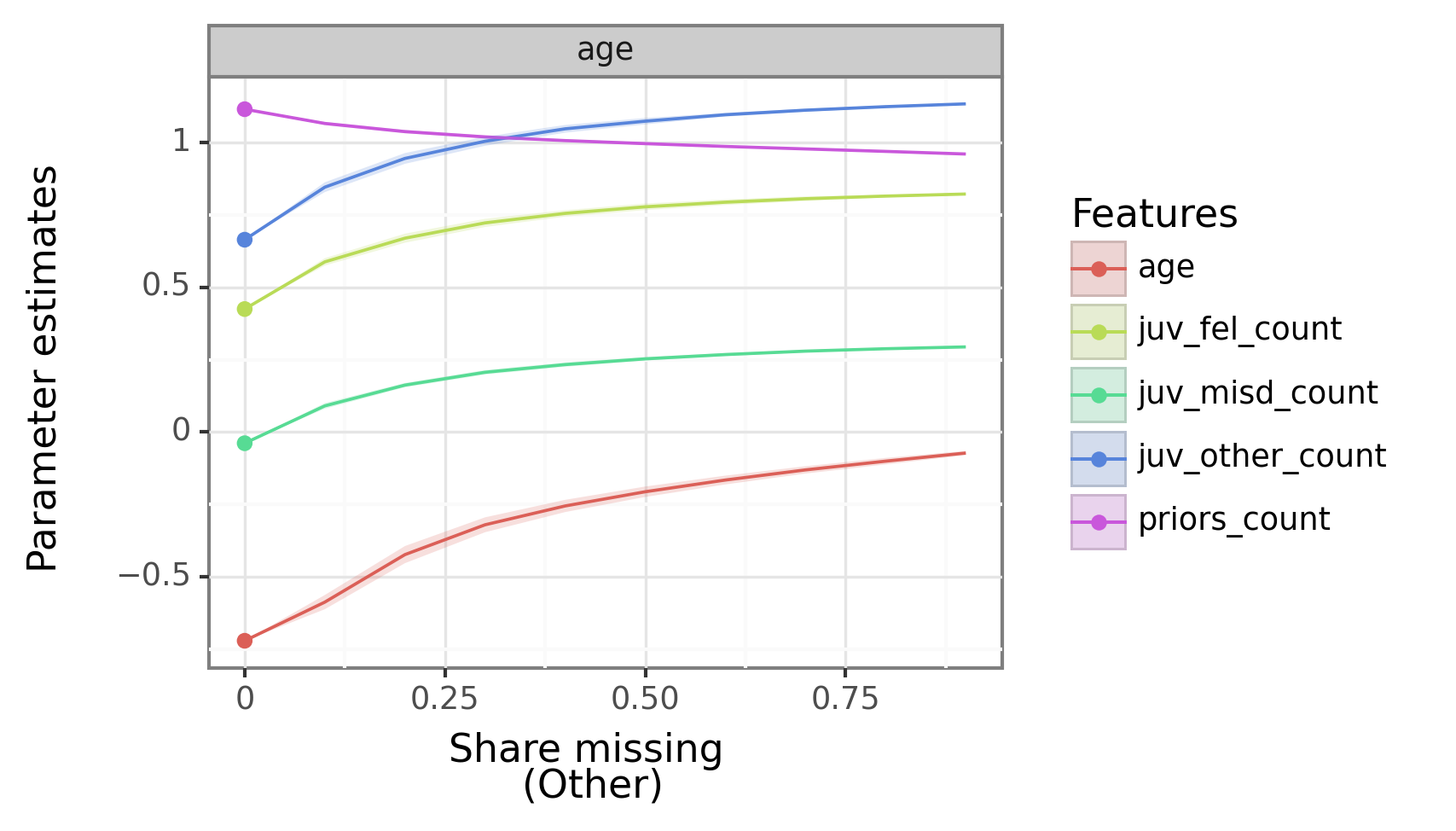}
    \caption{Excess selection rate of group Other (i.e. not African-American) at different population selection rates with synthetic outcomes using the COMPAS dataset, and the respective parameter estimates. Under-reporting is added to the feature `age' in group Other. Results are reported as averages over 30 simulation runs with shaded areas representing one standard deviation in each direction. Note that parameter estimates are only displayed for continuous count features and age to preserve readability. The models additionally take sex and the categorical feature charge degree into consideration.}
    \label{fig:case1compas}
\end{figure}

Figure~\ref{fig:case1compas} depicts the parameter estimates and excess selection rate of group Other (i.e. not African-American) when fitting a model on data with feature under-reporting in the feature `age' for group Other. We see that under-reporting in this setting leads to over-selection of the group with under-reporting. This over-selection is increasing with increasing levels of under-reporting. As the figure shows, the regression parameter for age is negative with an attenuation effect when under-reporting is injected. This means that in the semi-synthetic ground truth model and in the prediction models under under-reporting younger defendants are more likely to reoffend than older defendants. The feature correlations between age and juvenile crime counts (felony, misdemeanor, and other) are negative in the data while the correlation between age and the feature `priors count' is positive. This leads to parameter estimates that are increasing for increasing under-reporting in age for juvenile crime counts and decreasing for increasing under-reporting for priors count exactly as predicted by the theoretical analyses in Proposition~\ref{prop:propertieshatbetak}. Ultimately, this example shows how, in some settings, disparities may decrease as a function of under-reporting which aligns with Case 1 from the theoretical discussions in Section~\ref{sec:xx}. However, the example presented here is somewhat artificial and we find that typically disparities are increasing with differential feature under-reporting.

\end{document}